\documentclass[10pt,journal,compsoc]{IEEEtran}
\ifCLASSOPTIONcompsoc
  \usepackage[nocompress]{cite}
\else
  \usepackage{cite}
\fi
\ifCLASSINFOpdf
\else
\fi

\usepackage{amssymb}
\usepackage{hyperref}
\usepackage[dvipdfmx]{graphicx}
\usepackage{bmpsize}
\usepackage{mathtools}
\usepackage[table]{xcolor}
\usepackage{colortbl}
\usepackage{algorithmic}
\usepackage{algorithm}

\usepackage{booktabs}
\usepackage{multirow}
 \usepackage{enumitem}

\usepackage{tikz}
\usepackage{pgfplots}
\pgfplotsset{compat=1.17}
\usepgfplotslibrary{groupplots}
\pgfplotsset{
  every axis/.append style={
    no markers,
    grid=major,
    grid style={dashed},
    legend style={font=\tiny},
    ylabel style={font=\scriptsize},
    xlabel style={font=\scriptsize},
  },
  every axis plot/.append style={line width=1.2pt, line join=round},
  every axis legend/.append style={legend columns=1},
  group/group size=3 by 1,
  every x tick label/.append style={alias=XTick,inner xsep=0pt},
  every x tick scale label/.style={at=(XTick.base east),anchor=base west}
}

\definecolor{col1}{RGB}{53, 110, 175}
\definecolor{col2}{RGB}{204, 42, 42}
\definecolor{col3}{RGB}{255, 175, 35}
\definecolor{col4}{RGB}{79, 162, 46}
\definecolor{col5}{RGB}{97, 97, 97}
\definecolor{col6}{RGB}{103, 63, 153}
\definecolor{col7}{RGB}{0, 0, 0}
\definecolor{col8}{RGB}{123, 63, 0}

\tikzset{
  curve1/.style={col1},
  curve2/.style={col2},
  curve3/.style={col4},
  curve4/.style={col3},
  curve5/.style={col6},
  curve6/.style={cyan},
  curve7/.style={col5, dashdotted},
  curve8/.style={col7, dashed},
  curve9/.style={col8, densely dotted},
  curve10/.style={teal, densely dotted},
  curve11/.style={lime},
  curve12/.style={orange},
}

\newcommand{\lt}{<}
\newcommand{\gt}{>}

\newcommand{\discreteGradientField}{\mathcal{G}}
\newcommand{\discreteVectorField}{\mathcal{V}}
\newcommand{\dimensionality}{d}
\newcommand{\chainGroup}{\mathcal{C}}
\newcommand{\boundaryGroup}{\mathcal{B}}
\newcommand{\cycleGroup}{\mathcal{Z}}
\newcommand{\homologyGroup}{\mathcal{H}}

\newcommand{\domain}{\mathcal{K}}
\newcommand{\stableManifold}{\domain}
\newcommand{\unstableManifold}{\domain'}
\newcommand{\range}{\mathbb{R}}

\newcommand{\Star}{St}

\newcommand{\simplex}{\sigma}

\newcommand{\diagram}{\mathcal{D}}

\newcommand{\persistence}{\mathcal{P}}

\newcommand{\bigO}{\mathcal{O}}

\newcommand{\julien}[1]{\textcolor{black}{#1}}
\newcommand{\majorRevision}[1]{\textcolor{blue}{#1}}
\renewcommand{\majorRevision}[1]{\textcolor{black}{#1}}
\newcommand{\minorRevision}[1]{\textcolor{blue}{#1}}
\renewcommand{\minorRevision}[1]{\textcolor{black}{#1}}

\newcommand{\cutout}[1]{\textcolor{blue}{#1}}
\renewcommand{\cutout}[1]{}

\newcommand{\mycaption}[1]{
\vspace{-1.75ex}
\caption{#1}
}

\begin{document}
%
\title{Discrete Morse Sandwich:\\
Fast Computation of Persistence Diagrams for \julien{Scalar}
Data -- An Algorithm and A Benchmark}
%
%
%
%

\author{Pierre Guillou,
        Jules Vidal,
        and Julien Tierny
\IEEEcompsocitemizethanks{\IEEEcompsocthanksitem P. Guillou, J. Vidal and J.
Tierny are
with the CNRS and Sorbonne Université.
E-mail: \href{mailto:pierre.guillou@sorbonne-universite.fr,
jules.vidal@sorbonne-universite.fr,
julien.tierny@sorbonne-universite.fr}{\{firstname.lastname\}@sorbonne-universite.fr}
}
\thanks{Manuscript received May 21, 2021; revised May 11, 2021.}}

%
%

\markboth{Journal of \LaTeX\ Class Files,~Vol.~14, No.~8, May~2021}%
{Shell \MakeLowercase{\textit{et al.}}: Bare Demo of IEEEtran.cls for Computer Society Journals}
%



\IEEEtitleabstractindextext{%

\begin{abstract}
This paper introduces an efficient algorithm for persistence diagram computation,
given an input
piecewise linear scalar
field $f$ defined on a
$\dimensionality$-dimensional simplicial complex
$\domain$,
with $d \leq 3$.
\majorRevision{Our work revisits the seminal algorithm
\emph{``PairSimplices''} \cite{edelsbrunner02, zomorodianBook}
with discrete Morse theory (DMT) \cite{forman98, robins_pami11},}
%
%
which
\majorRevision{greatly}
reduces the number of input
simplices to consider.
\majorRevision{Further, we also extend to DMT and accelerate the stratification
strategy
described in \emph{``PairSimplices''} \cite{edelsbrunner02, zomorodianBook} for
the fast computation of
\minorRevision{the $0^{th}$ and $(d-1)^{th}$ diagrams, noted}
$\diagram_0(f)$ and $\diagram_{\dimensionality-1}(f)$.}
\minorRevision{Minima-saddle}
persistence
pairs ($\diagram_0(f)$) and saddle-maximum persistence pairs
($\diagram_{\dimensionality-1}(f)$) are efficiently computed by
processing\minorRevision{,} with a Union-Find\minorRevision{,} the unstable
sets of $1$-saddles and the stable
sets of $(\dimensionality-1)$-saddles.
\minorRevision{We}
provide a detailed description of the (optional) handling of
the boundary component of $\domain$ when processing
$(\dimensionality-1)$-saddles.
\majorRevision{This fast pre-computation \minorRevision{for} the dimensions $0$
and
$(\dimensionality-1)$ enables an aggressive
specialization of \cite{bauer13} to the 3D case,}
\majorRevision{which results in a drastic reduction of
%
the number of input simplices}
for the
computation of $\diagram_1(f)$,
the intermediate layer of the
\emph{sandwich}.
\majorRevision{Finally, we} document several performance improvements via
shared-memory
parallelism.
We provide an open-source implementation of our algorithm for reproducibility
purposes.
We also contribute
a reproducible benchmark package, which exploits three-dimensional data from a
public repository and compares our algorithm to a variety of publicly available
implementations.
Extensive experiments indicate that our algorithm improves  by two orders of
magnitude the time performance of the seminal
\majorRevision{\emph{``PairSimplices''}} algorithm
it extends.
Moreover, it also improves
memory footprint and
time
performance over a selection of
14
competing approaches,
with
a substantial gain
over the fastest available approaches, while
producing
a strictly identical output.
We illustrate the utility of our contributions with
an application to the fast
and robust extraction of persistent $1$-dimensional generators on surfaces, volume data and
high-dimensional
point clouds.
\end{abstract}


}

\maketitle

\IEEEdisplaynontitleabstractindextext

%
\IEEEpeerreviewmaketitle

\IEEEraisesectionheading{\section{Introduction}\label{sec:introduction}}

%
%
%
%

\begin{figure*}
\centering
\includegraphics[width=\linewidth]{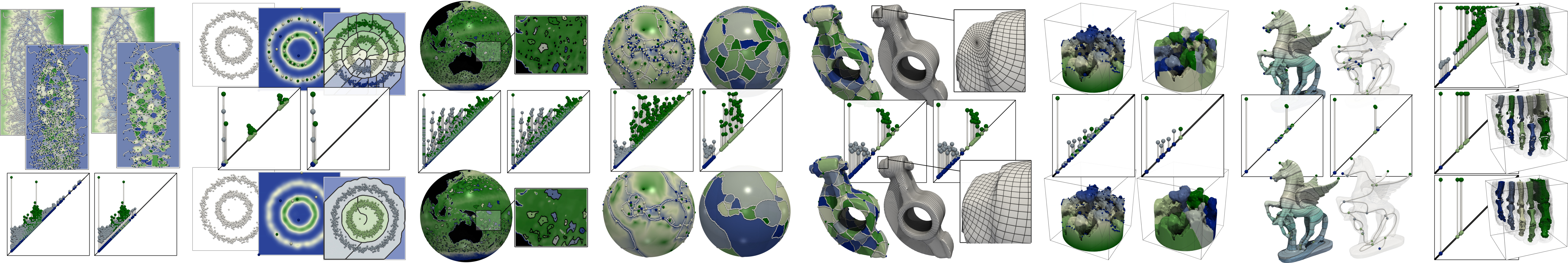}
\mycaption{Panorama of topological analysis pipelines in visualization and
graphics applications.
From left to right:
cell segmentation in microscopy data with the Morse-Smale complex (acquired 2D image),
persistence-driven clustering with the Morse-Smale complex (2D point cloud),
vortex extraction with the merge tree in climate data (acquired sea surface height, triangulated surface),
tectonic plate extraction with the Morse-Smale complex (simulated data, triangulated surface),
quad-meshing with the Morse-Smale complex (eigenfunction of the Laplace Beltrami operator, triangulated surface),
viscous finger extraction with the Morse-Smale complex (simulated data, tetrahedral mesh),
skeleton extraction with the Reeb graph (harmonic function, triangulated surface),
bone segmentation in a CT scan with the merge tree (acquired 3D image).
In each case, topological persistence plays a central role to distinguish features from noise,
enabling multi-scale analysis (left: original diagram, right: simplified diagram).
}
\label{fig_ttkOverview}
\end{figure*}

\IEEEPARstart{S}{\julien{calar}}
data is central to many fields of science and
engineering. It can be the result of an \emph{(i)} 
acquisition process (examples include CT-scans produced in medical imaging or 
one-dimensional time-series produced by punctual sensors) or it can be the 
result of a \emph{(ii)} numerical computation (examples include simulations in 
computational fluid dynamics, material sciences, etc). 
In both cases, the data is typically provided as a
\julien{low-dimensional}
scalar field (1D, 2D, or 3D) defined on the
vertices of either \emph{(i)} a regular grid (e.g. pixel or voxel images) or
\emph{(ii)} a mesh (e.g. polyhedral surfaces and volumes, AMR grids, etc.). An 
established strategy to generically process either cases of data provenance is 
to subdivide each cell of the input domain into simplices  \cite{freudenthal42, 
kuhn60}, hence converting the input data into a generic representation that 
facilitates subsequent processing, namely a piecewise linear scalar field 
defined over
a simplicial complex
(i.e. poly-lines in 1D,
triangulated surfaces in 2D and  tetrahedral meshes in 3D). However, such scalar
fields are provided in the applications with an ever-increasing size and 
geometrical complexity, which significantly challenges their interpretation by 
human users. This motivates the design of advanced data analysis tools, to 
support the interactive exploration and analysis of the features of interest 
present in large datasets. This is precisely the purpose of Topological Data
Analysis (TDA) \cite{edelsbrunner09}, which provides a toolbox of techniques for 
the generic, robust, and efficient extraction of structural features in data.

Topological methods have been investigated by the visualization community for 
more than twenty years \cite{heine16}, with applications to a variety of 
domains, including combustion \cite{laney_vis06, bremer_tvcg11, gyulassy_ev14}
fluid dynamics \cite{kasten_tvcg11, fangChen13}
material sciences \cite{gyulassy07, Lukasczyk17},
chemistry \cite{chemistry_vis14, harshChemistry, Malgorzata19}, 
or astrophysics
\cite{sousbie11, shivashankar2016felix} to name a few. Several 
topological data representations studied in TDA (such as the persistence 
diagram \cite{edelsbrunner09}, 
the contour tree \cite{tarasov98, carr00, gueunet_tpds19},
the Reeb graph \cite{reeb1946points, pascucci07, biasotti08, parsa12, 
gueunet_egpgv19} or the
 Morse-Smale complex \cite{BremerEHP04, GyulassyNPBH06, ShivashankarN12, 
gyulassy_vis14, Defl15, gyulassy_vis18}) have been specialized and used 
successfully in visualization, in particular for the explicit extraction and 
visual representation of structural patterns hidden in the data. An important
aspect of TDA is its ability to provide multi-scale hierarchies of the 
above topological data representations, which consequently enables multi-scale 
visualization, exploration and analysis. In that setting, \emph{Topological 
Persistence} \cite{edelsbrunner09} is an established importance measure which 
enables to distinguish the most salient topological structures present in the 
data from those corresponding to noise. In typical analysis pipelines (as 
shown in \autoref{fig_ttkOverview}), this importance measure drives the 
simplification of the above topological representations, resulting in 
interactive, multi-scale data explorations. In practice, topological 
persistence can be obtained by computing \emph{Persistence Diagrams} 
\cite{edelsbrunner09}.
Several algorithms have been proposed for their computation (see 
\autoref{sec_relatedWork}) and many software packages are publicly available. 
However, most of them generically target data defined in arbitrary dimension 
(with a specific focus towards high dimensional point clouds) and
provide only limited specialization for
low-dimensional
\julien{scalar}
data.

In this work, we introduce a novel algorithm for the fast computation of 
persistence diagrams for
\julien{scalar}
data defined on 1, 2 or 3-dimensional
domains.
In contrast to previous work, our approach specifically takes
advantage of the low dimensionality of
\julien{typical scalar}
data
\majorRevision{by revisiting a stratification strategy \cite{edelsbrunner02,
edelsbrunner09, bauer13}, that we call \emph{sandwiching}, from the perspective
of
discrete Morse theory \cite{forman98} (see \autoref{sec_overview} for an
overview).}
%
Our algorithm has several advantages over existing approaches. Our extensive 
experiments (\autoref{sec_results}) demonstrate a substantial gain over existing algorithms,
both in memory footprint and time performance, while
delivering a strictly identical
output. Moreover, it is output sensitive and
most of its internal procedures can be accelerated with shared-memory 
parallelism (\autoref{sec_parallel}). For reproducibility purposes, we provide a
C++
implementation of our approach. We also contribute a benchmark package,
which exploits three-dimensional data from a public repository 
\cite{openSciVisDataSets} and compares our approach to a variety of publicly
available
implementations.
We believe
such a
benchmark
has the
potential to become a reference experiment for future work on the topic.
Finally, we present an application (\autoref{sec_applications}) to the fast and robust extraction of
generators for surfaces, volume data or high-dimensional point clouds.


\subsection{Related work}
\label{sec_relatedWork}
This section describes the literature related to our work. First, we provide a 
quick overview of the usage of persistent homology in data visualization. 
Second, we briefly review the related computational methods.

\noindent
\textbf{Persistent Homology in Visualization}
Persistent Homology has originally been
introduced independently by several research groups \cite{B94, frosini99, robins99,
edelsbrunner02}. In many applications involving data analysis, 
topological persistence quickly established itself as an appealing importance 
measure that helps distinguish salient topological structures
in the data.

In data visualization, except a few approaches dealing with graph layouts 
\cite{graphLayout} and dimensionality reduction  \cite{OesterlingHJSH11, 
topoMap}, Persistent 
Homology has been mostly used in previous work in \emph{scientific 
visualization}, typically dealing with the interactive visual analysis of 
\julien{scalar}
data (coming from acquisitions or simulations). In that context,
topological persistence is typically used as a measure of importance driving 
the simplification of the input data itself \cite{tierny_vis12, 
Lukasczyk_vis20}, or the multi-scale hierarchical representation of topological 
abstractions \cite{heine16}, such as 
contour trees \cite{tarasov98, carr00, gueunet_tpds19},
Reeb graphs \cite{reeb1946points, pascucci07, biasotti08, parsa12, 
gueunet_egpgv19} or
 Morse-Smale complexes \cite{BremerEHP04, GyulassyNPBH06, ShivashankarN12, 
gyulassy_vis14, Defl15, gyulassy_vis18}. For instance, in the \emph{``Topology 
ToolKit''} (TTK) \cite{ttk17, ttk19} (an open-source library for topological 
data analysis and visualization), data is typically pre-simplified 
interactively, by removing low persistence features
\cite{tierny_vis12, 
Lukasczyk_vis20}, yielding a multi-scale hierarchy 
for the subsequent topological data representations 
(\autoref{fig_ttkOverview}). 
Similar analysis pipelines have been documented in a
number of applications, including 
combustion \cite{laney_vis06, bremer_tvcg11, gyulassy_ev14}
fluid dynamics \cite{kasten_tvcg11, fangChen13}
material sciences \cite{gyulassy07, Lukasczyk17},
chemistry \cite{chemistry_vis14, harshChemistry, Malgorzata19}, 
or astrophysics
\cite{sousbie11, shivashankar2016felix}. Topological persistence has also been 
used as an importance measure in several other
\julien{scalar}
data analysis tasks,
such as data segmentation  \cite{topoAngler, carr04}, isosurface extraction 
\cite{vanKreveld97}, data compression \cite{soler_pv18} or transfer function 
design for volume rendering \cite{WeberDCPH07}. 
The persistence diagram (\autoref{sec_persistenceDiagram}) is a popular 
topological data representation, which concisely and robustly captures the 
number and salience of the features of interest present in the data. As such, 
it is an effective visual descriptor of the population 
of features in data, for ensemble summarization \cite{favelier_vis18, 
vidal_vis19, KontakVT19} or feature tracking  \cite{soler_ldav18, soler_ldav19, 
LukasczykGWBML20}.

\noindent
\textbf{Algorithms for Computing Persistent Homology}
In general,
the standard
approach to the computation
of persistent homology involves the reduction of the boundary matrix
\cite{edelsbrunner09} (which
describes the facet/co-facet relations between the simplices of the input
domain). This approach is now the core procedure of many software packages. This 
includes for instance
\emph{PHAT} \cite{BauerKRW17} and
\emph{Dipha} \cite{dipha} (which feature
additional accelerations \cite{dualities, twist}, along with specific data 
structures for cubical cell complexes \cite{wagner11}), \emph{Gudhi} 
\cite{gudhi} (which also features specific accelerations \cite{SilvaMV11, 
DeyFW14, BoissonnatDM15, BoissonnatP20} and data structures 
\cite{BoissonnatM14}, in particular for cubical cell complexes \cite{wagner11}) 
and others \cite{dionysus2, javaplex}. Certain packages have a special focus 
towards
the persistent homology of Rips filtrations of high-dimensional point clouds,
such as \emph{Ripser} \cite{ripser} (adapted to cubical complexes
\cite{cubicalRipser}) or \emph{Eirene} \cite{eirene, henselmanghristl6}.
They have been integrated in
several data analysis libraries \cite{scikittda2019, TauzinLTPCMDH21}.
\majorRevision{Among the above techniques, several documented accelerations
share some
conceptual similarities with our approach.
Specifically, Bauer et al. \cite{bauer13} accelerate the global computation by
pre-computing persistence pairs (\autoref{sec_persistenceDiagram}) on localized
\emph{chunks} of the data. Similarly, Bauer
\cite{ripser} introduces the notion of \emph{apparent pairs}, which are
persistence pairs  involving a simplex and one
of its co-facets, and which can be efficiently detected in a pre-process.
Our approach uses
discrete
Morse theory \cite{forman98} to serve a similar purpose.
As further detailed in \autoref{sec_saddleSaddle}, our pre-computation
focuses on simplices
\emph{involved} in
zero-persistence pairs (specifically, within each vertex lower star),
but
then
pair them
greedily
according to
the expansion-based discrete gradient \cite{robins_pami11} of the
data. Bauer et al.
\cite{bauer13}
introduce a stratification strategy (``\emph{clearing} and
\emph{compression}'')
when computing the persistence diagram
in
dimension $p$,
which discards $(p+1)$-simplices
(respectively $p$-simplices) which
already
create (respectively destroy)
persistence pairs (typically computed within the above chunks).
Our
stratification strategy can be interpreted as an aggressive specialization
of the above
strategy for three-dimensional data. Specifically, when computing for the
dimension $1$, not only does our algorithm discard from the computation the
simplices involved in zero-persistence pairs, but it also discards \emph{all}
the simplices
involved in zero and two-dimensional persistence pairs.}
Some methods support parallel computations  \cite{BauerKRW17, dipha, Nigmetov20,
oineus}. All
\majorRevision{the above approaches}
are included in our benchmark
(\autoref{sec_benchmark}).

For low dimensional data, such as
\julien{typical \majorRevision{3D} scalar}
fields, specific
\majorRevision{stratification}
strategies can be considered.
\majorRevision{In their seminal paper introducing the algorithm
\emph{``PairSimplices''} \cite{edelsbrunner02}, Edelsbrunner et al. observe
that the persistence diagram can be efficiently computed for the dimension $0$
with a Union-Find data structure \cite{cormen}. They also observe that, by
symmetry, a Union-Find can also be used for the dimension $2$ (without
specifying, however, how these computations interact with the diagram of
dimension $1$).
This strategy has been the default computation method
in TTK \cite{ttk17, ttk19} since its initial release,
where the sub-level set
components of $f$ and $-f$ are
efficiently tracked
with
a parallel merge tree  algorithm
\cite{gueunet_ldav16, gueunet_tpds19}.
Recently,
Vidal et al.  presented \emph{progressive} \cite{vidal_tvcg21} and
\emph{approximate} \cite{vidal_ldav21}
variants,
based on a multiresolution representation of the input.
In this paper, we revisit this stratification strategy (originally introduced by
Edelsbrunner et al. \cite{edelsbrunner02}),
but from the perspective of discrete Morse theory \cite{forman98},
and we further accelerate it by restricting the sub-level set connectivity
tracking
%
%
to the unstable (and
stable) sets of $1$ and $(\dimensionality - 1)$ saddles.
This acceleration can be interpreted as an adaptation to discrete Morse theory
of earlier work on monotone paths
for merge tree
construction
\cite{ChiangLLR05, MaadasamyDN12,
CarrWSA16, smirnov17, vidal_tvcg21}.
}

Similarly to our work, previous approaches have investigated Morse theory 
\cite{morseQuote, milnor63}, specifically its discrete version 
\cite{forman98}, to accelerate the computation of persistence diagrams of 
scalar data. 
\majorRevision{To apply discrete Morse theory (DMT), one first needs to
compute a discrete gradient (\autoref{sec_discrete_morse_theory}) from the input
scalar data. For this, several algorithms have been proposed
\cite{gyulassy_vis08, robins_pami11, shivashankarTVCG12, ShivashankarN12,
gyulassy_vis14, ttk17, gyulassy_vis18}.
Shivashankar et al. \cite{shivashankarTVCG12, ShivashankarN12}
introduced the \emph{``AssignGradient''}
algorithm, which constructs discrete vectors
equivalent
to the \emph{apparent pairs} of Bauer \cite{ripser}.
However,
in practice, even for smooth datasets, this algorithm
generates
a number of critical simplices several orders of magnitude
larger than the number of piecewise linear (PL) critical points \cite{ttk17},
resulting in
a
large
majority of \emph{spurious} zero-persistence pairs in
the persistence diagrams (i.e. pairs not captured by the lower star
filtration, \autoref{sec_lowerStarFiltration}).
To address this, Tierny et al. \cite{ttk17}
performed an explicit reversal of the \emph{v-paths}
(\autoref{sec_discrete_morse_theory}) involving \emph{spurious} critical
simplices
(i.e.
not located in the star of any PL critical point).}
Robins et al. \cite{robins_pami11, diamorse}
\majorRevision{introduced}
an
algorithm for computing a discrete gradient field, \majorRevision{including an
implicit, localized gradient reversal (called \emph{``ProcessLowerStars''}).
This algorithm provides the
nice property that it generates \emph{no} spurious critical simplex: all the
resulting critical simplices are guaranteed to be located within the star of a
PL critical point (we directly exploit this property in our work,
\autoref{sec_saddleSaddle}). Hence, these simplices}
exactly
coincide with the topological changes of the lower star filtration of the
data \majorRevision{(\autoref{sec_lowerStarFiltration})}.
\majorRevision{Then,}
all the topological events occurring in the filtration of the
scalar
data can
be equivalently encoded with a filtration of its discrete Morse complex. As
\majorRevision{this}
complex is usually smaller in practice than the input data, this
pre-process
reduction procedure accelerates
subsequent, traditional algorithms for persistent homology 
\cite{zomorodianBook}.
Extensions of this idea have been investigated \cite{GuntherRWH12, 
perseusPaper, perseus, iuricich21},
\majorRevision{e.g.}
for the support of high
dimensional data.
In contrast,
our
work
specifically takes advantage of the low dimensionality of the data to
expedite \majorRevision{the process (to avoid the computation of the
\emph{full} Morse complex),
with
a stratification strategy adapted  from
\emph{``PairSimplices''}
to
discrete
Morse theory,
which we further accelerate
by restricting a Union-Find processing to the
unstable (and stable) sets of $1$ and $(d-1)$ saddles
(\autoref{sec_extremumDiagrams}).}

\begin{figure*}
\includegraphics[width=\linewidth]{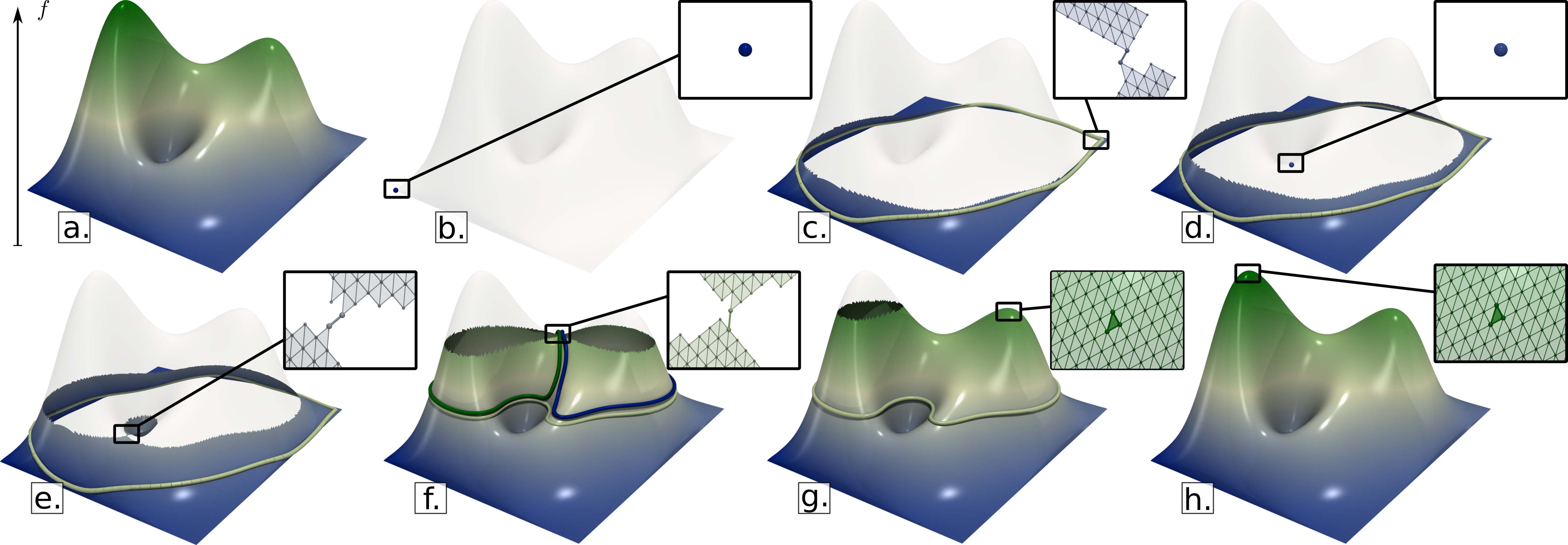}
 \mycaption{Lexicographic filtration of a toy example (elevation $f$ on a
terrain $\domain$, \emph{(a)}).
 At step \emph{(b)}, the introduction of the first vertex in the filtration
$\domain_b$ creates one connected component  and $\beta_0(\domain_b) = 1$. This
component later loops back to itself \emph{(c)}, creating a non trivial
$1$-cycle $c_c$ (light green). At this point, we have:
$\homologyGroup_1(\domain_c) = \{ 0, c_c \}$
 and $\beta_1(\domain_c) = rank\big(\homologyGroup_1(\domain_c)\big) =
log_2(|\homologyGroup_1(\domain_c)|) = 1$. At step \emph{(d)}, a new connected
component is created and  $\beta_0(\domain_d) = 2$. At step \emph{(e)}, two
connected components merge into one and $\beta_0(\domain_e) = 1$. At step
\emph{(f)}, the connected component of $\domain_f$ loops back to itself,
yielding three, independent, non trivial $1$-cycles $c_c$ (light green), $c_f$
(dark green) and $c_{f'}$ (dark blue). At this point, we have:
$\homologyGroup_1(\domain_f) = \{ 0, c_c, c_f, c_{f'} \}$
 and $\beta_1(\domain_f) = rank\big(\homologyGroup_1(\domain_f)\big) =
log_2(|\homologyGroup_1(\domain_f)|) = 2$. At step \emph{(g)}, the introduction
of a triangle fills the \emph{``hole''} left by the homology class $c_{f'}$
(dark blue, step \emph{(f)}), which
 becomes trivial
 and disappears. Moreover, the class $c_f$ (dark green, step \emph{(f)}) becomes
homologous to an \emph{older} class, $c_c$ (light green), and thus disappears
and we have $\beta_1(\domain_g) = 1$. Finally, at step \emph{(h)}, the
introduction of the last triangle fills the \emph{``hole''} left by the homology
class $c_c$ (light green, step \emph{(g)}) and we eventually have
$\beta_0(\domain_h) = 1$ and $\beta_1(\domain_h) = 0$. The persistent diagram
(\autoref{fig_diagram}) keeps track of all these events and records the birth,
death and overall lifespan of the topological features responsible for changes
in Betti numbers.}
%
 \label{fig_lowerStar}
\end{figure*}

\subsection{Contributions}
This paper makes the following new contributions:
\begin{enumerate}
  \item{\emph{A fast algorithm for the computation of persistence diagrams for
1D, 2D or 3D
\julien{scalar}
data}:
  \begin{itemize}
  \item \majorRevision{Our approach revisits the
algorithm
\emph{``PairSimplices''} \cite{edelsbrunner02,
zomorodianBook}
with
discrete Morse theory \cite{forman98, robins_pami11}.}
   \majorRevision{Simplices involved in zero-persistence pairs are efficiently
skipped,
which
drastically
reduces
the number of input simplices
to consider (\autoref{sec_saddleSaddle}).}
    \item
    \majorRevision{The persistence diagrams for the dimensions $0$ and
$(\dimensionality
- 1)$ are efficiently computed by restricting a Union-Find \cite{cormen}
processing to the unstable (and stable) sets of $1$ and $(\dimensionality
-
1)$ saddles (\autoref{sec_extremumDiagrams}).
This enables an aggressive specialization of \cite{bauer13} to the 3D case,
which
accelerates
the computation in dimension $1$ by further reducing the number of input
simplices.}
%

%
%
  \end{itemize}
  \majorRevision{Our algorithm provides
  practical
  gains
  over
reference
methods. It is output sensitive and
its sub-routines can be
 accelerated with shared-memory parallelism.
  }}
  \item{\emph{An open-source implementation}:
For reproduction purposes, we provide a C++
implementation
of our approach, which is officially integrated in the source tree of TTK \cite{ttk17, ttk19}
%
(Github commit: 
\majorRevision{\href{https://github.com/topology-tool-kit/ttk/commit/bb3089f07a5039433dfab6bab7455e23678ec6b3}{bb3089f}}).
}
  \item{\emph{A reproducible benchmark:}
We provide a  Python benchmark package
(\href{https://github.com/pierre-guillou/pdiags_bench}{https://github.com/pierre-guillou/pdiags\_bench}), which
uses three-dimensional data from a public repository
\cite{openSciVisDataSets} and compares the running times, memory footprints and 
output diagrams of a variety of publicly available implementations for 
persistence diagram computation. This reproducible benchmark may be used as a 
reference experiment for
future developments on the topic.}
\end{enumerate}

\section{Preliminaries}
This section presents the theoretical background of our work. It contains
definitions adapted from the Topology ToolKit \cite{ttk17, ttk19}. We refer the
reader to textbooks \cite{edelsbrunner09, zomorodianBook} for comprehensive
introductions to
computational topology.

\subsection{Input data}
\label{sec_inputData}
The input data is provided as a piecewise linear (PL) scalar field $f : \domain 
\rightarrow \range$ defined on
$d$-dimensional simplicial complex $\domain$,
with $d \leq 3$.
As discussed in the introduction,
this input representation generically and
homogeneously supports all types of
\julien{typical scalar}
data, in 1D, 2D or 3D, coming from
either acquisitions or numerical simulations. When the data is given on 
arbitrary cell complexes, cells are subdivided into simplices. In particular, 
regular grids are 
triangulated according to the Freudenthal 
triangulation \cite{freudenthal42, kuhn60} (yielding a 6-vertex neighborhood in 
2D and a 14-vertex neighborhood in 3D). 
Note that this triangulation is performed implicitly (i.e. no memory overhead), by emulating the
simplicial structure upon traversal queries \cite{ttk17}.

The input scalar field $f$ is typically provided on 
the vertices of $\domain$ and interpolated on the simplices of higher 
dimension. $f$ is also assumed to be injective on the vertices of $\domain$,
which is easily achieved in practice with a symbolic perturbation inspired from 
Simulation of Simplicity \cite{edelsbrunner90}.

\subsection{Lexicographic filtration}
\label{sec_lowerStarFiltration}
Given the input function $f$, a global order between the simplices of
$\domain$ can be introduced by
considering the so-called \emph{lexicographic} comparison, as detailed below.

Given a $d$-simplex $\sigma \in \domain$, let us consider the sequence
$\{f\big(v_0(\sigma)\big), f\big(v_1(\sigma)\big), \dots,
f\big(v_d(\sigma)\big)\}$ of its vertex data values, sorted in decreasing
order, where $f\big(v_i(\sigma)\big)$ denotes the $i^{th}$ largest value among
its vertices, i.e. $f\big(v_0(\sigma)\big) \gt f\big(v_1(\sigma)\big) \gt \dots
\gt
f\big(v_d(\sigma)\big)$.

Then, an order can be established between any two simplices $\sigma_i$ and
$\sigma_j$ by comparing the above sorted sequences. In particular,
$\sigma_i$
will be considered \emph{smaller} than $\sigma_j$ if $f\big(v_0(\sigma_i)\big)
\lt f\big(v_0(\sigma_j)\big)$.
On the contrary, if $f\big(v_0(\sigma_i)\big)
\gt f\big(v_0(\sigma_j)\big)$, $\sigma_i$
will be considered \emph{greater} than $\sigma_j$.
Otherwise,
if $f\big(v_0(\sigma_i)\big) =
f\big(v_0(\sigma_j)\big)$, a tiebreak needs to be performed and
the order will be decided by iteratively considering, similarly,
the following vertices in the sequence (i.e. $v_k(\sigma_i)$ and
$v_k(\sigma_j)$, with $k \in \{1, \dots, d\}$) until the conditions
$f\big(v_k(\sigma_i)\big) \lt
f\big(v_k(\sigma_j)\big)$ (i.e. $\sigma_i$ is smaller than $\sigma_j$) or
$f\big(v_k(\sigma_i)\big) \gt
f\big(v_k(\sigma_j)\big)$ (i.e. $\sigma_i$ is greater than $\sigma_j$) are
satisfied. In the case where the dimensions $d_i$ and $d_j$ of $\sigma_i$ and
$\sigma_j$ are such that $d_i \lt d_j$ and that $f\big(v_k(\sigma_i)\big) =
f\big(v_k(\sigma_j)\big), \forall k\in \{0, \dots,  d_i\}$ (i.e. $\sigma_i$
is a \emph{face} of $\sigma_j$), then $\sigma_i$ is considered \emph{smaller}
than $\sigma_j$. Since $f$ is injective on the vertices of $\domain$
(\autoref{sec_inputData}), this lexicographic comparison guarantees
a strict total order on the set of
simplices of $\domain$, such that all the faces of a simplex $\simplex$ are by
construction smaller than $\simplex$.

Let $\domain_i$ be the union of the first $i$ simplices of $\domain$,
given the above comparison.
Then, the global lexicographic order induces a
nested sequence of simplicial complexes $\emptyset = \domain_0 \subset \domain_1
\subset
\dots \subset \domain_n = \domain$ (where $n$ is the number of simplices of
$\domain$), which we call the \emph{lexicographic filtration} of $\domain$.
Intuitively, it can be seen as a time-varying process, where the
simplices of $\domain$ are added one by one, given the lexicographic comparison
of the vertex data values.

A central idea in Topological Data Analysis consists in encoding the evolution
of the topological structures of $\domain_i$ (for
\julien{typical scalar}
data: its connected
components, handles and voids, see \autoref{sec_homology})  \emph{along} the
filtration, as $i$ increases
from $0$ to $n$. In particular, as shown in \autoref{fig_lowerStar},
connected components progressively merge, handles get closed and voids get
filled. This evolution is captured by the persistence diagram introduced later
(\autoref{sec_persistenceDiagram}).

We now discuss an alternate filtration, often considered in previous work
\cite{edelsbrunner09, robins_pami11} and
we describe its relation (used in \autoref{sec_saddleSaddle}) to the
lexicographic filtration considered here.
Let $\Star(v)$ be the star of a vertex $v$, i.e. the set of all its
co-faces $\simplex$: $\Star(v) = \{ \simplex \in \domain ~ | ~ v < \simplex
\}$. Let $\Star^{-}(v)$ be the \emph{lower} star of $v$. It is the subset of
the simplices of the star of $v$, for which $v$ is the vertex with highest $f$
value: $\Star^{-}(v) = \{ \simplex \in \Star(v) ~|~ \forall u \in \sigma,
f(u) \leq f(v)\}$. Since $f$ is assumed to be injective on the vertices of
$\domain$, it follows that each simplex $\sigma \in \domain$ belongs to a
unique lower star.
Let $\domain'_j$ be the union of the first $j$ lower stars, i.e. the union of
the lower stars of the $j$-th lowest vertices of $f$. Then, the nested
sequence of simplicial complexes $\emptyset = \domain'_0 \subset \domain'_1
\subset \dots \subset \domain'_{n_v} = \domain$ (where $n_v$ is the number of
vertices in $\domain$) is called the \emph{lower star filtration} of $f$
\cite{edelsbrunner09}.
$\domain'_j$ is homotopy equivalent to the
sub-level
sets of $f(v_j)$ \cite{edelsbrunner09}  and
the topological
changes occurring in $\domain'_j$ during the lower star filtration thus
precisely
occur at the PL critical points \cite{banchoff70} of $f$.
Given
the above definition, it follows that each sub-complex $\domain'_j$ of the
lower star
filtration is equal to the sub-complex $\domain_{i-1}$ of the lexicographic
filtration, where $\sigma_i$ is the \emph{vertex} immediately after $v_j$ in the
global
vertex order.
In other words, the lower star filtration introduces simplices
by
\emph{chunks} of lower stars   (\autoref{fig_zeroPersistencePairs}), while the lexicographic filtration introduces
them one by one, yet in a compatible order.
Then, it follows that each PL critical point includes in its
lower star a simplex whose introduction via the lexicographic filtration
changes the topology of $\domain_i$.

\begin{figure}
\centering
\includegraphics[width=\linewidth]{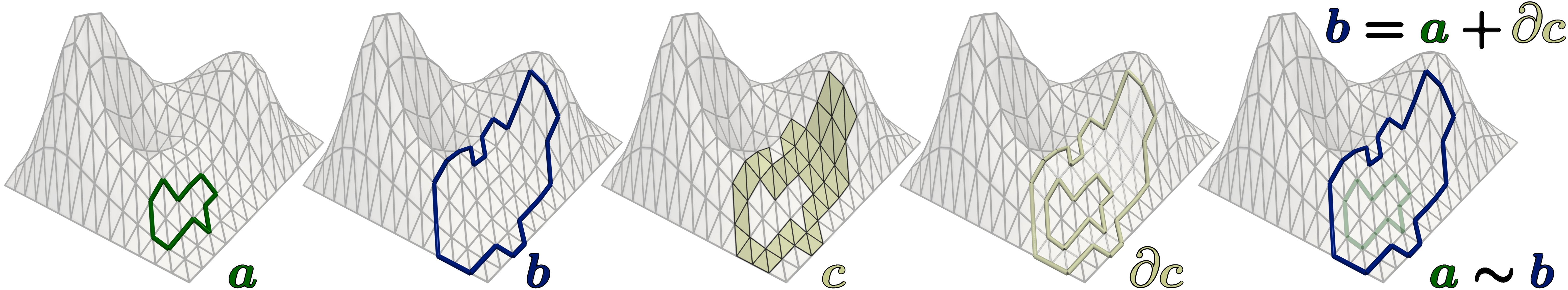}
 \mycaption{\majorRevision{Two $1$-cycles $a$ and $b$ are \emph{homologous}
(noted $a \sim b$) if there exists a $2$-chain $c$, such that $b = a + \partial
c$.
 Here, $\partial c = a + b$.
 By adding $a$ on both sides (modulo 2
coefficients), we indeed have: $b = a + \partial c$, thus $a \sim b$.}
%
 }
 \label{fig_homologousCycles}
\end{figure}

\subsection{Homology groups}
\label{sec_homology}
The topology of a simplicial complex can be described with its
homology groups, briefly summarized here from \cite{edelsbrunner09}.

We call a $p$-chain $c$ a formal sum (with modulo 2 coefficients) of 
$p$-simplices $\simplex_i$ of $\domain$: $c = \sum \alpha^c_i \simplex_i$ with
$\alpha^c_i \in \{0, 1\}$. 

Two 
$p$-chains $c =  \sum \alpha^c_i \simplex_i$ and $c' =  \sum \alpha^{c'}_i 
\simplex_i$ can be summed together componentwise
to 
form a new $p$-chain $c'' = \sum \alpha^{c''}_i \simplex_i$, where 
$\alpha^{c''}_i = \alpha^c_i + \alpha^{c'}_i$ and $\alpha^{c''}_i \in \{0, 1\}$.
Intuitively, a $p$-chain can be interpreted
as a selection of $p$-simplices,
modeled
with a bit mask,
where a $p$-simplex is present
in the selection
(i.e. with its coefficient valued at $1$) only if it has been 
added an odd number of times. Then, the set of all possible $p$-chains of 
$\domain$ (along with their modulo 2 addition) forms the \emph{group of
chains},
noted $\chainGroup_p(\domain)$.

The boundary of a
$p$-simplex $\sigma_i$, noted $\partial \sigma_i$, is given by the sum
of its faces of dimension $(p-1)$. Then, the \emph{boundary} of a $p$-chain 
$c$, noted $\partial c$, is the sum of the boundaries of the simplices of $c$:
$\partial c = \sum \alpha^c_i \partial \sigma_i$. Note that $\partial c$ 
is itself a $(p-1)$ chain, i.e. $\partial : \chainGroup_p(\domain) 
\rightarrow \chainGroup_{p-1}(\domain)$, and that
the boundary operator
commutes with addition, i.e. $\partial (c + c') = \partial c + \partial c'$.

A $p$-cycle $c$ is a $p$-chain such that $\partial c = 0$ and the group of all 
possible $p$-cycles is noted $\cycleGroup_p(\domain)$. 
A $p$-boundary is a 
$p$-chain $c \in \chainGroup_p(\domain)$ which is the boundary of a 
$(p+1)$-chain $c' \in \chainGroup_{p+1}(\domain)$: $c = \partial c'$. The group 
of $p$-boundaries is noted $\boundaryGroup_p(\domain)$.
The 
fundamental lemma of homology states that $\partial \partial c = 0$ for every 
$p$-chain $c$, for any $p$ \cite{edelsbrunner09}. This implies that 
$p$-boundaries are necessarily $p$-cycles ($\boundaryGroup_p \subseteq
\cycleGroup_p$), but not the other way around: all $p$-cycles are not 
necessarily $p$-boundaries. Such cycles are specifically captured with the 
notion of \emph{homology group}, which is the quotient group given by:
$\homologyGroup_p(\domain) =
\cycleGroup_p(\domain) / \boundaryGroup_p(\domain)$. 
Specifically, two $p$-cycles $a$ and $b$ of $\cycleGroup_p$ are called
\emph{homologous} (noted $a \sim b$), if $b = a + \partial c$ where $\partial c$ is a
$p$-boundary ($\partial c \in
\boundaryGroup_p$). Intuitively, this means that two cycles $a$ and $b$ are
homologous if one can be
transformed into the other, by the
addition of the boundary of a $(p+1)$ chain $c$ (\autoref{fig_homologousCycles}), as further discussed
in \autoref{sec_pairCells}.
The 
set of all cycles which are homologous defines a \emph{homology class} (from 
which anyone can be chosen as a
\majorRevision{representative}).
The \emph{order} of
$\homologyGroup_p(\domain)$ is given by its cardinality, i.e. the number
of homology classes.
Given the modulo-2 addition between
\majorRevision{representatives},
the \emph{rank} of
$\homologyGroup_p(\domain)$
is given by the maximum number of linearly independent
classes (called \emph{generators}) and it is called the  $p$-th Betti number of
$\domain$, noted
$\beta_p(\domain)$. Intuitively, the $p$-th Betti number gives the number of 
\emph{$p$-dimensional holes} in $\domain$, which cannot be filled with a 
$(p+1)$ chain of $\domain$.
In practice,
given a
$3$-dimensional simplicial complex
$\domain$
embedded in $\range^3$,
$\beta_0(\domain)$ corresponds to its number of 
connected components, $\beta_1(\domain)$ is its number of handles 
and $\beta_2(\domain)$ is its number of voids.


\begin{figure}
\centering
\includegraphics[width=\linewidth]{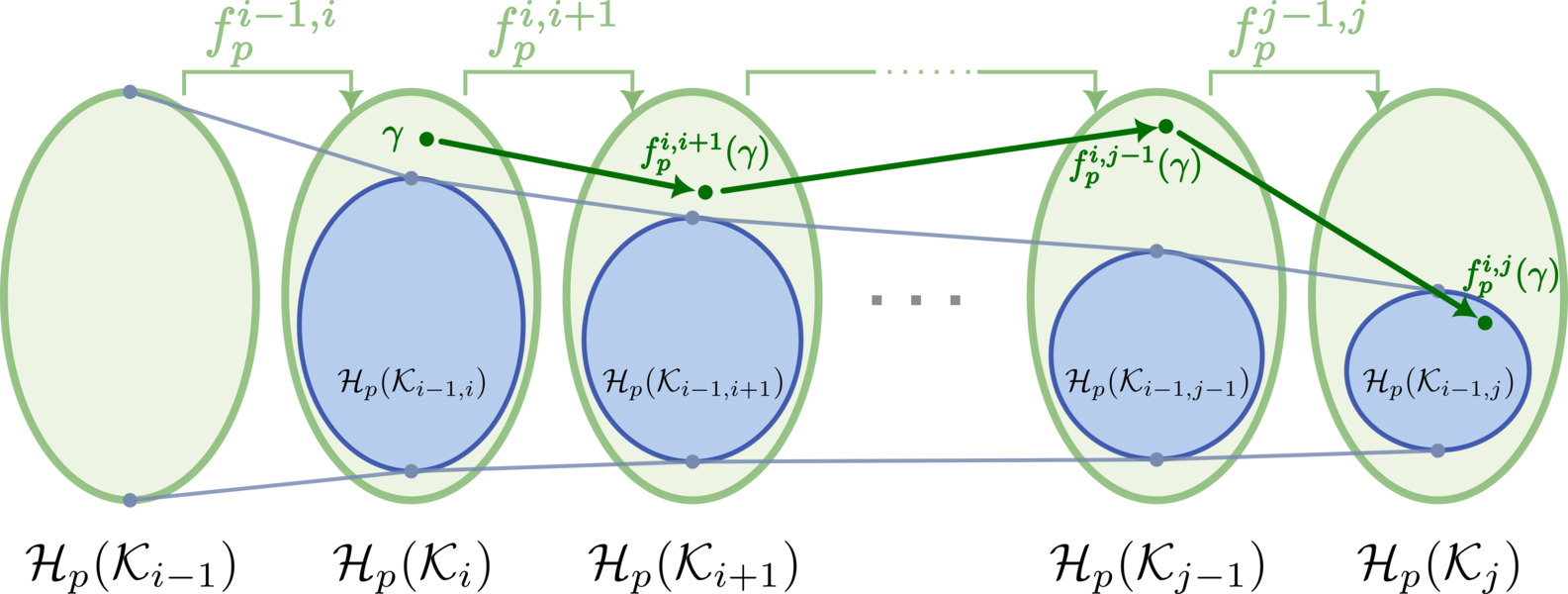}
\mycaption{Tracking homology classes along the filtration with homomorphisms
(dark green arrows, illustration adapted from \cite{edelsbrunner09}). The class
$\gamma$ is born in $\domain_i$: it is not the image through the homomorphism
$f_p^{i-1, i}$ of any pre-existing class (i.e.
it does not belong to $\homologyGroup_p(\domain_{i-1,i})$, blue set).
$\gamma$ dies in $\domain_j$ as it merges
with a pre-existing class, i.e. the image through the homomorphism $f_p^{i-1, j}$ of a
class already existing at step $i-1$ and still persistent at step $j$ (blue set).
In \autoref{fig_lowerStar}, the dark green cycle $c_f$ has a similar trajectory: it is born at step \emph{(f)} and at step \emph{(g)}, it \emph{merges} with the pre-existing cycle $c_c$ (i.e. $c_f$ becomes homologous to $c_c$).
}
\label{fig_persistentHomologyGroups}
\end{figure}

\subsection{Persistence diagrams}
\label{sec_persistenceDiagram}
Persistent diagrams are concise topological data representations which track
the
evolution of the homology groups during a filtration. In the remainder, we 
focus on the lexicographic filtration introduced in
\autoref{sec_lowerStarFiltration}.
Since $\domain_i \subseteq
\domain_j$ for any $0 \leq i \leq j \leq n$, it follows that there exists a
homomorphism \cite{edelsbrunner09} between the homology groups
$\homologyGroup_p(\domain_i)$ and $\homologyGroup_p(\domain_j)$, noted 
$f_p^{i, j} : \homologyGroup_p(\domain_i) \rightarrow 
\homologyGroup_p(\domain_j)$.
This
homomorphism $f_p^{i, j}$
keeps track of the relations between the
homology classes along the filtration, from $\domain_i$ to $\domain_j$ (\autoref{fig_persistentHomologyGroups}).


Formally, for any $0 \leq i \leq j \leq n$, the $p$-th persistent homology 
group, noted $\homologyGroup_p(\domain_{i,j})$, is the image of the 
homomorphism $f_p^{i, j}$, noted $\homologyGroup_p(\domain_{i,j}) =  f_p^{i,
j}\big(\homologyGroup_p(\domain_i)\big)$.


Specifically, we say that a homology class $\gamma$ is \emph{born at 
$\domain_i$} if $\gamma \in \homologyGroup_p(\domain_i)$ and $\gamma \notin 
\homologyGroup_p(\domain_{i-1, i})$ (see 
\autoref{fig_persistentHomologyGroups}):
$\gamma$ is present in $\homologyGroup_p(\domain_i)$ but it is not included in 
the image by $f_p^{i-1, i}$ of the homology groups of the previous complex in 
the filtration, $\domain_{i-1}$. In other words, $\gamma$ is present in 
$\homologyGroup_p(\domain_i)$ but it is not associated to any pre-existing 
class of $\homologyGroup_p(\domain_{i-1})$ by $f_p^{i-1, i}$.

Symmetrically,
we say that a homology class 
$\gamma$ born at $\domain_i$ \emph{dies at $\domain_j$} if \emph{(i)} $f_p^{i, 
j-1}(\gamma) \notin \homologyGroup_p(\domain_{i-1, j-1})$ and \emph{(ii)} 
$f_p^{i, j}(\gamma) \in \homologyGroup_p(\domain_{i-1, j})$, see \autoref{fig_persistentHomologyGroups}. In other words,
\emph{(i)} the class $\gamma$ did not exist prior to $i$ and \emph{(ii)} it 
merged (through $f_p^{j-1, j}$) with another, pre-existing class $\gamma'$, 
itself created before $i$. This destruction of a class upon its merge with 
another \emph{older} class is often called the \emph{Elder rule}
\cite{edelsbrunner09}.
Note that the birth of a $p$-dimensional homology class
$\gamma$ occurs on a $p$-simplex $\sigma_i$ of $\domain_i$,
while its death occurs on a $(p+1)$-simplex $\sigma_j$ of
$\domain_j$. The pair $(\sigma_i, \sigma_j)$ is called a \emph{ persistence
pair}.

The persistence of a homology class $\gamma$ which \majorRevision{is} born in
$\domain_i$ and
which
\majorRevision{dies}
in $\domain_j$ is given by the difference in the
corresponding scalar values $\persistence(\gamma) = f(v_j) - f(v_i)$, where 
$f(v_j)$ and $f(v_i)$ are respectively the maximum vertex data values of the
simplices $\sigma_j$ and $\sigma_i$.
Note that a homology class $\gamma$ which was born in $\domain_i$ and whose 
image by $f_p^{i, n}$ is still included in
$\homologyGroup_p(\domain_{i,
n})$ is said to have \emph{infinite persistence} (i.e. it is still present in
the final complex $\domain_n = \domain$).

\begin{figure}
\includegraphics[width=\linewidth]{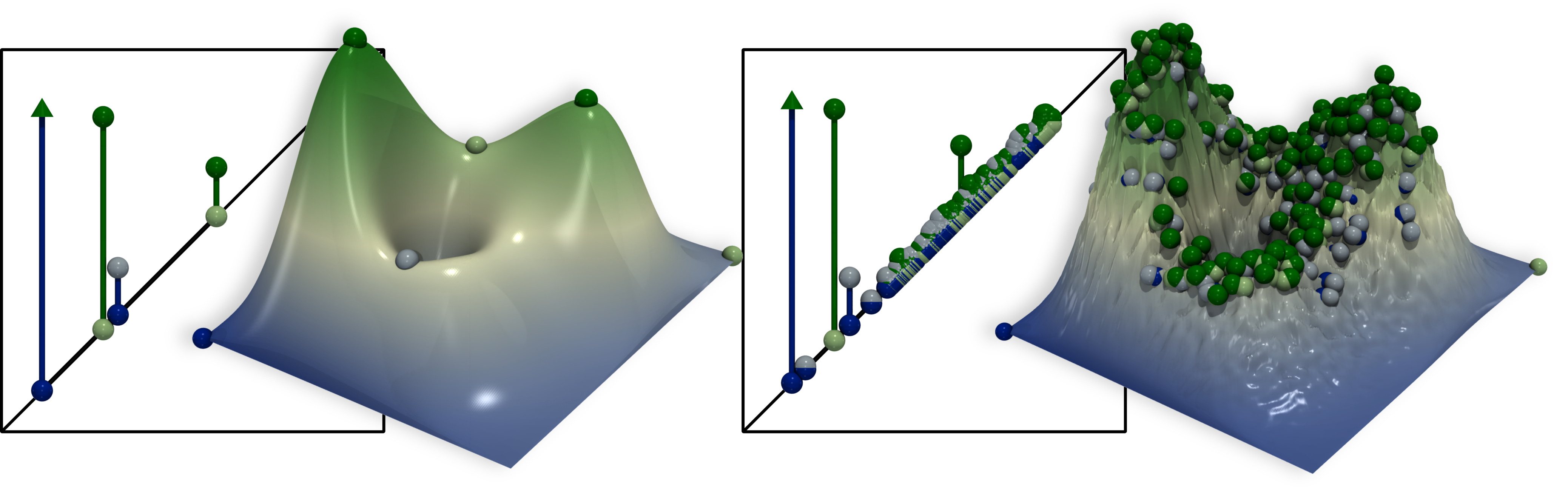}
 \mycaption{Persistence diagrams for the lexicographic filtration of a clean
(left) and noisy (right) version the terrain toy example (blue bars:
$\diagram_0(f)$, green bars: $\diagram_1(f)$).
 Classes with infinite persistence are shown with an upward arrow. Critical simplices are reported with spheres (dark blue: minima, dark green: maxima, other: saddles).
 The Betti numbers of any step $i$ of the filtration can be directly read from the diagram, by counting the intersections between a horizontal line at height $i$ with the bars of the diagram. The \emph{persistence} of each topological feature is given by the height of each bar. In practice, noise in data tends to create short bars (right) which can be easily distinguished from the main signal (long bars), in this case, two prominent hills and one salient pit.}
 \label{fig_diagram}
\end{figure}

The persistence diagram of dimension $p$, noted $\diagram_p(f)$, is a concise
encoding of the $p$-dimensional
persistent homology groups. In particular, it embeds each persistent generator
$\gamma$ in the 2D birth/death plane at position
$\big(f(v_i), f(v_j)\big)$ and its persistence can be therefore directly read 
from its height to the diagonal. This has the practical implication that 
generators
with large persistence (typically corresponding to salient features in
the data) are located far away from the diagonal, whereas
generators
with small
persistence (typically corresponding to noise) are located in the vicinity of 
the diagonal, as illustrated in \autoref{fig_diagram}.

\begin{figure}
\includegraphics[width=\linewidth]{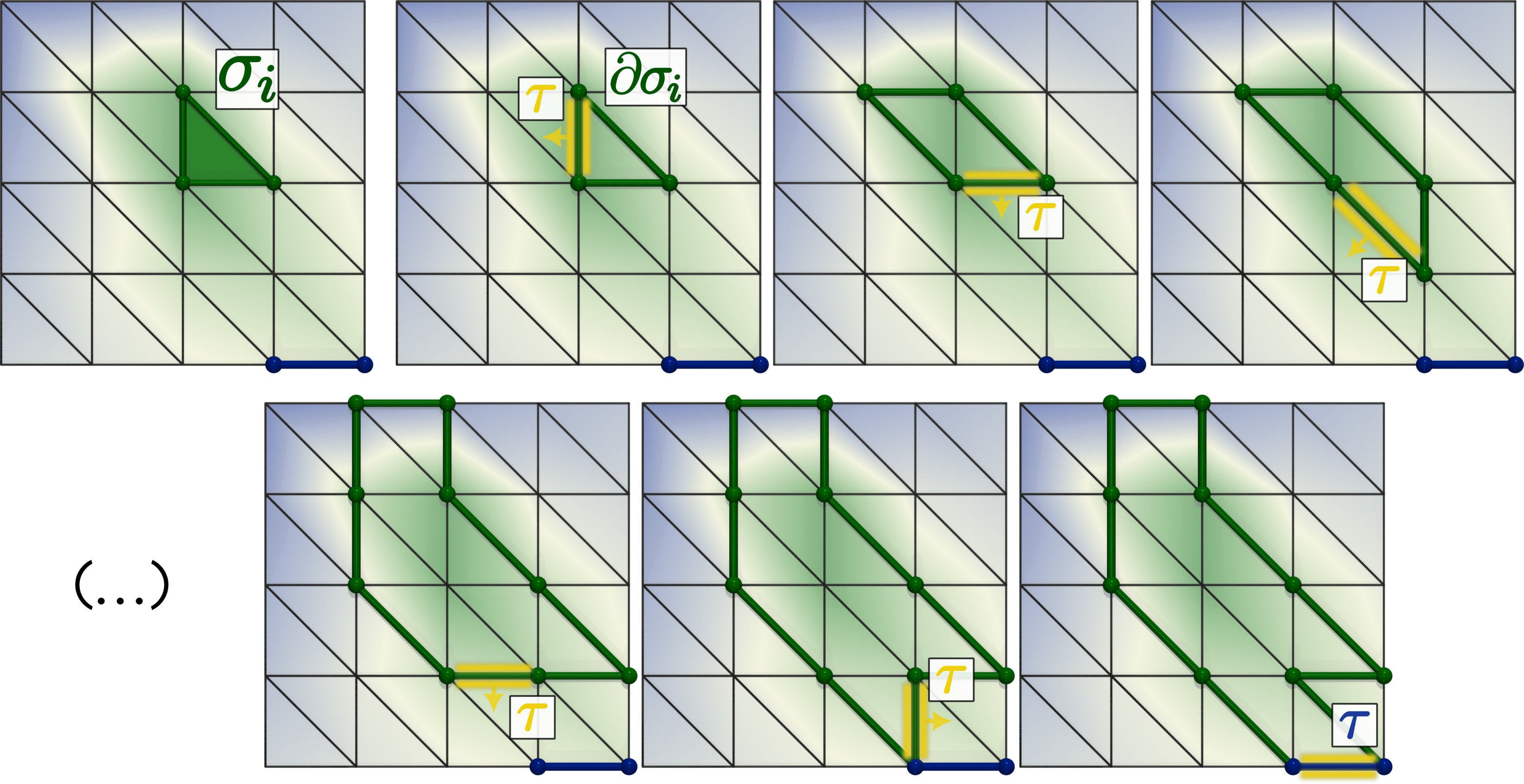}
 \mycaption{\emph{Homologous \majorRevision{propagation}}
 given a $(\dimensionality_i + 1)$-simplex $\simplex_i$ (left to right, top to bottom). The boundary $\partial \simplex_i$ (green curve) is iteratively expanded by:
 \emph{(i)} selecting its highest $(\dimensionality_i)$-simplex $\tau$ (yellow, line \ref{algo_pairCells_selectSimplex}, \autoref{algo_pairCells}) and
 \emph{(ii)} adding the boundary of the expanded $(\dimensionality_i + 1)$-chain
paired with $\tau$ (line \ref{algo_pairCells_growingChain},
\autoref{algo_pairCells}), here the triangle adjacent to $\tau$.
 The
 \majorRevision{propagation}
 stops (line \ref{algo_pairCells_foundCycle},
\autoref{algo_pairCells}) when an unpaired $(\dimensionality_i)$-simplex (blue
edge) is included in the expanded boundary (bottom right). Then, a persistence
pair $(\tau, \simplex_i)$ is created if the expanded boundary is not empty (line
\ref{algo_pairCells_addPair}, \autoref{algo_pairCells}).
 }
 \label{fig_seminalPairCell}
\end{figure}

\begin{algorithm}
    \small
    \algsetup{linenosize=\tiny}
    \caption{\textcolor{black}{\footnotesize Reference
\majorRevision{\emph{``PairSimplices''}
\cite{edelsbrunner02, zomorodianBook}}}}
\label{algo_pairCells}
\hspace*{\algorithmicindent} \textbf{Input}: Lexicographic filtration of
$\domain$ by $f$.

\hspace*{\algorithmicindent} \textbf{Output}: Persistence diagrams
$\diagram_0(f)$, $\diagram_1(f)$ and $\diagram_2(f)$.

\begin{algorithmic}[1]

\FOR{$j \in [1, n]$}
  \STATE // Process the $(\dimensionality_i+1)$-simplex $\sigma_j$
  \STATE $Pair(\sigma_j) \leftarrow \emptyset$
  \STATE $Chain(\sigma_j) \leftarrow \sigma_j$
  \STATE // Homologous \majorRevision{propagation}
  of $\partial \sigma_j$
  \WHILE{$\partial\big(Chain(\sigma_j)\big) \neq 0$}
  \label{algo_pairCells_startExtension}
    \STATE $\tau \leftarrow \max\Big(\partial\big(Chain(\sigma_j)\big)\Big)$
    \label{algo_pairCells_selectSimplex}
    \IF{$Pair(\tau) == \emptyset$}
      \STATE // $\tau$ created a $(d_i)$-cycle
      \STATE \textbf{break}
      \label{algo_pairCells_foundCycle}
    \ELSE
      \STATE // Expand chain (with homologous boundary)
      \STATE $Chain(\sigma_j) \leftarrow Chain(\sigma_j) +
Chain\big(Pair(\tau)\big)$
      \label{algo_pairCells_growingChain}
    \ENDIF
  \ENDWHILE
  \label{algo_pairCells_endExtension}
  \IF{$\partial \big(Chain(\sigma_j)\big) \neq 0$}
    \STATE // A
    non-trivial cycle homologous to
$\partial
\sigma_j$ exists (l. \autoref{algo_pairCells_foundCycle})
    \STATE $\tau \leftarrow \max\Big(\partial\big(Chain(\sigma_j)\big)\Big)$
    \STATE $Pair(\sigma_j) \leftarrow \tau$
    \STATE $Pair(\tau) \leftarrow \sigma_j$
    \STATE $\diagram_{\dimensionality_{i}}(f) \leftarrow
\diagram_{\dimensionality_{i}}(f) \cup (\tau, \sigma_j)$
    \label{algo_pairCells_addPair}
  \ENDIF
\ENDFOR
\end{algorithmic}
\end{algorithm}

\begin{figure}
  \includegraphics[width=\linewidth]{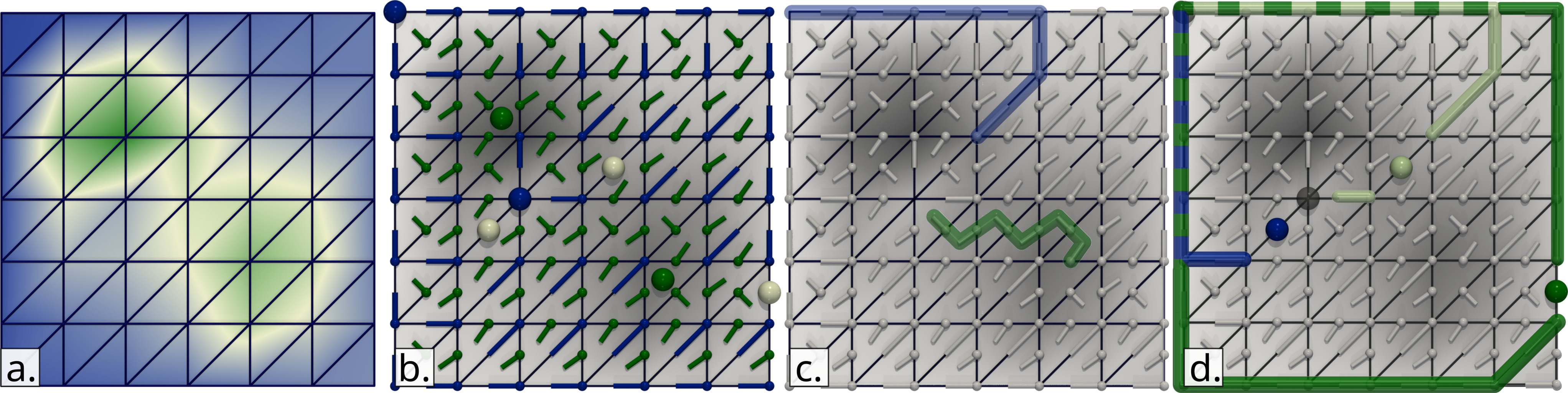}
 \mycaption{Given an input PL scalar field $f$ \emph{(a)}, a \emph{discrete
gradient field} $\discreteGradientField$
 pairs $i$-simplices with $(i+1)$-simplices
 \emph{(b)} (blue: vertex-edge arrows, green: edge-triangle arrows). The
remaining unpaired simplices are
 \emph{critical}
 (large spheres, blue: minima, white: saddles, green: maxima).
 Two \emph{discrete integral lines} (or \emph{``v-paths''}) are shown in
\emph{(c)} (blue: vertex-edge integral line, green: edge-triangle integral
line).
 The \emph{discrete unstable set} of each saddle, \emph{(d)}, is the collection
of all discrete integral lines starting from it (each discrete unstable set is
colored according to its saddle).
%
%
%
 }
 \label{fig_discreteMorseTheory}
\end{figure}

\subsection{The algorithm \majorRevision{\emph{``PairSimplices''}}}
\label{sec_pairCells}
\majorRevision{Edelsbrunner et al.
\cite{edelsbrunner02, edelsbrunner09, zomorodianBook} describe} an
iterative algorithm called
\majorRevision{\emph{``PairSimplices''}} for the computation of persistence
diagrams. We sketch
its main steps here as our approach builds on top of it.
\majorRevision{This description is adapted from Zomorodian's textbook
\cite{zomorodianBook}.}
This algorithm (\autoref{algo_pairCells}) observes, for each step $i$ of the
input filtration, the effect of the insertion of a $\dimensionality_i$-simplex
$\sigma_i$ on the set of $(\dimensionality_i-1)$-cycles homologous to its
boundary $\partial \sigma_i$.
In particular,
if $\partial \sigma_i$ was not already trivial (i.e. homologous to an empty
cycle) in $\domain_{i-1}$, then the insertion of
$\sigma_i$ in
$\domain_i$ will now make $ \partial \sigma_i$
trivial ($\partial
\sigma_i \sim 0$).
By transitivity, all the
cycles $c$ homologous to $\partial \sigma_i$ (its homology class) now become
trivial as well,
hence
completing
a
persistence pair
in $\diagram_{\dimensionality_{i}-1}(f)$,
that is, \emph{filling} a $\dimensionality_{i-1}$-dimensional hole of
$\domain_{i-1}$.

Thus, for each step $i$ of the filtration, the algorithm reconstructs
$(\dimensionality_i-1)$-cycles in $\domain_i$ which are homologous to
$\partial \sigma_i$. This is achieved by
a process that we call \emph{homologous
\majorRevision{propagation}}
(\autoref{fig_seminalPairCell}), which
iteratively expands a chain
$Chain(\sigma_i)$, whose boundary $\partial \big(Chain(\sigma_i)\big)$ is
homologous to $\partial \sigma_i$ by construction (\autoref{algo_pairCells},
lines \autoref{algo_pairCells_startExtension}  to
\autoref{algo_pairCells_endExtension}).
This
\majorRevision{propagation}
is achieved by considering $(\dimensionality_i-1)$-simplices
in decreasing filtration order, i.e. by selecting at each iteration the
highest
simplex (\autoref{algo_pairCells}, line
\autoref{algo_pairCells_selectSimplex}), and by stopping at the first unpaired
$(\dimensionality_{i}-1)$-simplex $\tau$ (\autoref{algo_pairCells}, line
\autoref{algo_pairCells_foundCycle}), responsible for the creation of the latest
(i.e. youngest) homologous $(\dimensionality_{i}-1)$-cycle in $\domain_{i-1}$,
hence effectively enforcing the
\emph{Elder rule} (\autoref{sec_persistenceDiagram}). Then the persistence pair
$(\tau, \sigma_i)$ is created in $\diagram_{d_i-1}(f)$.
If $\partial
\sigma_i$ was trivial initially in $\domain_{i-1}$ when starting the
\majorRevision{propagation},
$Chain(\sigma_i)$ is extended until its boundary becomes empty and no pair
will be created in $\diagram_{d_i-1}(f)$.
\majorRevision{Note that the above algorithm can be viewed as a
geometric interpretation of boundary matrix reduction, which is at the core of
most modern approaches (\autoref{sec_relatedWork}). In particular, selecting
the highest boundary simplex $\tau$ (\autoref{algo_pairCells}, line
\autoref{algo_pairCells_selectSimplex}) is equivalent to the identification,
during  matrix reduction, of
the lowest
non-zero entry for the column $j$ of the boundary matrix\minorRevision{.
Considering}
the
boundary of the chain sum (line
\autoref{algo_pairCells_growingChain}) is equivalent to summing columns
\cite{edelsbrunner09} (Sec. VII.1).
}


\subsection{Discrete Morse Theory (DMT)}
\label{sec_discrete_morse_theory}

We now conclude this section of preliminaries with notions of discrete Morse
theory \cite{forman98}, or DMT for short (which we restrict here to
simplicial complexes), as it is instrumental in our approach to accelerate the
algorithm
\majorRevision{\emph{``PairSimplices''}}.

We call a discrete vector a pair formed by a simplex $\simplex_i 
\in \domain$ (of dimension $i$) and one of its co-facets $\simplex_{i+1}$ 
(i.e. one of its co-faces of dimension $i+1$), noted $\{\simplex_i < 
\simplex_{i+1}\}$. 
$\simplex_{i+1}$ is usually referred to as the \emph{head} of the vector, while 
$\simplex_i$ is its tail.
Examples of discrete vectors include a pair between a vertex 
and one of its incident edges, or a pair between an edge and
a triangle containing it
(see \autoref{fig_discreteMorseTheory}). A \emph{discrete vector field} on
$\domain$ is then
defined as a collection $\discreteVectorField$ of pairs $\{\simplex_i < 
\simplex_{i+1}\}$ such that each simplex of $\domain$ is involved in at most 
one pair. A simplex $\simplex_i$ which is involved in no discrete vector of 
$\discreteVectorField$ is called a \emph{critical simplex}. 

A discrete integral line, or \emph{v-path}, is a sequence of discrete vectors 
$\big\{\{\simplex_i^0 < \simplex_{i+1}^0\}, \dots, \big\{\{\simplex_i^k < 
\simplex_{i+1}^k\} \big\}$ 
such that \emph{(i)} $\simplex_i^j \neq \simplex_i^{j+1}$ (i.e. the tails of 
two consecutive vectors are distinct) and \emph{(ii)} $\simplex_i^{j+1} < 
\simplex_{i+1}^j$ (the tail of a vector in the sequence is a face of the head 
of the previous vector in the sequence) for any $0 < j < k$.
We say that a discrete integral line \emph{terminates} at a critical simplex 
$\simplex_i$ if $\simplex_i$ is a face\majorRevision{t} of the head of its last 
vector
$\{\simplex_i^k < 
\simplex_{i+1}^k\}$ (i.e. $\sigma_i < \simplex_{i+1}^k$). 
Symmetrically, we say that a discrete integral line \emph{starts} at a critical 
simplex $\sigma_{i+1}$ if $\sigma_{i+1}$ \majorRevision{is} a 
co-face\majorRevision{t} of
the tail of its first vector $\sigma_i^0$ (i.e. $\sigma_i^0 < \sigma_{i+1}$).
By analogy with the 
smooth setting, this notion of discrete integral lines therefore starts and 
terminates at  critical points.
The collection of all the discrete integral
lines terminating in a given critical simplex $\sigma_i$ is called the 
\emph{discrete stable set} of $\sigma_i$ and it is noted 
$\stableManifold(\sigma_i)$. Symmetrically, the collection of all the discrete 
integral lines starting at a given critical simplex $\sigma_i$ is called the 
\emph{discrete unstable set} of $\sigma_i$  (\autoref{fig_discreteMorseTheory}) and it is noted
$\unstableManifold(\sigma_i)$.

A discrete vector field
such that all of its possible discrete integral
lines
are acyclic is called a \emph{discrete gradient
field} \cite{forman98}, noted $\discreteGradientField$.
Then, the critical simplices of $\discreteGradientField$ are discrete analogs
to the
critical points from the smooth setting \cite{morseQuote, milnor63}. Their
dimension $i$ corresponds to the
smooth notion of
index
(number of negative
eigenvalues of the Hessian): local minima
occur on
vertices,
$i$-saddles on $i$-simplices and maxima
on $d$-simplices.
\majorRevision{For}
\julien{typical scalar}
data, the input is
generically provided as a PL scalar field $f$
\majorRevision{(\autoref{sec_inputData})}.
Given this input,
Robins et al. introduced
an algorithm based on
expansions \cite{robins_pami11} \majorRevision{(in the sense of simple homotopy
theory)}, which
guarantees that each resulting critical $d_i$-simplex $\sigma_i$ belongs to
the lower star of a PL critical point of index $d_i$.

%
%

\begin{figure*}
  \centering
  \includegraphics[width=\linewidth]{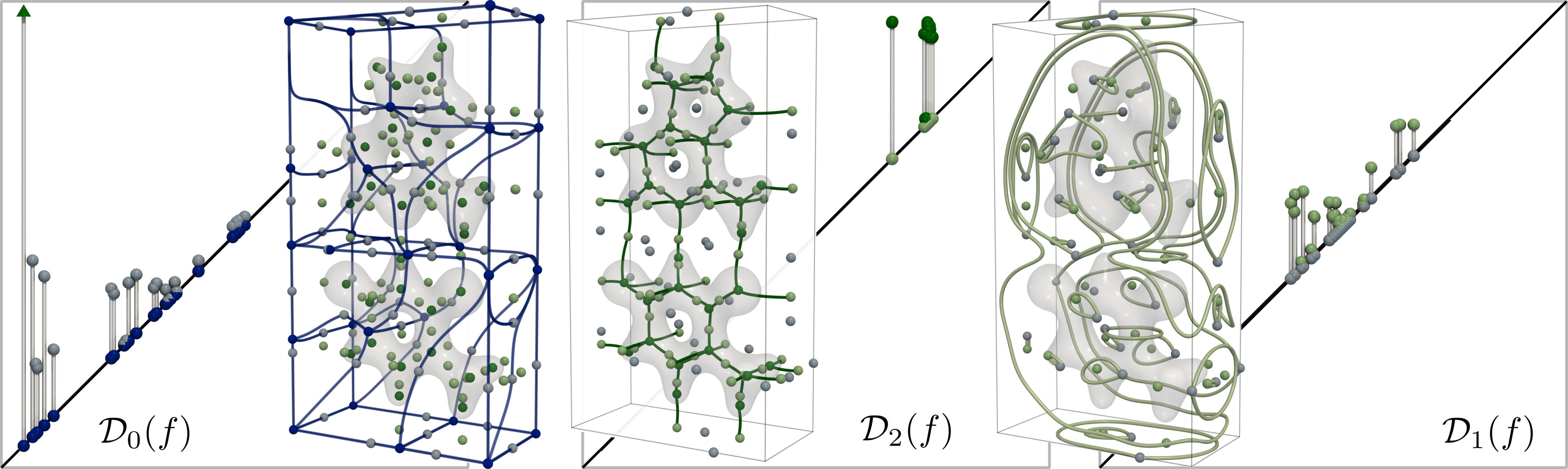}
  \mycaption{Overview of our approach on a quantum chemistry dataset (grey
surface: electron density
level set).
  The diagram $\diagram_0(f)$ is first efficiently computed (left) by processing
the unstable sets (blue curves) of the $1$-saddles (light blue spheres) of $f$
(\autoref{sec_diagram0}). Simultaneously (center), $\diagram_{\dimensionality -
1}(f)$ is computed symmetrically, by processing the stable sets (green curves)
of the $(\dimensionality-1)$-saddles (light green spheres) of $f$
(\autoref{sec_maxDiagram}). Finally (right), only the remaining, unpaired
$2$-saddles (light green spheres) are considered by our
algorithm
\emph{``PairCriticalSimplices''} (\autoref{sec_saddleSaddle}) to efficiently
compute $\diagram_1(f)$ by homologous
\majorRevision{propagation}
of their associated $1$-cycles
(\majorRevision{light green}
curves). This stratification strategy
  drastically reduces the number of unpaired critical simplices at each step,
  leaving only a small portion for the last and most computationally expensive
part of our approach
  (right).}
%
%
%
  \label{fig_overview}
\end{figure*}

\section{Overview}
\label{sec_overview}




Figure \ref{fig_overview} provides an overview of our approach. We assume that
$\domain$ is connected (otherwise, each connected component is processed
independently by our algorithm). While we focus on simplicial complexes in our work (\autoref{sec:introduction}), our algorithm can be applied in principle to arbitrary cell complexes.


First, the discrete gradient of the input data is computed along with its
critical simplices via
\majorRevision{expansions} \cite{robins_pami11}.
The
rest
of
our approach consists in grouping the resulting critical simplices
into persistence
pairs.
We describe for that
\majorRevision{an}
extension (\autoref{sec_saddleSaddle}) of
the
algorithm
\majorRevision{\emph{``PairSimplices''} \cite{edelsbrunner02, zomorodianBook}},
which is expressed in
the DMT setting for improved performances, and that we
call \emph{``PairCriticalSimplices''}
\majorRevision{(\autoref{algo_pairCriticalSimplices})}. While each diagram
$\diagram_0(f)$,
$\diagram_1(f)$ and $\diagram_2(f)$ could be computed
with this
algorithm, we describe instead a stratification strategy,
called
\emph{sandwiching},
(described below),
which further
improves
performances.

Second (\autoref{fig_overview}, left),
the
diagram
$\diagram_0(f)$ is obtained  by processing the
unstable sets of the
$1$-saddles of $f$ (\autoref{sec_diagram0}).

Third (\autoref{fig_overview}, center) the diagram
$\diagram_{\dimensionality -1}(f)$ is obtained   by
processing the stable sets of the ($\dimensionality - 1)$-saddles of $f$
(\autoref{sec_maxDiagram}).

Next (\autoref{fig_overview}, right) for 3D data only, the diagram
$\diagram_{1}(f)$ is computed by restricting our novel algorithm
\emph{``PairCriticalSimplices''} (\autoref{sec_saddleSaddle}) to the remaining
set of
unpaired $2$-saddles.


Last, the remaining critical simplices are necessarily involved in classes of
infinite persistence,
capturing the
(infinitely persistent)
homology groups of
$\domain$ (\autoref{sec_infinitePersistence}).

\section{Pairing Critical Simplices}
\label{sec_saddleSaddle}

This section presents our adaptation of the seminal algorithm 
\majorRevision{\emph{``PairSimplices''}}
(\autoref{sec_pairCells}) to the
DMT
setting
(\autoref{sec_discrete_morse_theory}), resulting in substantial performance
gains.

\subsection{Observations}
\label{sec_observations}
This section describes three main observations regarding the algorithm
\majorRevision{\emph{``PairSimplices''}} (\autoref{algo_pairCells}), which
are at the basis of our 
adaptation, described in \autoref{sec_algorithmPairCriticalSimplices}.

\noindent
\textbf{\emph{(a)} Dimension separability} First, one can observe that the different
persistence diagrams $\diagram_0(f)$, $\diagram_1(f)$ and $\diagram_2(f)$ can
be computed in a separated manner, one after the other.
Indeed, a given $d_{i}$-simplex $\simplex_i$ can only be involved in
\emph{(i)}
the destruction of a
$(d_i - 1)$-cycle (if $\partial \simplex_i$ was not trivial), or
\emph{(ii)}
the creation of a $d_i$-cycle (if $\partial \simplex_i$ was trivial). For instance, the addition
of a $1$-simplex in the lexicographic filtration  connects two
vertices belonging either \emph{(i)} to distinct connected components (in which
case a persistence pair is added to $\diagram_0(f)$, line
\autoref{algo_pairCells_addPair}, \autoref{algo_pairCells}) or \emph{(ii)} to
the same connected component (in which case a new $1$-cycle is created
line \autoref{algo_pairCells_foundCycle}, \autoref{algo_pairCells}, to be
later added to $\diagram_1(f)$). Thus, if the persistence diagram
$\diagram_{i-1}(f)$ is available, the diagram $\diagram_{i}(f)$ can be
efficiently computed by restricting \autoref{algo_pairCells} to the $i+1$
simplices of $\domain$ (still processed in lexicographic order). Then, each
$i$-simplex which has not been paired yet in $\diagram_{i-1}(f)$ will be
guaranteed to be the creator of a $i$-cycle, and thus involved in a persistence
pair of $\diagram_{i}(f)$. This dimension separability is at the
basis
of
our \emph{sandwiching} stratification strategy.
\majorRevision{This observation is not novel. It is also at the basis of
previous approaches, e.g. the \emph{clearing} and \emph{compression}
acceleration \cite{bauer13}.}

\noindent
\textbf{\emph{(b)} Boundary caching}
\majorRevision{As it is described in Zomorodian's textbook
\cite{zomorodianBook}, the}
algorithm
\majorRevision{\emph{``PairSimplices''}} proceeds
to the homologous
\majorRevision{propagation}
of $\dimensionality_i$-cycles by iteratively growing
$(\dimensionality_i + 1)$-chains (line \autoref{algo_pairCells_growingChain},
\autoref{algo_pairCells}), and by explicitly extracting their boundary when
needed
(\minorRevision{e.g.}
line \autoref{algo_pairCells_startExtension},
\autoref{algo_pairCells}). However, each of these extractions requires a pass
which is linear with the size of the chain. This can be improved
\minorRevision{(as described in
\autoref{sec_algorithmPairCriticalSimplices})}
by \emph{caching} the boundary of the chain created at each simplex, and by
manipulating boundaries directly in the
\majorRevision{propagation}
process, instead of
\minorRevision{manipulating}
chains \minorRevision{(similarly to approaches based on boundary
matrix
reduction)}.
This adaptation
of the
\majorRevision{propagation}
process
still require\minorRevision{s} a linear pass
(\minorRevision{for}
the
modulo-$2$ addition of boundary simplices), but this time on a much smaller set
(boundaries are
in practice
much smaller than their
chains).

\begin{figure}
  \includegraphics[width=\linewidth]{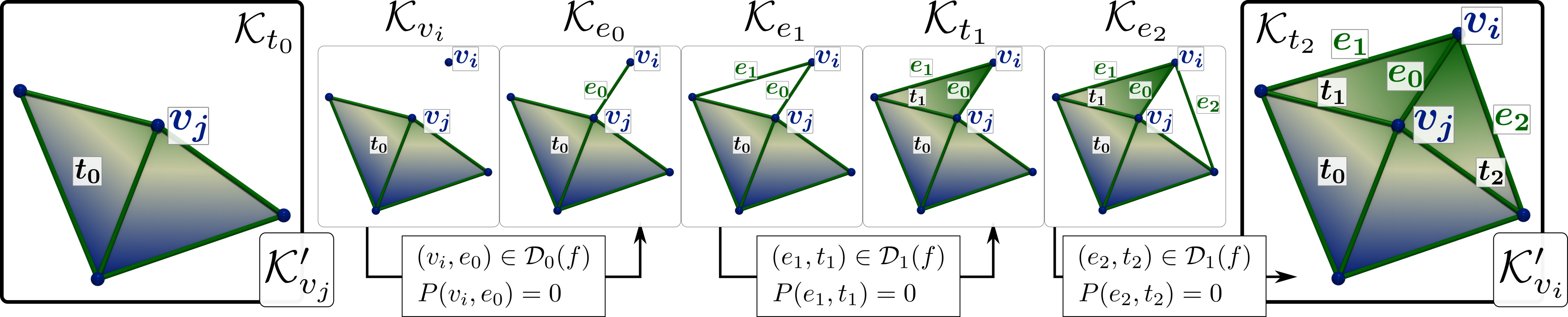}
 \mycaption{The \emph{lexicographic} filtration introduces simplices one by one
(left to right) and
 can capture many zero-persistence pairs ($(v_i, e_0), (e_1, t_1)$ and $(e_2, t_2)$), which are not captured by the \emph{lower-star} filtration, which introduces simplices by chunks of lower stars (steps $\domain'_{v_j}$ and $\domain'_{v_i}$).}
%
  \label{fig_zeroPersistencePairs}
\end{figure}

\noindent
\textbf{\emph{(c)} Zero\majorRevision{-}persistence skip}
By definition of a filtration (\autoref{sec_lowerStarFiltration}), a given
simplex cannot be inserted in the filtration before its
facets.
A practical implication of this observation is that many,
zero-persistence pairs are created by the
algorithm
\majorRevision{\emph{``PairSimplices''}}.
For
instance, the insertion of the last edge $e_1$ of a given triangle
$t_1$
often creates a new $1$-cycle
(step $\domain_{e_1}$, \autoref{fig_zeroPersistencePairs})
which is immediately filled by the
subsequent insertion of $t_1$ (step $\domain_{t_1}$, \autoref{fig_zeroPersistencePairs}), creating a persistence pair $(e_1, t_1)$ in $\diagram_1(f)$.
However, since the persistence of a pair is
given by the difference between the maximum vertex data values of $e_1$
and $t_1$ (\autoref{sec_persistenceDiagram}), we have in such cases
$\persistence(e_1, t_1) = 0$, as $e_1$  and $t_1$
have to share the highest vertex of $t_1$, given the lexicographic order.
Thus, a significant time
(\majorRevision{c.f.}
\autoref{sec_perfAnalysis}) is spent in
practice by
the 
algorithm
\majorRevision{\emph{``PairSimplices''}}
to construct persistence pairs with zero\majorRevision{-}persistence, which
consequently do not contribute any information
to
the output diagram.
We address this issue with
DMT.
In
particular, the discrete gradient $\discreteGradientField$ computed via
\majorRevision{expansions} \cite{robins_pami11} guarantees that
each critical simplex
belongs to
the lower star of a PL critical point of $f$,
which exactly coincide themselves to
changes in the
homology
groups of the lower star filtration (\autoref{sec_discrete_morse_theory}). Then,
all the remaining regular simplices (involved in a discrete vector of
$\discreteGradientField$) induce homology changes which are \emph{not}
captured by the lower star filtration (steps marked with a black frame in
\autoref{fig_zeroPersistencePairs}, $\domain'_{v_j}$ and $\domain'_{v_i}$), and
which, equivalently, have
zero\majorRevision{-}persistence. Then, it follows that all the zero-persistence
pairs of the
lexicographic filtration
can be efficiently skipped in a
pre-process, by
discarding from the computation all the simplices which are not critical, as
only the critical simplices will induce non-zero persistence homology changes
(i.e., captured by the lower star filtration,
\autoref{sec_discrete_morse_theory}).
\subsection{Algorithm}
\label{sec_algorithmPairCriticalSimplices}

Our adaptation of the algorithm
\majorRevision{\emph{``PairSimplices''}}
to the
DMT
setting, called \emph{``PairCriticalSimplices''}, directly results from
the above observations.

\begin{algorithm}[t]
    \small
    \algsetup{linenosize=\tiny}
    \caption{\textcolor{black}{\footnotesize
    Our pre-processing algorithm
    \emph{``Zero\majorRevision{-}persistence skip''}}}
    \label{algo_skippingZeroPersistence}
\hspace*{\algorithmicindent} \textbf{Input}: Lexicographic filtration of
$\domain$ by $f$.

\hspace*{\algorithmicindent} \textbf{Output}: Ordered sets
$C_{\dimensionality_i}$ or critical $\dimensionality_i$-simplices.

\begin{algorithmic}[1]

\STATE // Discrete gradient (\majorRevision{via expansions}
\cite{robins_pami11})
\STATE $\discreteGradientField \leftarrow DiscreteGradient(\domain, f)$
\label{algo_zeroPersistenceSkip_discreteGradient}

\FOR{$i \in [1, n]$}
    \STATE // Process a $\dimensionality_i$-simplex $\simplex_i$
    \IF{$\exists ~ \{\simplex_i \lt \simplex_j \}$ or $\exists ~ \{\simplex_j
\lt \simplex_i \}$}
      \STATE // $\simplex_i$ is involved in a discrete vector with $\simplex_j$
        \STATE // Skip zero-persistence pair
        \STATE $Pair(\simplex_i) \leftarrow \simplex_j$
        \STATE $Pair(\simplex_j) \leftarrow \simplex_i$
        \label{algo_zeroPersistenceSkip_zeroPair}
    \ELSE
      \STATE // $\simplex_i$ is a critical simplex
      \STATE $Pair(\simplex_i) \leftarrow \emptyset$
      \label{algo_zeroPersistenceSkip_critical}
      \STATE $C_{\dimensionality_i} \leftarrow C_{\dimensionality_i} \cup
\simplex_i$
      \label{algo_zeroPersistenceSkip_criticalSet}
    \ENDIF
\ENDFOR

\FOR{$\dimensionality_i \in [0, d]$}
  \STATE $Sort(C_{\dimensionality_i})$ $\quad$ // by lexicographic order
  \label{algo_zeroPersistenceSkip_sort}
\ENDFOR
\end{algorithmic}
\end{algorithm}

\begin{algorithm}[t]
    \small
    \algsetup{linenosize=\tiny}
    \caption{\textcolor{black}{\footnotesize
    Our algorithm \emph{``PairCriticalSimplices''}.}}
    \label{algo_pairCriticalSimplices}
\hspace*{\algorithmicindent} \textbf{Input}: Ordered set $C_{\dimensionality_i +
1}$ of critical $(\dimensionality_i+1)$-simplices

\hspace*{\algorithmicindent} \textbf{Output}: Persistence diagrams
$\diagram_{\dimensionality_i}(f)$.

\begin{algorithmic}[1]

\FOR{$j \in C_{\dimensionality_i + 1}$}
  \label{algo_pairCriticalSimplices_firstFor}
  \STATE // Process the $(\dimensionality_i + 1)$-simplex $\simplex_j$
  \STATE $Boundary(\sigma_j) \leftarrow \partial \simplex_j$
  \STATE // Homologous \majorRevision{propagation}
  of $\partial \sigma_j$
  \WHILE{$Boundary(\sigma_j) \neq 0$}
  \label{algo_pairCriticalSimplices_startExtension}
    \STATE $\tau \leftarrow \max\big(Boundary(\sigma_j)\big)$
    \label{algo_pairCricitalSimplices_selectSimplex}
    \IF{$Pair(\tau) == \emptyset$}
      \label{algo_pairCriticalSimplices_foundUnpaired}
      \STATE // $\tau$ is unpaired and thus created a $d_i$-cycle.
      \STATE \textbf{break}
      \label{algo_pairCriticalSimplices_foundCycle}
    \ELSE
      \STATE // Expand boundary
      \STATE $Boundary(\sigma_j) \leftarrow $
      \label{algo_pairCriticalSimplices_expansion}
      \STATE $\quad \quad Boundary(\sigma_j) + Boundary\big(Pair(\tau)\big)$
      \label{algo_pairCriticalSimplices_mod2}
    \ENDIF
  \ENDWHILE
  \label{algo_pairCriticalSimplices_endExtension}
  \IF{$Boundary(\sigma_j) \neq 0$}
    \label{algo_pairCriticalSimplices_nonEmpty}
    \STATE // A
    non-trivial cycle homologous to
$\partial
\sigma_j$ exists (l. \autoref{algo_pairCriticalSimplices_foundCycle})
    \STATE $\tau \leftarrow \max\big(Boundary(\sigma_j)\big)$
    \STATE $Pair(\sigma_j) \leftarrow \tau$
    \STATE $Pair(\tau) \leftarrow \sigma_j$
    \STATE $\diagram_{\dimensionality_{i}}(f) \leftarrow
\diagram_{\dimensionality_{i}}(f) \cup (\tau, \sigma_j)$
    \label{algo_pairCriticalSimplices_pairCreated}
  \ENDIF
\ENDFOR
\end{algorithmic}
\end{algorithm}

\noindent
\textbf{\emph{(a)} Zero-persistence skip} First, \autoref{algo_skippingZeroPersistence} is
used in a pre-process to skip the zero-persistence pairs of the
lexicographic filtration. This algorithm first computes the discrete gradient
field $\discreteGradientField$ given the input PL scalar field $f : \domain
\rightarrow \mathbb{R}$ with
\majorRevision{expansions} \cite{robins_pami11} (line
\autoref{algo_zeroPersistenceSkip_discreteGradient}).
\majorRevision{At this stage, for any  discrete vector $\{\simplex_i,
\simplex_{i+1}\} \in \discreteGradientField$,
it
is guaranteed that both $\simplex_i$ and $\simplex_{i+1}$ are involved in
zero-persistence pairs (\autoref{sec_observations}). Thus, by convention, we
pair $\simplex_i$ and $\simplex_{i+1}$ together}
(line
\autoref{algo_zeroPersistenceSkip_zeroPair}). Otherwise, critical
$\dimensionality_i$-simplices are marked as unpaired (line
\autoref{algo_zeroPersistenceSkip_critical}) and are added to the set
$C_{\dimensionality_i}$ of critical
$\dimensionality_i$-simplices (line
\autoref{algo_zeroPersistenceSkip_criticalSet}). Once all discrete vectors have
been processed, each set $C_{\dimensionality_i}$
is sorted by increasing lexicographic order (line
\autoref{algo_zeroPersistenceSkip_sort}).

\noindent
\textbf{\emph{(b)} Pair Critical Simplices} We now present our algorithm
\emph{``PairCriticalSimplices''} (\autoref{algo_pairCriticalSimplices}). For a
given simplex dimension $\dimensionality_i + 1$, this algorithm takes as an
input the ordered set $C_{\dimensionality_i + 1}$ of critical
$(\dimensionality_i + 1)$-simplices and produces the diagram
$\diagram_{\dimensionality_i}(f)$. This assumes that the diagram
$\diagram_{\dimensionality_i-1}(f)$ has already been computed (see the
\emph{Dimension separability} property, \autoref{sec_observations}) and that
consequently, the critical $\dimensionality_i$-simplices involved in
$\diagram_{\dimensionality_i-1}(f)$ have already been paired.
Since all the regular simplices inserted in between two critical simplices by
the lexicographic filtration are guaranteed to belong to zero-persistence pairs
(\emph{Zero\majorRevision{-}persistence skip} property,
\autoref{sec_observations}),
\autoref{algo_pairCriticalSimplices} simply processes the critical simplices of
$C_{\dimensionality_i + 1}$ in increasing lexicographic order. For each critical simplex
$\simplex_j$, the standard, downwards homologous
\majorRevision{propagation}
of the classical
algorithm
\majorRevision{\emph{``PairSimplices''}}
is employed (line
\autoref{algo_pairCriticalSimplices_startExtension})\majorRevision{, possibly
visiting in the process some simplices $\tau$ (line
\autoref{algo_pairCricitalSimplices_selectSimplex}) which are not critical. In
that case, $Pair(\tau)$ will return the simplex with which $\tau$ forms a
discrete vector (c.f. \autoref{algo_skippingZeroPersistence})}.
%
\majorRevision{As} discussed in
\autoref{sec_observations} (\emph{Boundary caching} property), our algorithm
directly manipulates boundaries instead of the corresponding chains. Then, when
a boundary homologous to $\partial \simplex_j$ is expanded (line
\autoref{algo_pairCriticalSimplices_expansion}), a modulo-2 addition is
employed by manipulating a bit mask (indicating if a simplex $\simplex_i$ is
already preset in $Boundary(\simplex_j)$). The rest of the algorithm is
identical to the original algorithm
\majorRevision{\emph{``PairSimplices''}}:
if the expanded
boundary for the simplex $\simplex_j$ is not-empty (line
\autoref{algo_pairCriticalSimplices_nonEmpty}), this means that a critical
simplex $\tau$, creating a $\dimensionality_i$-cycle, has been found during the
downward homologous \majorRevision{propagation}
(line
\autoref{algo_pairCriticalSimplices_foundCycle}). Then, a persistence pair
$(\tau, \simplex_j)$ is created between $\simplex_j$ and the highest
$\dimensionality_i$-simplex $\tau$ of its expanded boundary
$Boundary(\simplex_j)$ (line \autoref{algo_pairCriticalSimplices_pairCreated}).
\majorRevision{Then, our algorithm \emph{``PairCriticalSimplices''} is a direct
geometric interpretation of boundary matrix reduction (as discussed for
\emph{``PairSimplices''}, \autoref{sec_pairCells}), but restricted to the
columns of the boundary matrix corresponding to critical simplices.}

\section{Extremum-Saddle Persistence Pairs}
\label{sec_extremumDiagrams}
Our algorithm \emph{``PairCriticalSimplices''}
(\autoref{sec_algorithmPairCriticalSimplices}) could be used as-is to compute
the diagrams $\diagram_0(f)$, $\diagram_1(f)$ and $\diagram_2(f)$ one after the
other, already resulting in substantial performance gains
over
the
seminal algorithm
\majorRevision{\emph{``PairSimplices''}}
(see \autoref{sec_perfAnalysis}). In
this section, we further exploit the \emph{Dimension separability} property
(\autoref{sec_saddleSaddle}) to further speedup the process.

%
%



\subsection{Minimum-Saddle Persistence Pairs}
\label{sec_diagram0}

This section introduces a faster alternative to the algorithm
\emph{``PairCriticalSimplices''}, for the specific case of $\diagram_0(f)$.

\begin{figure}
  \centering
  \includegraphics[width=\linewidth]{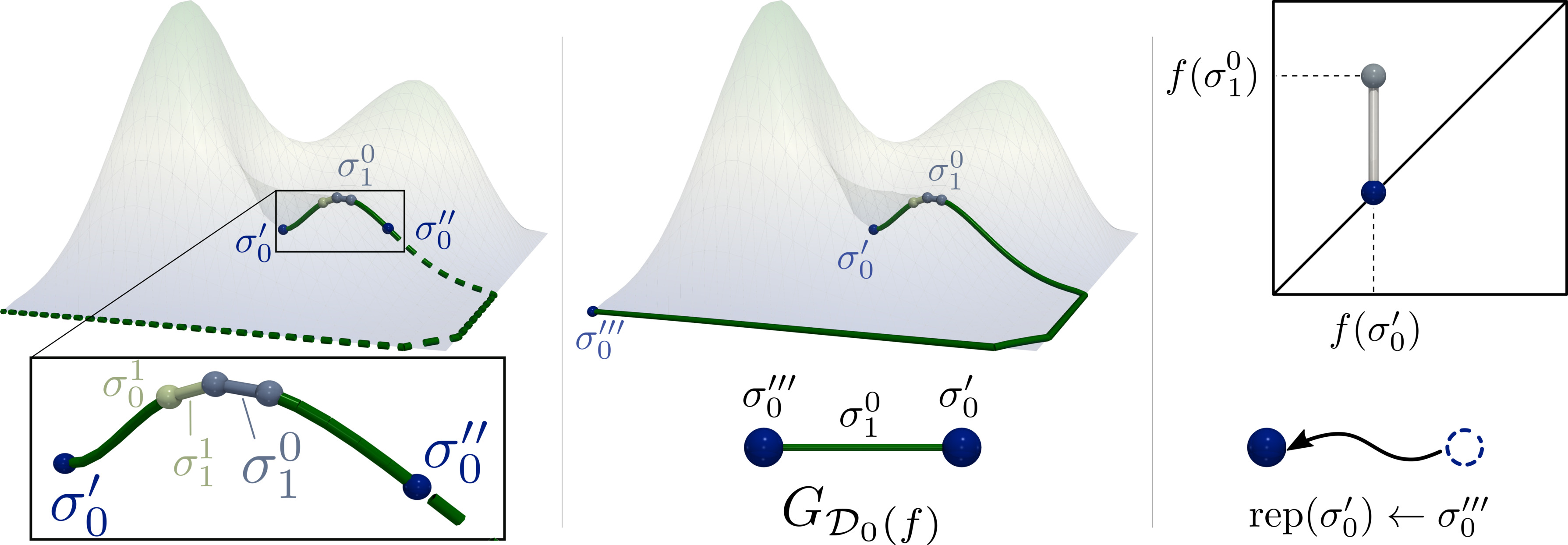}
 \mycaption{Overview of our algorithm for the computation of $\diagram_0(f)$.
 \emph{Left:} the homologous
 \majorRevision{propagation}
 of \autoref{algo_pairCriticalSimplices} applied to a critical edge
$\simplex_1^0$ turns out to be restricted to the discrete unstable sets of
$\simplex_1^0$, until it reaches a first unpaired vertex ($\simplex'_0$). There,
the
\majorRevision{propagation}
pauses and a persistence pair
 $(\simplex'_0, \simplex_1^0)$ is created. At this point, the expanded boundary $Boundary(\simplex_1^0)$ is equal to $\{\simplex'_0 + \simplex''_0\}$.
 \emph{Center:} It is possible that the expanded boundary of another critical
edge later hits the now paired vertex $\simplex'_0$, in which case it will
return the expanded boundary of its paired simplex ($\simplex_1^0$), that is
$Boundary(\simplex_1^0)$. This has the effect of \emph{resuming} the
 \majorRevision{propagation}
 from $\simplex''_0$, still on the discrete unstable set of
$\simplex_1^0$, down to $\simplex'''_0$. Our overall algorithm expedites this
homologous
\majorRevision{propagation}
by collapsing all regular simplices along the unstable sets
of critical edges, resulting in a graph $G_{\diagram_0(f)}$ (bottom), whose arcs
are the critical edges and the nodes the vertices at the end of their unstable
sets.
 \emph{Right:} The graph $G_{\diagram_0(f)}$ is processed with a Union-Find data structure by adding its arcs in increasing (original) lexicographic order and a new persistence pair is created in $\diagram_0(f)$ after the insertion of each arc.
  }
  \label{fig_overview_minSaddle}
\end{figure}

\noindent
\textbf{\emph{(a)} Unstable set restriction}
This algorithm is based on the key observation that, for the specific case of
$\diagram_0(f)$, given a critical $1$-simplex $\simplex_1^0$, the homologous
\majorRevision{propagation}
described in \autoref{algo_pairCriticalSimplices} exactly coincides
with the discrete unstable set (\autoref{sec_discrete_morse_theory}) of
$\simplex_1^0$ (see \autoref{fig_overview_minSaddle}). In particular, at the
first iteration of the algorithm, $\tau$
will be selected as one of the two vertices of $\simplex_1^0$, noted
$\simplex_0^0$. If $\simplex_0^0$ is not a minimum itself, it has to be paired
(given the discrete gradient $\discreteGradientField$,
\autoref{algo_skippingZeroPersistence}) with another edge $\simplex_1^1$, being
one of its co-facets, with $\simplex_1^1 \neq \simplex_1^0$. Since in simplicial
complexes, edges are guaranteed to connect distinct vertices, we then have the
property that the only other facet of $\simplex_1^1$ is another vertex
$\simplex_0^1 \neq \simplex_0^0$. Thus, so far, the first iteration of the
homologous
\majorRevision{propagation}
visited a sequence of edges and vertices $\{\simplex_1^0,
\simplex_0^0, \simplex_1^1, \simplex_0^1\}$, such that for each item
$\sigma_i^j$ in this sequence we have:
\emph{(i)} $\sigma_i^j \neq \sigma_i^{j+1}$ and \emph{(ii)} $\sigma_i^{j+1}
\lt \sigma_{i+1}^{j}$, which exactly coincides with the definition of a
discrete integral line (\autoref{sec_discrete_morse_theory}). Then,
along the iterations of \autoref{algo_pairCriticalSimplices}, two integral
lines, started at each vertex of $\simplex_1^0$, will be iteratively
constructed, by selecting at each iteration the highest extremity of the two
integral lines (line \autoref{algo_pairCricitalSimplices_selectSimplex}). This process
terminates when one of the two integral lines reaches a minimum $\sigma_0'$
(i.e., an unpaired vertex, line
\autoref{algo_pairCriticalSimplices_foundCycle}).
At this point, we have $Boundary(\sigma_1^0) = \{\sigma_0' + \sigma_0''\}$,
where $\sigma_0''$ is the extremity of the other integral line (\autoref{fig_overview_minSaddle}, left). Then, if
$\sigma_0' \neq \sigma_0''$ (i.e. $\sigma_1^0$ did not create a $1$-cycle), we
have $Boundary(\sigma_1^0) \neq \emptyset$
(line \autoref{algo_pairCriticalSimplices_nonEmpty}) and a
persistence
pair $(\sigma_0', \sigma_1^0)$ is created (line
\autoref{algo_pairCriticalSimplices_pairCreated}).
Then, at this stage, the boundary
\majorRevision{propagation}
completed the first integral line
from $\sigma_1^0$ down to $\sigma_0'$ and paused the second integral line at
$\sigma_0''$ and we have,
by construction,
$Boundary(\sigma_1^0) \sim \partial
\sigma_1^0$. Next, it is possible that later in the algorithm, the
expanded boundary $Boundary(\sigma_1')$ of another critical $1$-simplex
$\sigma_1'$ hits the minimum $\sigma_0'$. In such a case, the expanded boundary
of its paired simplex (i.e. $Boundary(\sigma_1^0)$) will then be added
(modulo-2) to $Boundary(\sigma_1')$
(line
\autoref{algo_pairCriticalSimplices_mod2}) and the second integral line started
in $\sigma_1^0$ (paused at $\sigma_0''$) will eventually be resumed from
$\sigma_0''$ until it hits another minimum (\autoref{fig_overview_minSaddle}, center).

Overall, for $\diagram_0(f)$,
\autoref{algo_pairCriticalSimplices} will exactly visit the edges and vertices of
$\domain$ which are located on the unstable sets of the critical $1$-simplices.
It follows that
 the homologous
 \majorRevision{propagation}
 of a critical $1$-simplex $\simplex_1$ can be
accelerated
 by directly considering \majorRevision{its} unstable set, whose boundary
($\{\simplex_0'+\simplex_0'''\}$, c.f. above) is homologous by construction to
$\partial \simplex_1$.

Note that this observation no longer holds in higher dimensions. For instance, when constructing  $\diagram_1(f)$ on a 3-dimensional simplicial complex,
in contrast to the case of $\diagram_0(f)$ described above,
the unstable set of a critical $2$-simplex $\simplex_2$ may become non-manifold (as described by Gyulassy and Pascucci \cite{gyulassy11} in the study of Morse-Smale complexes, yielding multiple integral lines between a given pair of critical simplices). In such a case, the boundary  of the unstable set of $\simplex_2$ is no longer exactly homologous to $\partial \simplex_2$ (due to the non-manifold elements of the surface) and the acceleration described
above
is
no longer
applicable.

\noindent
\textbf{\emph{(b)} Unstable set compression}
The computation of $\diagram_0(f)$
can be
further accelerated by \emph{compressing} all
unstable sets.
Given the unstable sets of the critical
$1$-simplices,
we collapse all their regular
edges (which are involved in zero-persistence pairs, \autoref{sec_observations}).
This collapse
eventually results in
a graph
$G_{\diagram_0(f)}$ (\autoref{fig_overview_minSaddle}, center), whose nodes and arcs respectively correspond to the
vertices and edges of $\domain$ which are left unpaired by
$\discreteGradientField$ and whose adjacency relations are determined by
the
input unstable sets. At this point, $G_{\diagram_0(f)}$ can be directly given as an input to \autoref{algo_pairCriticalSimplices} to compute  $\diagram_0(f)$.

\noindent
\textbf{\emph{(c)} Connectivity tracking}
Given $G_{\diagram_0(f)}$, we further accelerate the process and simplify \autoref{algo_pairCriticalSimplices} by exploiting the specific dimensionality of $\diagram_0(f)$. In particular,
in the case of $\diagram_0(f)$, \autoref{algo_pairCriticalSimplices}
visits the arcs of $G_{\diagram_0(f)}$ in increasing (original) lexicographic order.
For a given arc $\simplex_1$, two cases can occur.

First \emph{(i)}, the highest node $\simplex_0$ of $\simplex_1$ has not been
visited yet by any
\majorRevision{propagation}
(line \autoref{algo_pairCriticalSimplices_foundUnpaired}) and a
persistence pair $(\simplex_0, \simplex_1)$ is created in  $\diagram_0(f)$ (line
\autoref{algo_pairCriticalSimplices_pairCreated}).
Then the arc $\simplex_1$ can be collapsed (similarly to regular edge
compression, above paragraph (b)) to indicate that it can no longer be paired by
the algorithm. This collapse can be modeled by a \emph{union} operation,
indicating that the other node $\simplex_0'$ of $\simplex_1$ becomes the
\majorRevision{\emph{representative}}
of $\simplex_0$ (which can no longer be paired).

Second \emph{(ii)}, the highest vertex $\simplex_0$ of $\simplex_1$ has already
been visited by a prior
\majorRevision{propagation},
in which case we need to efficiently \emph{find} its other boundary
vertex $\simplex_0'$ (line \autoref{algo_pairCriticalSimplices_mod2}) to resume
the
\majorRevision{propagation}
there.

Overall, $\diagram_0(f)$ can be computed from $G_{\diagram_0(f)}$ by collapsing
its arcs as they are visited and recording these collapses with a \emph{union}
operation, such that boundary nodes can  later be  retrieved with a \emph{find}
operation. This can be efficiently implemented with a Union-Find data structure
\cite{cormen}, since for $\diagram_0(f)$, each node needs to record only one
\majorRevision{representative}
(the
\majorRevision{representative}
of the other node of
its paired arc).

\noindent
\textbf{\emph{(d)} Summary}
Overall (\autoref{fig_overview_minSaddle}), our algorithm computes $\diagram_0(f)$ by first constructing the unstable sets of each critical $1$-simplex.
Next, each regular edge in these unstable sets is collapsed
to create the graph $G_{\diagram_0(f)}$. Finally, $G_{\diagram_0(f)}$ is processed with a Union-Find data structure \cite{cormen} to compute $\diagram_0(f)$. Initially a Union-Find node $UF(\simplex_0)$ is created for each node $\simplex_0$ of $G_{\diagram_0(f)}$ and the arcs of $G_{\diagram_0(f)}$ are processed in increasing (original) lexicographic order.
Given an arc  $\simplex_1$, its two expanded boundary nodes $\simplex_0$ and $\simplex_0'$ are efficiently retrieved by applying the \emph{find} operation on the two nodes of $\simplex_1$.
Then, if $\simplex_0$ is strictly higher than $\simplex_0'$, the persistence
pair $(\simplex_0, \simplex_1)$ is created in $\diagram_0(f)$ and a \emph{union}
operation is performed between the nodes $UF(\simplex_0)$ and $UF(\simplex_0')$,
and the unpaired node $UF(\simplex_0')$ is used as a
\majorRevision{representative}.

\subsection{Saddle-Maximum Persistence Pairs}
\label{sec_maxDiagram}

\begin{figure}
  \includegraphics[width=\linewidth]{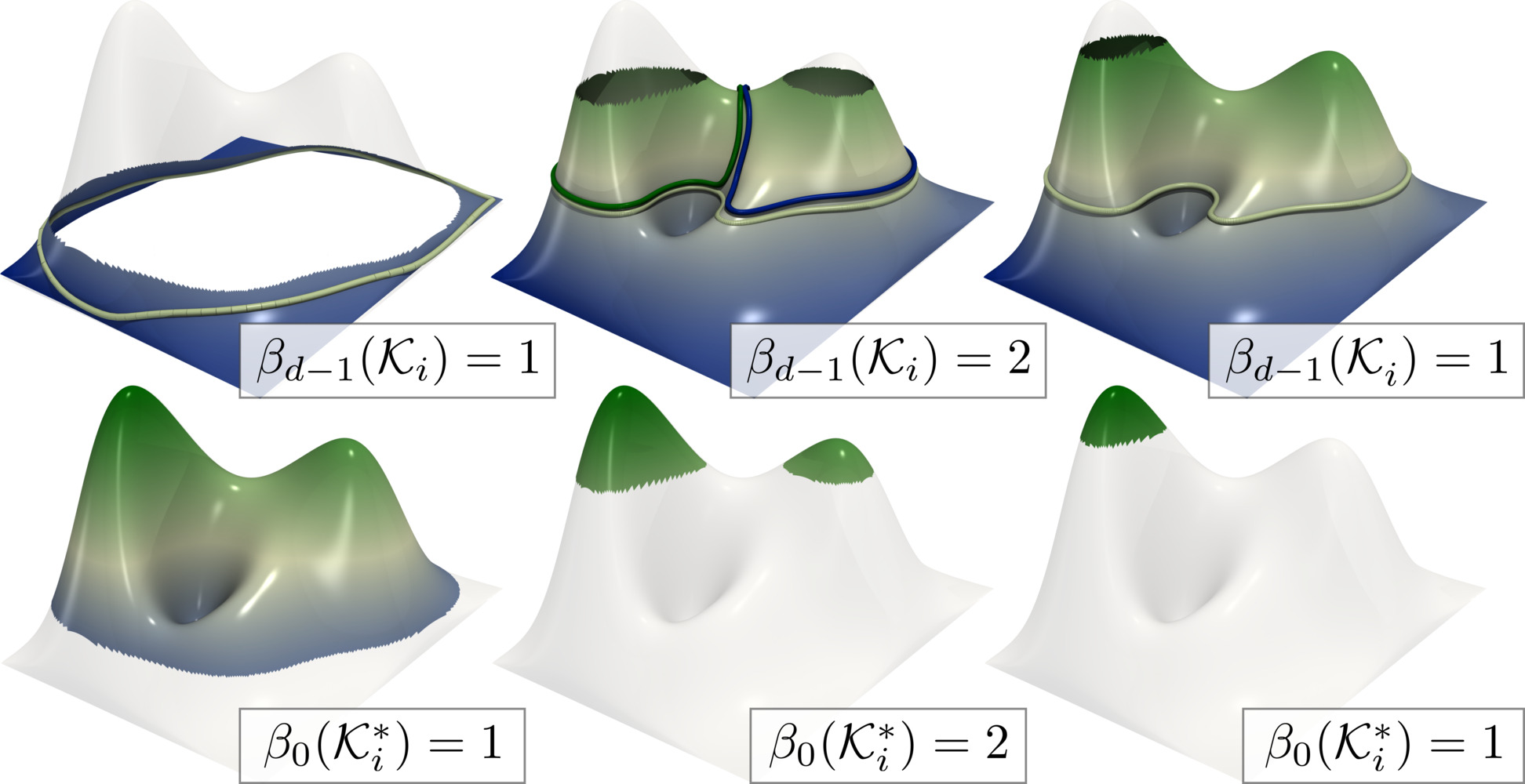}
  \mycaption{The Betti number \julien{$\beta_{\dimensionality - 1}(\domain_i)$}
of a given step of the filtration $\domain_i$ (top) equals the Betti number
$\beta_0(\domain^*_i)$ of the complement $\domain^*_i$ of $\domain_i$ (bottom),
assuming that the boundary $\partial \domain$ of $\domain$ is fully included in
$\domain_i$.}
%
  \label{fig_duality}
\end{figure}

In this section, we detail our strategy for the computation of
$\diagram_{\dimensionality - 1}(f)$.
In particular, we exploit within the
DMT
setting
%
the duality argument
discussed by Edelsbrunner \majorRevision{et al.
\cite{edelsbrunner02, edelsbrunner09}},
recently
\majorRevision{revisited}
for general cell complexes in higher dimensions
\cite{garin20}\majorRevision{, and we document its implementation in the
DMT setting.}
\majorRevision{This}
duality argument
\majorRevision{(}\autoref{fig_duality}) states that, at a given step $i$ of the
filtration, the $(d-1)$-dimensional voids of $\domain_i$, under certain
conditions, exactly coincide with the connected components of the
\emph{complement} $\domain_i^*$ of $\domain_i$.

%




Formally, let $\domain^*$ be the dual cell complex of $\domain$. Specifically, each $(d-i)$-simplex $\simplex_i$ of $\domain$ is represented by an $i$-dimensional cell $\simplex_i^*$ in $\domain^*$. Moreover, given two simplices $\simplex_i < \simplex_j$
in $\domain$, we have $\simplex_j^* < \simplex_i^*$ in $\domain^*$ (i.e. face-coface relations are reversed). Then, it follows that each diagram $\diagram_{d-k-1}(f)$ of the lexicographic filtration of $\domain$ is equal to the opposite of the diagram $\diagram_{k}(-f)$ (i.e. the diagram of the \emph{backward} lexicographic filtration, $-f$) of $\domain^*$ (see Garin et al. \cite{garin20}, Theorem 2.1). This implies in particular that $\diagram_{\dimensionality - 1}(f)$ can be computed very efficiently by applying
the algorithm for $\diagram_0(f)$ described in \autoref{sec_diagram0} to the \emph{backward} filtration (i.e.$-f$, reverse order) of the dual $\domain^*$ of $\domain$. This observation nicely translates to the
DMT
setting, as described next.

\begin{figure}
  \includegraphics[width=\linewidth]{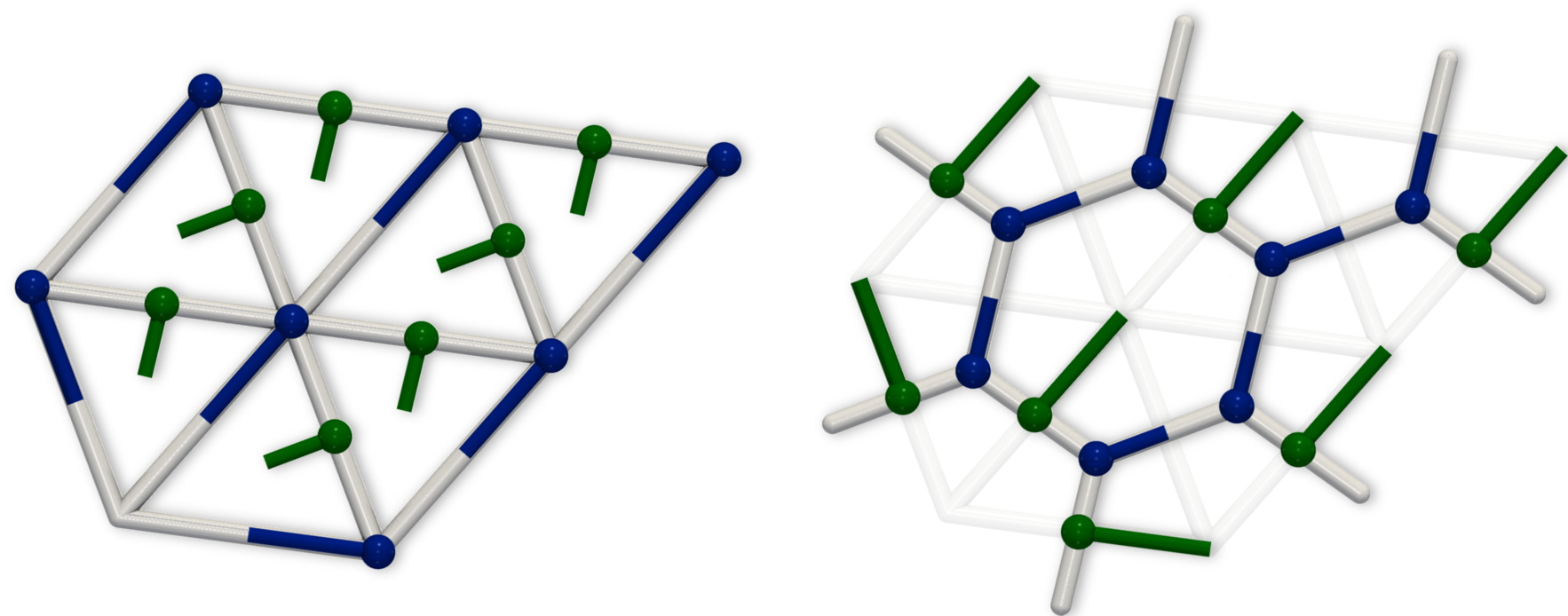}
  \mycaption{Given a discrete gradient field $\discreteGradientField$ defined
on $\domain$ (left), its \emph{dual} discrete gradient field
$\discreteGradientField^*$ (right) is obtained by considering the dual cell
complex $\domain^*$ and reverting each arrow of $\discreteGradientField$: each
vertex-edge arrow (blue, left) becomes an edge-face arrow (green, right) while
each edge-triangle arrow (green, left) becomes a vertex-edge arrow (blue,
right), along which unstable sets can be easily defined and computed.}
  \label{fig_duality_gradient}
\end{figure}

A dual discrete gradient vector field $\discreteGradientField^*$ (\autoref{fig_duality_gradient}) can be easily
defined on the dual $\domain^*$ of $\domain$ by reverting each discrete vector of $\discreteGradientField$. In particular, each discrete vector $\{\simplex_{d-1}, \simplex_d\}$ between a $(d-1)$-simplex $\simplex_{d-1}$ of  $\domain$ and one its cofacets can be reverted into
$\{\simplex_d^*, \simplex_{d-1}^*\}$, where $\simplex_d^*$ and $\simplex_{d-1}^*$ are the simplices dual to $\simplex_{d}$ and $\simplex_{d-1}$ in $\domain^*$. Then $\{\simplex_d^*, \simplex_{d-1}^*\}$ is a discrete vector between a $0$-simplex ($\simplex_d^*$) and a $1$-simplex ($\simplex_{d-1}^*$). Once this is established, the algorithm described in \autoref{sec_diagram0} can be applied as-is on $\discreteGradientField^*$.
Additionally, one can observe that the critical $1$-simplices of $\discreteGradientField^*$ will be, by construction, critical $(d-1)$-simplices of $\discreteGradientField$ and that their \emph{unstable} sets in $\discreteGradientField^*$ will exactly coincide to \emph{stable} sets in  $\discreteGradientField$.


Thus, our algorithm for computing $\diagram_0(f)$ (\autoref{sec_diagram0}) can be easily adapted to compute $\diagram_{d-1}(f)$ as follows. The stable sets of each critical $(d-1)$-simplex are first constructed. Next, each discrete vector in these stable sets is collapsed, to create a graph $G_{\diagram_{d-1}(f)}$, where each node represents a critical $d$-simplex and each arc a critical $(d-1)$-simplex. Finally, $G_{\diagram_{d-1}(f)}$ is processed with a Union-Find data structure
(\autoref{sec_diagram0}), but
in decreasing lexicographic order, and a persistence pair $(\simplex_{d-1}, \simplex_d)$ is created in  $\diagram_{d-1}(f)$ for each connected component of $G_{\diagram_{d-1}(f)}$ created in $\simplex_d$ and merged into another by the addition of the arc representing $\simplex_{d-1}$.

\noindent
\textbf{Domains with boundary}
When $\domain$ is not closed, a slight variation of the above algorithm
is
considered.
For domains with boundary, in specific configurations, the connected components
of the backward lexicographic filtration of $\domain^*$
%
may no longer exactly coincide with the voids of the forward lexicographic filtration of
$\domain$. In particular, when a connected component of
the backward filtration of $\domain^*$
first
hits the
outer
boundary component
of $\domain$
(by construction, on
a critical $(\dimensionality - 1)$-simplex), it no longer
describes
a
void \emph{inside} the object, as it merges with the rest of the
outside space (thus
deleting the corresponding cavity). To take this into account, we
assign a
\emph{virtual} discrete maximum with infinite function value to the outer boundary component of $\domain$
(representing the outside space)
%
and apply the rest of the above algorithm as-is. Then, when a
connected component of
backward filtration
hits the outer boundary, it is
considered, given the above adjustment, to die there as it merged with an
(infinitely) older component (the outside space).


Note that this specific adjustment comes with no additional computational
overhead as the rest of our algorithm is used as-is (only one, extra virtual
maximum is considered by the algorithm). In our implementation, this adjustment
is optional as its practical relevance can be questionable for real-life data,
as illustrated in
Appendix \majorRevision{A}.

\begin{figure*}
\includegraphics[width=\linewidth]{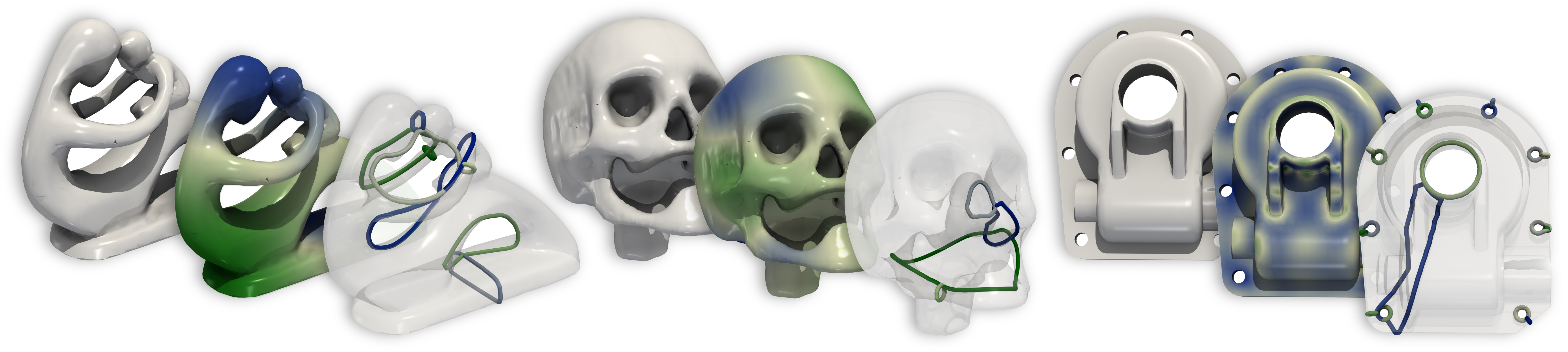}
\vspace{-2ex}
\mycaption{Extracting surface generators with eigenfunctions of the
Laplace-Beltrami operator (center). The infinitely persistent $1$-cycles of this
very smooth scalar field (curves, right) smoothly capture each topological
handle, facilitating further surface post-processing (e.g. parametrization).}
 \label{fig_applicationGenerator}
\end{figure*}

\begin{figure}
        \includegraphics[width=\linewidth]{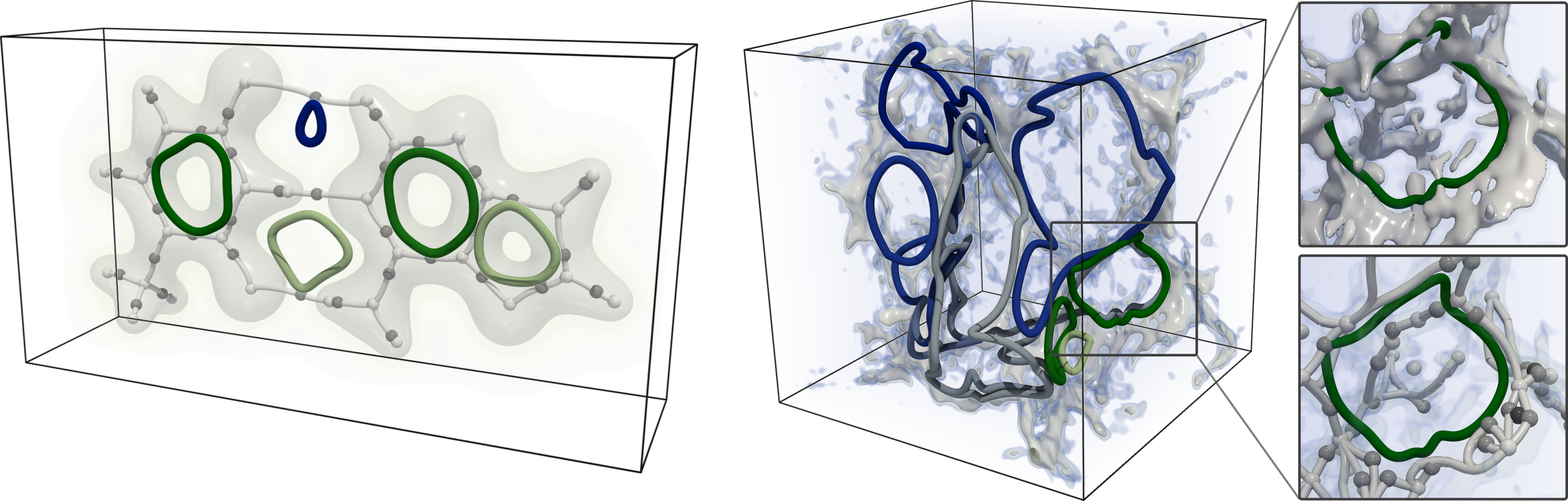}
 \mycaption{Persistent $1$-cycles (colored by persistence) in volume data
capture prominent loops in the sub-level sets.
 In a quantum chemistry example (electron density, left), the $3$ most
persistent $1$-cycles exactly coincide with the $3$ carbon rings of the
molecular system, while the $2$ least persistent cycles coincide with weaker,
non-covalent interactions. In an astrophysics example (dark matter density,
right), persistent $1$-cycles capture prominent loops in the \emph{cosmic web}
\cite{sousbie11}, indicating a very dense, circular pattern of galaxies,
arranged around a prominent void (inset zoom).
 In both cases,  $1$-dimensional separatrices of the Morse-Smale complex are
shown in the background for visual context.}
 \label{fig_applicationVoxelData}
\end{figure}

\section{Critical Simplices of Infinite Persistence}
\label{sec_infinitePersistence}

To summarize, our overall approach first
computes $\diagram_0(f)$
(\autoref{sec_diagram0}) and $\diagram_{d-1}(f)$ (\autoref{sec_maxDiagram}).
%
%
%
Finally, if $d=3$, $\diagram_1(f)$ is computed with our novel algorithm \emph{``PairCriticalSimplices''} (\autoref{sec_saddleSaddle}).
%
%
%
During this process,
%
certain critical simplices may remain unpaired
after the above algorithms have finished. These correspond to
homology classes with infinite persistence, which exactly characterize the
homology of $\domain$. Specifically, each remaining unpaired $i$-simplex
$\sigma_i$ yields a persistence class with infinite persistence in
$\diagram_i(f)$, which we embed, by convention at location
$\big(f(\sigma_i), f^*\big)$, where $f^*$ denotes the maximum $f$ value.
Such points in the diagrams are marked with a specific flag (\autoref{fig_diagram}), as they
describe more the domain $\domain$ itself than the data
$f$
defined on it.

%
%
%
%
%




\section{Computational Aspects}
\label{sec_parallel}

This section details the computational aspects of our algorithm, including time complexity and parallelism.

\subsection{Time complexity}
\label{sec_timeComplexity}
The first stage of our approach consists in establishing the lexicographic filtration with a global sort in
$\bigO\big(n \majorRevision{\log}(n)\big)$ steps (where $n$ is the total number
of simplices in $\domain$).

The second stage computes a discrete gradient by
\majorRevision{expansions}
\cite{robins_pami11}.
This operation takes $\bigO(n_v)$ where $n_v$ is the number of vertices in $\domain$.

The third stage consists in computing $\diagram_0(f)$ (\autoref{sec_diagram0}). The first step of this algorithm computes the unstable sets of each $1$-saddle to construct the graph $G_{\diagram_{0}(f)}$, which is done in  $\bigO(n_e)$ steps in practice, where $n_e$ is the number of edges in $\domain$. Processing $G_{\diagram_{0}(f)}$ with a Union-Find data-structure to finally construct $\diagram_0(f)$ takes $\bigO\big(n_e \alpha(n_e)\big)$ steps, where $\alpha()$ is the extremely slowly-growing  inverse of the Ackermann function. Computing $\diagram_{\dimensionality-1}(f)$ requires the same steps.

The fourth stage of our approach, only
for
$\dimensionality = 3$, applies our algorithm
\emph{``PairCriticalSimplices''} (\autoref{algo_pairCriticalSimplices}).
Similarly to the seminal algorithm
\majorRevision{\emph{``PairSimplices''}
\cite{edelsbrunner02, zomorodianBook}},
our algorithm requires $\bigO(n^3)$ steps in the worst case.
For each critical simplex (in the worst case, $n$ steps, line
\autoref{algo_pairCriticalSimplices_firstFor}), an homologous
\majorRevision{propagation}
is
performed (in the worst case, in $n$ steps, lines
\autoref{algo_pairCriticalSimplices_startExtension} to
\autoref{algo_pairCriticalSimplices_endExtension}), which itself requires at
each step a possibly linear pass to expand the boundary of the current critical
simplex with modulo-2 additions (line
\ref{algo_pairCriticalSimplices_expansion}). However, as documented in
\autoref{sec_perfAnalysis}, our algorithm \emph{``PairCriticalSimplices''}
performs in practice significantly faster than the algorithm
\majorRevision{\emph{``PairSimplices''}}
since:
\emph{(i)} it only considers the critical simplices (and not all the simplices of $\domain$),
\emph{(ii)} the critical simplices already present in $\diagram_0(f)$ and $\diagram_{\dimensionality-1}(f)$ are discarded from the computation (which
provides further accelerations),
\emph{(iii)} it maintains the expanded boundary of the considered critical simplex and not its expanded chain (which is significantly bigger).

Overall, our approach has
the advantage of being
output-sensitive. In particular, the
size (number of nodes) of the graphs $G_{\diagram_{0}(f)}$ and $G_{\diagram_{\dimensionality-1}(f)}$ corresponds to
the number of minima and maxima of $f$ and consequently to
the size of $\diagram_0(f)$ and $\diagram_{d-1}(f)$. The time complexity of our algorithm \emph{``PairCriticalSimplices''} is parameterized by the number of remaining saddle-saddle pairs, which corresponds to the size of $\diagram_1(f)$. Then our approach will provide superior performances when considering smooth data sets, as typically found in various simulation domains.

\subsection{Shared memory parallelism}
\label{sec_parallelism}
Our approach can benefit from further accelerations thanks to shared-memory parallelism.
The first stage (establishing the lexicographic filtration) can be done with  parallel sorting  (see the GNU parallel sort for an implementation example).
The second stage (discrete gradient computation \cite{robins_pami11}) is trivially parallelizable on the vertices of $\domain$.
Regarding the third stage, computing $\diagram_0(f)$, the computation of the unstable sets is parallelized on a per $1$-saddle basis and the processing of $G_{\diagram_{0}(f)}$ with the Union-Find data-structure is then done sequentially.
In practice $\diagram_0(f)$ and $\diagram_{\dimensionality - 1}(f)$ are computed in parallel thanks to a task pool mechanism.
Regarding the fourth stage (\autoref{algo_pairCriticalSimplices}),
\majorRevision{the homologous
\majorRevision{propagations}
can be computed in parallel for each
$2$-saddle,
using lightweight synchronizations \cite{gueunet_tpds19,
Nigmetov20}. Specifically,
when
the homologous
\majorRevision{propagation}
of a
$2$-saddle
$\simplex_2$ hits an unpaired
$1$-saddle
$\simplex_1$, a
\emph{temporary} persistence pair $(\simplex_1, \simplex_2)$ is created in
$\diagram_1(f)$. Later, if the
homologous
\majorRevision{propagation}
of another
$2$-saddle
$\simplex'_2$
also
hits
$\simplex_1$, an
atomic \emph{compare-and-swap} operation is performed. Specifically, if
$\simplex'_2$ is anterior to $\simplex_2$ in the lexicographic filtration, the
pair $(\simplex_1, \simplex_2)$ is updated into $(\simplex_1, \simplex'_2)$ and
the
\majorRevision{propagation}
of $\simplex_2$ is resumed at $\simplex_1$.}


\begin{figure}
\includegraphics[width=\linewidth]{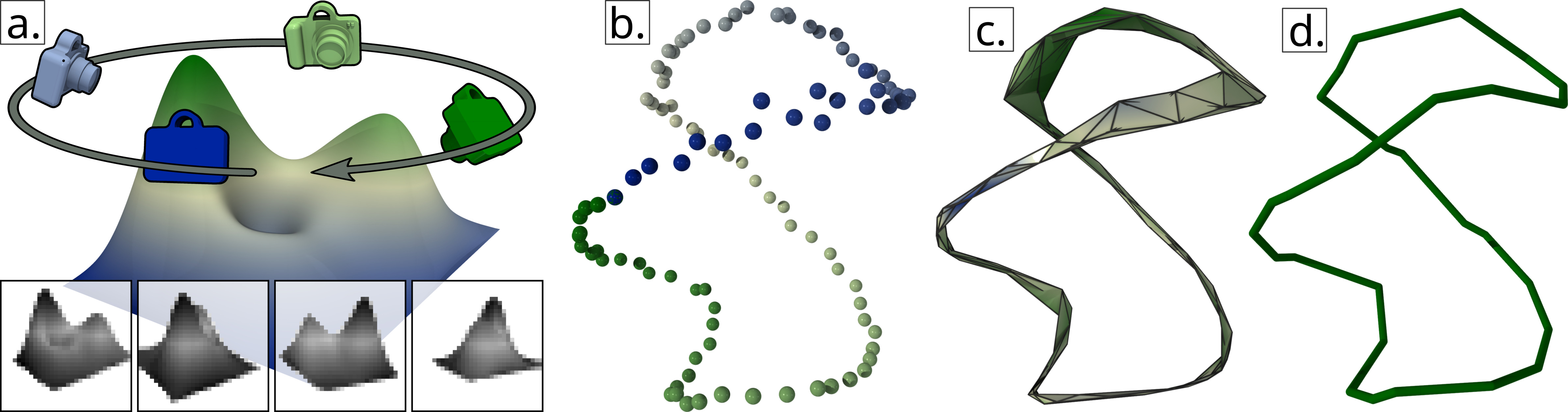}
 \mycaption{Detecting circular patterns in high dimensional data.
 \emph{(a)} $100$ greyscale pictures of resolution $32\times32$ (bottom) of a
synthetic terrain (center) are taken from $100$ viewpoints (colored cameras)
arranged along a circle (camera color: arc-length parameterization along the
circle).
 \emph{(b)} This set of images
 can be interpreted as $100$ points in $\mathbb{R}^{32\times32}$, which can be
projected down to 3D via Multi-Dimensional Scaling (MDS) \cite{mds} (color:
camera arc-length parameterization).
  \emph{(c)} The $2$-dimensional Rips complex can be computed from the point
cloud in the high-dimensional space ($\mathbb{R}^{32\times32}$) to infer the
structure of the space sampled by the point cloud, by adding a triangle in the
complex if its \emph{diameter} (the maximum pairwise distance between its
vertices) is smaller than a threshold $\epsilon$.
  The Rips complex is shown in 3D (via MDS projection), although it is computed
in $\mathbb{R}^{32\times32}$.
  \emph{(d)} The infinitely persistent $1$-cycle of the \emph{diameter} function
(for each vertex, average of the diameter of its adjacent triangles) robustly
captures the circular pattern synthetically injected in the data (\emph{(a)}),
hence confirming the ability of persistent $1$-cycles to recover circular
patterns in high-dimensional data.
%
 }
 \label{fig_highDimensionalData}
\end{figure}

\section{Application to Generator Extraction}
\label{sec_applications}

This section presents an application of our contributions to the fast
extraction of persistent $1$-dimensional generators.
While the topological persistence computed by our algorithm is a central simplification criterion in data visualization (\autoref{fig_ttkOverview}), the information maintained by our algorithm can additionally be exploited directly for visualization purposes.
Specifically,
Iurichich \cite{iuricich21} suggested to extract, for a given persistence pair $(\sigma_i, \sigma_j)$, a representative $\dimensionality_i$-cycle homologous to $\partial \sigma_j$, specifically, the earliest homologous $\dimensionality_i$-cycle, created at $\sigma_i$. For that,
Iurichich
introduced a specific post-processing algorithm \cite{iuricich21}, requiring the persistence diagram to be computed in a pre-processing step.

In contrast, in our work, this information is precisely maintained throughout the entire computation, for all $1$-dimensional persistence pairs, and is then readily available when our persistence diagram computation algorithm has finished, resulting in further accelerations.
Specifically,
for each critical $2$-simplex $\sigma_j$, the homologous
\majorRevision{propagation}
of
\autoref{algo_pairCriticalSimplices} (lines
\ref{algo_pairCriticalSimplices_startExtension} to
\ref{algo_pairCriticalSimplices_endExtension}) iteratively reconstructs with
$Boundary(\sigma_j)$ a sequence of $1$-cycles homologous to $\partial \sigma_j$
and any of these can be chosen as a representative generator. Specifically, we
store the earliest cycle \majorRevision{(which is not necessarily optimal
\cite{DeyHM19})}, \majorRevision{which is the cycle} precisely obtained at the
end of the homologous
\majorRevision{propagation}
(line \ref{algo_pairCriticalSimplices_nonEmpty}), when the first
unpaired simplex
$\tau$ is visited. Note that the problem of extracting persistent  $0$ and $(\dimensionality-1)$-dimensional generators is significantly simpler and has already been addressed via merge tree based segmentations \cite{carr04, gueunet_ldav16, topoAngler}.

\begin{figure}
\includegraphics[width=\linewidth]{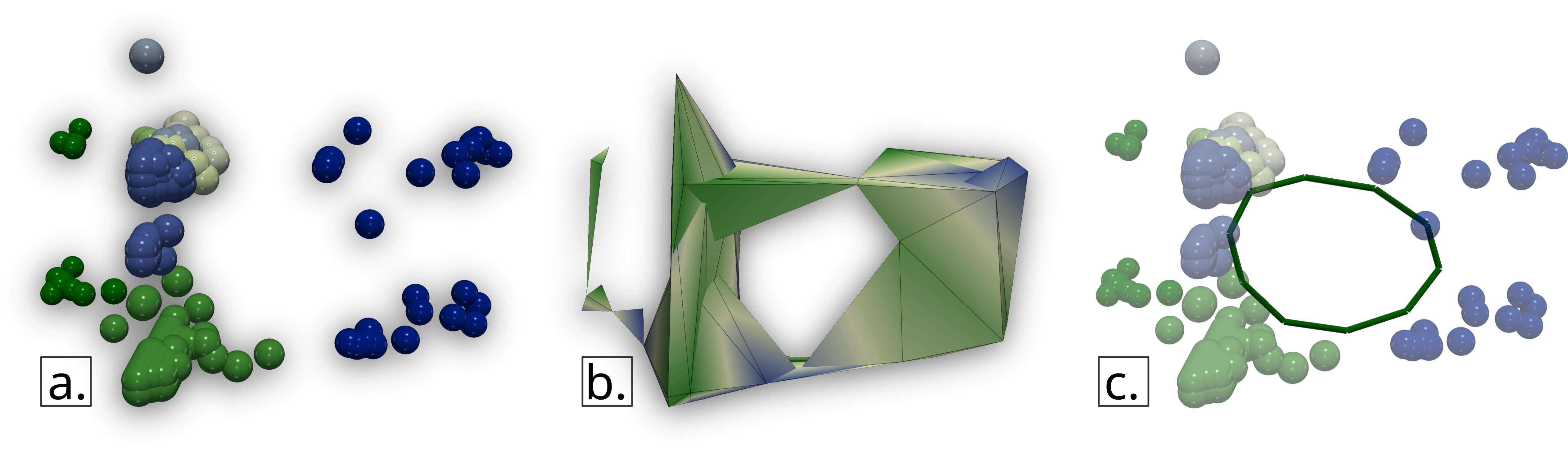}
\mycaption{Seven measurements of electric power consumption (including active
power, voltage, intensity) for a single household over two years (daily
sampling) \cite{householdElectricityData}.
This dataset can be interpreted as $700$ points (one per day) in $\mathbb{R}^7$ (projected in 3D in \emph{(a)} via MDS \cite{mds}).
\emph{(a)} The $k$-means clustering ($k=8$) applied to this point cloud identifies dense clusters (point colors), corresponding to distinct consumption modes.
\emph{(b)} The 2D Rips complex (computed in $\mathbb{R}^7$, but projected in 3D via MDS \cite{mds}) exhibits a prominent topological handle.
\emph{(c)} The \emph{diameter} function (see \autoref{fig_highDimensionalData}) yields a highly persistent $1$-cycle (green curve) between four clusters
(light blue, white, dark blue, light green).
This indicates that these four clusters are organized along a circular pattern in $\mathbb{R}^7$, implying in practice that continuous displacements from the white  consumption state to the light green one,
are likely to
imply a transition along the cycle, either through the light blue or dark blue consumption states.}
\label{fig_realHighDimensionalData}
%
\end{figure}

Figs. \ref{fig_applicationGenerator}, \ref{fig_applicationVoxelData}, \ref{fig_highDimensionalData}
and \ref{fig_realHighDimensionalData}
illustrate the ability of persistent $1$-cycles to robustly capture circular patterns on surfaces, volume data and high-dimensional point clouds respectively.


%
%
%
%
%




\section{Results}
\label{sec_results}
This section presents experimental results obtained with a C++/OpenMP implementation of our approach,
integrated in TTK (commit: 
\majorRevision{\href{https://github.com/topology-tool-kit/ttk/commit/bb3089f07a5039433dfab6bab7455e23678ec6b3}{bb3089f}}).
Our experiments were mostly run on a commodity desktop computer with two Xeon
CPUs (\julien{3.0 GHz, 2x4 cores, 64 GB of RAM}), while specific scalability 
experiments were run on a large shared-memory system with 128 Xeon CPUs (2.6 
GHz, 128x8 cores, 16TB of RAM).

\subsection{Experimental data}
\label{sec_data}

We consider a list of 34
\julien{scalar}
datasets available on a public repository
\cite{openSciVisDataSets}, provided as
\majorRevision{3D}
regular grids of
various resolutions and data types.
These datasets come from diverse
fields (bio-imaging, material sciences, combustion, quantum
chemistry, fluid dynamics) and have been either acquired
(e.g. CT scans)
or simulated.
We
\majorRevision{also}
consider two extreme cases:
\emph{(i)} an elevation function (yielding the smallest possible output, a single bar in $\diagram_0(f)$, with infinite persistence),
\emph{(ii)} a random function (yielding the largest outputs in practice).
Since our approach is output sensitive, we re-sampled all datasets to a common resolution ($192^3$),
to better observe runtime variations solely based on the output size.
This common resolution has been chosen such that most of the implementations
considered in our benchmark (\autoref{sec_benchmark}) could run on our
setup.
Some
available datasets
\cite{openSciVisDataSets} were too large
to fit in the memory of our desktop computer
and could not be downsampled to the common resolution.
These have not been
considered
in the benchmark
as we believe our desktop computer to be representative of the machines used by  potential benchmark users.

We generated 2D datasets by taking a slice of each original 3D regular grid along the
Z-coordinate (at mid-value).
These 2D datasets were re-sampled to
$4,096^2$.
Finally, we generated 1D datasets by considering a line of each 2D dataset
(Y-coordinate, mid-value).
These 1D datasets were re-sampled to
a common resolution of
$1,048,576$ vertices.

\begin{table*}
        \caption{Public implementations considered in our benchmark
(\autoref{sec_benchmark}).}
        \vspace{-1ex}
\rowcolors{3}{gray!20}{white}
        \resizebox{\linewidth}{!}{
        \begin{tabular}{|l|rrcrcccr|}
\hline
Implementation & Ref. & Version & Category & Language & Simplicial Support & Grid Support & Parallelism & Distance\\\hline
\majorRevision{PairSimplices}
& \cite{edelsbrunner02, zomorodianBook} &
\href{https://github.com/topology-tool-kit/ttk/tree/
362e69c8c10871bdc039c37c2d1aac47b236663b}{Github 362e69c} & Explicit Homologous
\majorRevision{Propagation} & C++ & Native & Implicit & No & $\textbf{0.0}$\\
\hline
CubicalRisper & \cite{cubicalRipser} & \href{https://github.com/CubicalRipser/CubicalRipser_3dim/tree/a063dac8ef646ff838ff10c14c0adc9acefd9972}{Github a063dac}
& Boundary Matrix Reduction & C++ & No & Native & No& NA\\
Dionysus2 & \cite{dionysus2} &
  Pypi v2.0.8
  & Boundary Matrix Reduction & C++ & Native & No & No& $\textbf{0.0}$\\
DIPHA & \cite{dipha} &
  \href{https://github.com/DIPHA/dipha/tree/0b874769fbd092c07a12cebc2459adb02117c2fd}{Github 0b87476} & Boundary Matrix Reduction & C++ & Native & Native & Controllable & $\textbf{0.0}$\\
Eirene.jl & \cite{eirene} &
  Julia 1.3.6 & Boundary Matrix Reduction & Julia & Native & No & No&
  $9.0 \times 10^3$
  \\
Gudhi & \cite{gudhi} &
  \href{https://github.com/GUDHI/gudhi-devel/tree/845b02ff408eb50207165b8e11136e4b1888612a}{Github 845b02ff} & Boundary Matrix Reduction & C++ &  Native & Native & Observed &
  $15.3 \times 10^3$\\
Javaplex & \cite{javaplex} &
\href{https://github.com/appliedtopology/javaplex/releases/tag/4.3.4}{Github v4.3.4}
  & Boundary Matrix Reduction & Java & Native & No & Observed& $\textbf{0.0}$\\
Oineus (Python API) & \cite{oineus} &
\href{https://github.com/grey-narn/oineus/tree/f2dd92ea00e4cf65da068b09cb6061b7c573740a}{Github f2dd92e} & Boundary Matrix
Reduction & C++ & No & Implicit & Controllable& NA\\
PHAT (Spectral Seq.) & \cite{BauerKRW17} &
\href{https://bitbucket.org/phat-code/phat/src/master/}{Bitbucket 264f0a7} &
Boundary Matrix Reduction & C++ & Native & No & Controllable&
$466.6 \times 10^3$\\
Ripser.py & \cite{ripser,ctralie2018ripser} & Pypi v0.6.0 & Boundary Matrix Reduction & C++ & Native & No & No& NA\\
\hline
Diamorse & \cite{diamorse} &
\href{https://github.com/AppliedMathematicsANU/diamorse/tree/3416d7a6ffa13b2fce7d83c560ac6bc83f1faa44}{Github 3416d7a} & Discrete Morse Theory & C++ & No & Native & No& NA\\
Perseus & \cite{perseus} & \href{http://people.maths.ox.ac.uk/nanda/perseus/}{Author WebPage}
& Discrete Morse Theory & C++ & Native & Native & No& NA\\
PersistenceCycles & \cite{iuricich21} & \href{https://github.com/IuricichF/PersistenceCycles}{Github b68ae3e} & Discrete Morse Theory & C++ & Native & Implicit & Controllable&
$97.5 \times 10^3$
\\
\hline
TTK-FTM & \cite{gueunet_tpds19} & v0.9.9 & Merge-Tree (2D) & C++ & Native & Implicit & Controllable&
$122.5 \times 10^6$\\
%
\hline
\end{tabular}
}
\label{table_implementations}
\end{table*}

Each of these 1D, 2D and 3D datasets were then triangulated into a simplicial complex by breaking up each cell into two triangles in 2D, and five tetrahedra in 3D.
As discussed in \autoref{sec:introduction}, our approach focuses on this generic input representation based on simplicial complexes and we will therefore consider these representations for our experimentations. This results overall in 108 input datasets.

As described in \autoref{sec_benchmark},
 some of the public implementations considered in our benchmark are specialized (or include specialized backends) for regular grids. However, they do not all interpret the input data in a consistent manner. For instance, some implementations (such as Gudhi) consider the input scalars to be defined on a per voxel basis, while others (such as Dipha or CubicalRipser) consider them as defined on a per vertex basis, which results in cell complexes of
 significantly
 different sizes (in particular, penalizing Gudhi).
Moreover, some implementations (such as Oineus,
\majorRevision{PairSimplices},
PersistenceCycles, TTK-FTM, DMS) implicitly triangulate the input
regular grid data \cite{freudenthal42, kuhn60}, which also changes the size of
the input complex.
First, since these internal data representations differ, the generated outputs will, consequently, not be exactly identical. Second, since these differences in internal representation result in cell complexes of
significantly
different sizes, they also induce a strong bias in runtime comparison.
For these reasons, we decided to focus our analysis on the methods which natively support simplicial complexes, for which a direct and unbiased comparison can be performed.

For completeness, we provide performance numbers for regular grids in
Appendix \majorRevision{B},
but we stress that the inconsistencies in the internal representations and in the generated outputs prevent a direct and unbiased comparison.

\subsection{Performance analysis}
\label{sec_perfAnalysis}

This section evaluates the time performance of our overall approach, named
\emph{``DiscreteMorseSandwich''}
(\emph{DMS}),
and details the
gains provided by each steps of our algorithm, in comparison to the original
algorithm
\majorRevision{\emph{``PairSimplices''}}.

\begin{figure}
  \centering
  \resizebox{\linewidth}{!}{
    \input{plots/plot_variants_expl_para}
  }
  \mycaption{
\majorRevision{Computation speeds
(simplices per second, log scale),
as a function of the output
size, for the distinct accelerations of our
  algorithm, in comparison to the seminal algorithm
  \majorRevision{\emph{``PairSimplices''}}
  (red,
\autoref{algo_pairCells}).
Each non-red curve corresponds to a
specific variant of our algorithm:
  \emph{``PairCriticalSimplices''} (green, \autoref{algo_pairCriticalSimplices}),
  with boundary caching (yellow, \autoref{sec_algorithmPairCriticalSimplices}),
  with \emph{``Sandwiching''} (purple, \autoref{sec_diagram0}),
  in parallel (blue, $8$ cores, \autoref{sec_parallelism}).
  On average, our parallel algorithm
  computes in 
  \majorRevision{$0.04$}
  (1D), 
  \majorRevision{$1.10$}
  (2D) and 
  \majorRevision{$\textbf{5.21}$}
  (3D) seconds,
and achieves a parallel efficiency of
\majorRevision{$30.68\%$}
(1D), 
\majorRevision{$59.50\%$}
(2D), 
and 
\majorRevision{$\textbf{78.89\%}$}
(3D) for an
overall speedup
over
\majorRevision{\emph{``PairSimplices''}}
of
\majorRevision{$\times 8$}
in 1D, 
\majorRevision{$\times 1,046$}
in 2D, and 
\majorRevision{$\times \textbf{323}$}
in 3D.}
}
  \label{fig_perfCurves}
\end{figure}

\autoref{fig_perfCurves} provides time performance curves for the 1D, 2D and 3D
versions of our 36 input datasets, where computation speeds (in simplices per
second, log scale) are reported as a function of the output size, and where the
seminal algorithm
\majorRevision{\emph{``PairSimplices''}}
is compared to four variants of our approach, to evaluate
the performance gain of each acceleration
introduced in our algorithm.

This figure confirms the output-sensitive behavior of our overall approach (DMS, blue curves), as computation speeds decrease for increased output sizes.
As expected by our time complexity analysis (\autoref{sec_timeComplexity}), the 
lowest speeds occur for 3D datasets 
(\majorRevision{$2.95 \times 10^7$} simplices/sec on average) since there, the 
computation of $\diagram_1(f)$ (the intermediate layer of the sandwich) has a 
less favorable time complexity than for $\diagram_0(f)$ and $\diagram_2(f)$. 
This is confirmed by the increased speed in 2D ($9.17 \times 10^7$ simplices/sec 
on average), while in 1D speed slightly decrease again as the unstable sets of 
$1$-saddles now cover the entire domain (whereas they constitute only a small 
subset in 2D).
In 1D (left), our overall approach (blue) is only about an order of magnitude
faster than the seminal algorithm
\majorRevision{\emph{``PairSimplices''}}.
In 2D (center), the simplest variant \emph{``PairCriticalSimplices''} (green)
starts to provide a significant acceleration (about one order of magnitude
speedup) with regard to
\majorRevision{\emph{``PairSimplices''}}
(red),
\julien{while boundary caching provides another order of magnitude speedup}. The effect of the sandwiching strategy becomes clearly visible in 3D (right).
%
There,
\majorRevision{\emph{``PairSimplices''}}
and \emph{``PairCriticalSimplices''} both timeout after 30
minutes of computation. In 3D, the benefit of the sandwiching approach is
substantial (about a $\times 4$ speedup), as illustrated by the gap between the
\julien{yellow and purple} curves. For all dimensions, the parallelization of
our overall approach (blue, 8 cores) provides about another order of magnitude
speedup over the other variants.
Overall, in 3D, our approach provides an average speedup of two orders of 
magnitude over
\majorRevision{\emph{``PairSimplices''}}
(\majorRevision{$\times 323$}, when considering the runs where 
\majorRevision{\emph{``PairSimplices''}}
did not timeout after 30 minutes),
with computations in less than 
$10$ seconds, 
with about a 
\majorRevision{$79\%$}
parallel efficiency, which can be considered 
as an efficient parallelization.
\majorRevision{Appendix C provides detailed time
statistics, including breakdowns of the computation times for each sub-step of
our approach.}

\subsection{Performance benchmark}
\label{sec_benchmark}

This section describes our benchmark for evaluating and comparing
various public implementations for persistent diagram computation.
Our Python benchmark package
(\href{https://github.com/pierre-guillou/pdiags_bench}{https://github.com/pierre-guillou/pdiags\_bench})
\emph{(i)} downloads and prepares the benchmark data (\autoref{sec_data}), \emph{(ii)} downloads, builds and executes each implementation (\autoref{sec_benchmarkImplementations}) and \emph{(iii)} aggregates the output information to produce the results provided in this section.


%
%

\subsubsection{Implementations}
\label{sec_benchmarkImplementations}
Our benchmark includes the implementation of our algorithm  \emph{``DiscreteMorseSandwich''} (DMS)
as well as 14 other implementations, whose specifications are reported in \autoref{table_implementations}.
A few clarifications are needed regarding certain implementations.
In particular, \emph{TTK-FTM} only computes $\diagram_0(f)$ and $\diagram_{\dimensionality-1}(f)$.
Ripser and its scikit-tda version both reported integer overflows for relatively large inputs (issue communicated to the authors).
A number of implementations (among the category \emph{``Boundary Matrix Reduction''}) require an explicit boundary matrix as an input, which we compute in a pre-processing stage with TTK.

\begin{figure}
  \centering
  \resizebox{\linewidth}{!}{
    \input{plots/plot_expl_seq}
  }
  \mycaption{Benchmark of sequential computation speeds (simplices/second, log
scale) as a function of the output size.
  }
  \label{fig_perfComparisonSequential}
\end{figure}

\subsubsection{Output comparison}
\label{sec_comparison}
To check the correctness of our implementation, we compute the
$L_2$-Wasserstein distance \cite{edelsbrunner09}
between our output
and the one computed by each implementation included in the benchmark, for each
dataset, for each output dimension.
To enable the direct comparison of this distance across datasets,
we consider as input scalar field $f$ the vertex order, after sorting all input data values (i.e. $f(v) \in [0, n_v - 1]$).
%
The average distance for all datasets, for all output dimensions, is reported in the column \emph{``Distance''} of \autoref{table_implementations}.
These numbers show that our implementation generates outputs which are strictly
identical to most other implementations
(\majorRevision{PairSimplices},
Dionysus2, DIPHA, Javaplex, etc.). Variations from $0$ seem to indicate slight
inaccuracies for the corresponding implementation. For instance,
for handling boundary effects, TTK-FTM only implements the second strategy described in
Appendix \majorRevision{A}
(i.e. it ignores the virtual maximum on the boundary), which impacts distance evaluations for $\diagram_{\dimensionality-1}(f)$
(see Appendix \majorRevision{A}).
Note that this distance is only reported for the
(non timed out)
implementations natively supporting simplicial complexes,
as outputs differ significantly in the case of regular grids (depending on the implementation's data interpretation, see \autoref{sec_data}).

\subsubsection{Performance metrics}
We evaluate performance along two major aspects: computation time and memory requirement.

Regarding computation time, we consider the timings reported by each implementation, from which we remove the input/output times (for reading the input from disk and writing the output to disk).
We also do \emph{not} include the pre-processing time dedicated to boundary matrix computation, for the implementations which require this input form.
Thus, our timings only include the core computation phase and we report in the following computation speeds, expressed in number of simplices per second. To enable an acceptable overall runtime (for the entire benchmark), we decided to interrupt all computations after a pre-defined timeout threshold of 15 minutes for all experiments.

Memory usage is evaluated with Python's standard library \href{https://docs.python.org/3/library/resource.html}{resources} and we report the maximum resident set size of each implementation (run in a dedicated subprocess).


\begin{figure}
  \centering
  \resizebox{\linewidth}{!}{
    \input{plots/plot_expl_para}
  }
  \mycaption{\majorRevision{Benchmark of parallel computation speeds
(simplices/second, log scale, 8 cores) as a function of the output size.}
}
  \label{fig_perfComparisonParallel}
\end{figure}

\subsubsection{Benchmark results}
\label{sec_benchmarkResults}

\autoref{fig_perfComparisonSequential} first reports, for each input dimension,
the computation speed
in sequential mode for all the (non timed out) implementations supporting simplicial complexes natively, for which no parallelism was observed or for which the number of threads could be explicitly set to one.
There, as can be expected, the only non-C++ implementation (\emph{Eirene.jl}) provides the lowest speeds.
Overall, the other C++ based implementations report computation speeds between 
$10^5$ and $10^8$ simplices per second. \emph{PHAT} and our method \emph{DMS} 
report the fastest sequential runtimes for all dimensions, with \emph{DMS} 
improving over \emph{PHAT} in 1D and 2D by $48\%$ and $51\%$ respectively, while 
\emph{PHAT} provides the best average sequential times in 3D, with a gain of 
\majorRevision{$38\%$}
over \emph{DMS}.
In \autoref{fig_perfComparisonSequential},
\emph{PersistenceCycles} \cite{iuricich21} is the method which is
the most related
to our approach.
However, similar to other
\majorRevision{DMT-based}
approaches
\cite{robins_pami11, GuntherRWH12,
perseusPaper, perseus}, it computes the \emph{full} Morse
complex
(prior to running a standard boundary matrix  reduction on it). In contrast, our 
approach computes only the necessary \emph{subparts} of the Morse complex: the 
(1D) unstable sets of $1$-saddles for $\diagram_0(f)$, the (1D) stable sets of 
$2$-saddles for $\diagram_2(f)$ -- which are much smaller than their (2D) 
unstable sets, and the (2D) unstable sets for only a \emph{small subset} of the 
$2$-saddles, specifically, only those involved in
$\diagram_1(f)$. This careful
selection already provides a significant performance gain
\majorRevision{(Appendix C)}. Next, the
construction of $\diagram_0(f)$ and $\diagram_2(f)$ uses a Union-Find 
data-structure, which is much more efficient than boundary matrix reduction. 
Overall, this results in runtime gains of $92\%$ and 
\majorRevision{$89\%$}
of \emph{DMS} over \emph{PersistentCycles} in 2D and 3D respectively.


Next, \autoref{fig_perfComparisonParallel} reports, for each input dimension,
the computation speed
in parallel mode (using 8 cores) for all the (non timed out) implementations supporting simplicial complexes natively, for which parallelsim was observed (typically using all available cores, in certain phases of the algorithm, for instance sorting) or for which the number of threads could be controlled explicitly.
In this figure, note that only the pre-processing sorting step of Gudhi benefits from parallelism, the core of its algorithm being sequential. Similarly to the sequential case, the only non-C++ implementation (\emph{JavaPlex}) provides the lowest speeds, as can be expected. The other C++ based implementations all report computation speed increases when parallelism is activated.
In comparison to \emph{PersistenceCycles} specifically (the other method of \autoref{fig_perfComparisonParallel} based on
DMT),
our method \emph{DMS} improves runtimes by $87\%$ in 2D and 
\majorRevision{$92\%$}
in 3D.
Overall, \emph{DMS} reports the fastest parallel runtimes for all dimensions, 
improving runtimes by 
\majorRevision{$74\%$},
\majorRevision{$77\%$}
and 
\majorRevision{$56\%$}
over the fastest competing technique, in 1D, 2D and 3D 
respectively.
These fastest runtimes can be explained by an improved parallel efficiency over competing techniques. In particular, the discrete gradient computation is trivially parallel, while several other steps
of our approach are also parallelized efficiently (\autoref{sec_parallelism}).

\begin{figure}
  \centering
  \resizebox{\linewidth}{!}{
    \begin{tikzpicture}
\begin{groupplot}[
  group style={group name=plots,},
  xlabel=\#cores]

\nextgroupplot[ymode=log, ylabel=Computation speed (simplices/second),xtick=data, xticklabels = {1,,,,, 32, 64,, 128},]
\addplot[curve1] coordinates { (1, 19067899.20481147) (2, 34753594.76818409) (4, 59001748.34606986) (8, 91615460.15811472) (16, 111921954.99868363) (32, 95552381.15507115) (64, 99790463.06450187) (96, 105542514.76973209) (128, 105071323.53515023) };
\addlegendentry{DMS}
\addplot[curve2] coordinates { (1, 1506949.1517428833) (2, 2551067.4236052823) (4, 3720524.0721502826) (8, 6049028.844719571) (16, 9968445.36425728) (32, 15652415.147760352) (64, 20442404.08616163) (96, 21210058.09240151) (128, 23488743.866013262) };
\addlegendentry{Dipha}
\addplot[curve3] coordinates { (1, 13325946.921204612) (2, 11052195.950335467) (4, 12019771.462772356) (8, 12252788.149698095) (16, 4948253.288240948) (32, 4701278.415262371) (64, 4227025.438665879) (96, 4042180.501221805) (128, 3954421.2260618955) };
\addlegendentry{\majorRevision{PHAT (Spectral Seq.)}}
\addplot[curve4] coordinates { (1, 1506432.8801600127) (2, 2751526.0716485926) (4, 4430832.208958508) (8, 7879996.83369648) (16, 12915119.706902787) (32, 15063579.293027531) (64, 20072738.4934323) (96, 22855792.64250456) (128, 23434474.64075437) };
\addlegendentry{PersistenceCycles}
\addplot[curve5] coordinates { (1, 13486098.717061162) (2, 20120540.975189112) (4, 29533690.147211857) (8, 37975126.84442636) (16, 32488506.731598523) (32, 29354801.86071993) (64, 27221991.844297636) (96, 26490863.187063366) (128, 25428772.75623553) };
\addlegendentry{TTK-FTM}
\legend{}
\nextgroupplot[legend to name=grouplegend_mesu, ymode=log, xtick=data, xticklabels = {1,,,,, 32, 64,, 128},]
\addplot[curve1] coordinates { (1, 2922730.4615708804) (2, 5436921.077134138) (4, 9597865.09351434) (8, 16618984.962338712) (16, 21198264.46934139) (32, 25411796.161951676) (64, 26579102.110863704) (96, 22231062.68707899) (128, 20686288.44994761) };
\addlegendentry{DMS}
\addplot[curve2] coordinates { (1, 776664.3670115144) (2, 1314458.0219663098) (4, 1979949.0956496857) (8, 3361054.12828399) (16, 5389695.610865479) (32, 4935783.764614987) (64, 6413806.055749778) (96, 10711594.211294744) (128, 10549531.718541494) };
\addlegendentry{Dipha}
\addplot[curve3] coordinates { (1, 6936370.029197979) (2, 6478859.456755871) (4, 7253816.957930937) (8, 7329361.269603599) (16, 2928111.7268160474) (32, 2315464.654972316) (64, 2254980.670386993) (96, 2213562.2183111752) (128, 2172371.3393607945) };
\addlegendentry{\majorRevision{PHAT (Spectral Seq.)}}
\addplot[curve4] coordinates { (1, 331223.6387191613) (2, 602890.5767477149) (4, 832492.8940776959) (8, 1502585.894872712) (16, 2463333.6432396434) (32, 2273828.369993428) (64, 4017967.831500132) (96, 5483557.654920317) (128, 6946362.239693834) };
\addlegendentry{PersistenceCycles}
\addlegendimage{curve5}
\addlegendentry{TTK-FTM}

\end{groupplot}
\node at (plots c2r1.east)[inner sep=0pt, xshift=12ex]
{\pgfplotslegendfromname{grouplegend_mesu}};
\end{tikzpicture}
  }
  \mycaption{\majorRevision{Benchmark of parallel scalability (average speed,
  as a function of the number of used cores) in 2D (left) and 3D (right).}
}
  \label{fig_perfComparisonParallelMESU}
\end{figure}

We further investigate parallel scalability in \autoref{fig_perfComparisonParallelMESU}, which reports for the most efficient parallel implementations, in 2D (left) and 3D (right), their average computation speeds (average over all datasets) when increasing the number of used cores.  This figure indicates that no implementation scales significantly when increasing the number of cores and that most implementations reach a high plateau of speed (essentially due to the remaining sequential parts of the algorithms) beyond $32$ cores in 2D and $96$ cores in 3D.
Moreover, \emph{PHAT}, which presented encouraging parallel performances on a desktop computer, seems to suffer drastically from NUMA effects on our large system, resulting in an absence of parallel acceleration.
Overall, \emph{DMS} reports the fastest performances on this system, improving runtimes with 128 cores by $76\%$ and $52\%$ over the fastest competing technique, in 2D and 3D respectively.

Together, Figures \ref{fig_perfComparisonParallel} and \ref{fig_perfComparisonParallelMESU}
also indicate that \emph{DMS} is more versatile than other approaches, as it
outperforms the most
\minorRevision{appropriate}
implementation for each system (\emph{PHAT} for the desktop computer, and \emph{DIPHA} for the large system).

Finally, \autoref{table_memoryFootPrint} reports the memory footprint for all the implementations supporting simplicial complexes natively, and for which the computation completed successfully.
There, one can observe that the methods supporting parallelism have a very similar (if not identical) memory footprint when parallelism is activated. Overall, the methods taking a boundary matrix as an input tend to have the largest memory footprints.
In contrast, \emph{DMS} uses \emph{TTK}'s
internal
triangulation data-structure \cite{ttk17} for modeling the input simplicial complex, which can be interpreted as a sparse representation of the boundary matrix,
resulting in substantial improvements over the most competitive techniques, by $25\%$, $5\%$ and $15\%$ in 1D, 2D and 3D respectively.

\subsection{Limitations}
\begin{table}
        \caption{\majorRevision{Maximum memory footprint over a single run for
each implementation, in mega-bytes (average over all datasets, bold: smallest
footprint).}}
\vspace{-1ex}
\rowcolors{3}{gray!20}{white}
        \resizebox{\linewidth}{!}{
        \begin{tabular}{|l|rrr|rrr|}
\hline
Implementation & Seq. 1D & Seq. 2D & Seq. 3D & Para. (8c) 1D & Para. (8c) 2D & Para. (8c) 3D\\\hline
Dionysus2 & 417.9 & 19,626.2 & 31,418.0 & NA & NA & NA\\
DIPHA& 271.1 & 11,835.0 & 19,672.1 & NA & 12,380.6 & 20,597.5\\
Eirene.jl& 43,885.9 & NA & NA & NA & NA & NA\\
Gudhi& NA & NA & NA & 246.1 & 10,644.8 & 16,770.3\\
JavaPlex& NA & NA & NA & 1,676.7 & NA & NA\\
\majorRevision{PHAT (Spectral Seq.)} & 251.0 & 10,399.1 & 15,326.8 & 250.0
& 10,396.9 & 15,472.1\\
PersistenceCycles & NA & 13,153.7 & 32,505.8 & NA & 13,154.4 & 32,878.9\\
TTK-FTM& NA & 5,692.9 & NA & NA & 5,701.8 & NA\\
\hline
DiscreteMorseSandwich & \textbf{188.3} & \textbf{5,388.9} & \textbf{12,996.7} & \textbf{188.2} & \textbf{5,390.7} & \textbf{13,312.6}\\
\hline
\end{tabular}
}
\label{table_memoryFootPrint}
\end{table}


Similarly to the orignal algorithm
\majorRevision{\emph{``PairSimplices''}
\cite{edelsbrunner02, zomorodianBook}}, our variant
\emph{``PairCriticalSimplices''} can work in principle in arbitrary dimension.
However, Robin's
expansion \majorRevision{algorithm} \cite{robins_pami11} provides strong
guarantees -- regarding the correspondence between critical simplices and PL
critical points -- for input datasets in up to three dimensions. Beyond, such a
correspondence is no longer guaranteed and our \emph{zero-persistence skip}
procedure (\autoref{algo_skippingZeroPersistence}) may no longer be valid.


%

\section{Conclusion}
This paper introduced an efficient algorithm for the computation of persistence diagrams for
\julien{scalar}
data.
Specifically, we
\majorRevision{documented}
a stratification strategy, which \emph{(i)} computes
the \emph{easiest} diagrams first ($\diagram_0(f)$ and
$\diagram_{\dimensionality-1}(f)$) with an efficient Union-Find based processing
applied to a carefully selected subset of the stable and unstable sets of
saddles and which then \emph{(ii)} efficiently computes the remaining diagram
($\diagram_1(f)$) by revisiting the seminal algorithm
\majorRevision{\emph{``PairSimplices''}
\cite{edelsbrunner02, zomorodianBook}}
in the context of discrete Morse
theory.
\majorRevision{Experiments} on 36 public datasets validated the performance
improvements of our approach over the
\majorRevision{algorithm \majorRevision{\emph{``PairSimplices''}}},
with two orders of magnitude speedups in 3D. A comprehensive benchmark
including 14 public implementations for persistent homology computation
indicated that our approach provides
the lowest memory footprints, as well as
the fastest parallel performances. Additionally, our experiments illustrated the versatility
of our approach,
as it outperforms (in 1D, 2D and 3D)
\minorRevision{the most appropriate methods for}
each tested system (\emph{PHAT} for
the desktop computer and \emph{DIPHA} for the large system), providing users
with performance confidence  irrespective of their system.

%


In the future, we will investigate alternative strategies for discrete gradient computation, in order to extend our \emph{zero-persistence skip} procedure to arbitrary dimensions.
Moreover, we will explore strategies for distributed computation,  to further improve
parallel scalability on large systems.
\ifCLASSOPTIONcaptionsoff
  \newpage
\fi



%
%
%

\section*{Acknowledgments}
\small{
This work is partially supported by the
European Commission grant
ERC-2019-COG
\emph{``TORI''} (ref. 863464,
\url{https://erc-tori.github.io/}). The authors thank Joshua Levine
\majorRevision{for
proofreading, and the reviewers for their constructive and helpful feedback}.
}

\bibliographystyle{abbrv-doi}
\bibliography{paper}

\begin{IEEEbiography}[{\includegraphics[width=1in,height=1.25in,clip,
keepaspectratio]{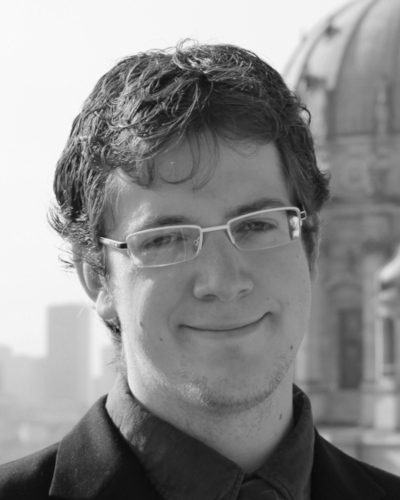}}]{Pierre Guillou}
is a research engineer at Sorbonne Université. After graduating from
MINES ParisTech, a top French engineering school in 2013, he received
his Ph.D., also from MINES ParisTech, in 2016. His Ph.D. work revolved
around parallel image processing algorithms for embedded
accelerators. Since 2019, he has been an active contributor to TTK and
the author of many modules created for the VESTEC and TORI projects.
\end{IEEEbiography}

\begin{IEEEbiography}[{\includegraphics[width=1in,height=1.25in,clip,
keepaspectratio]{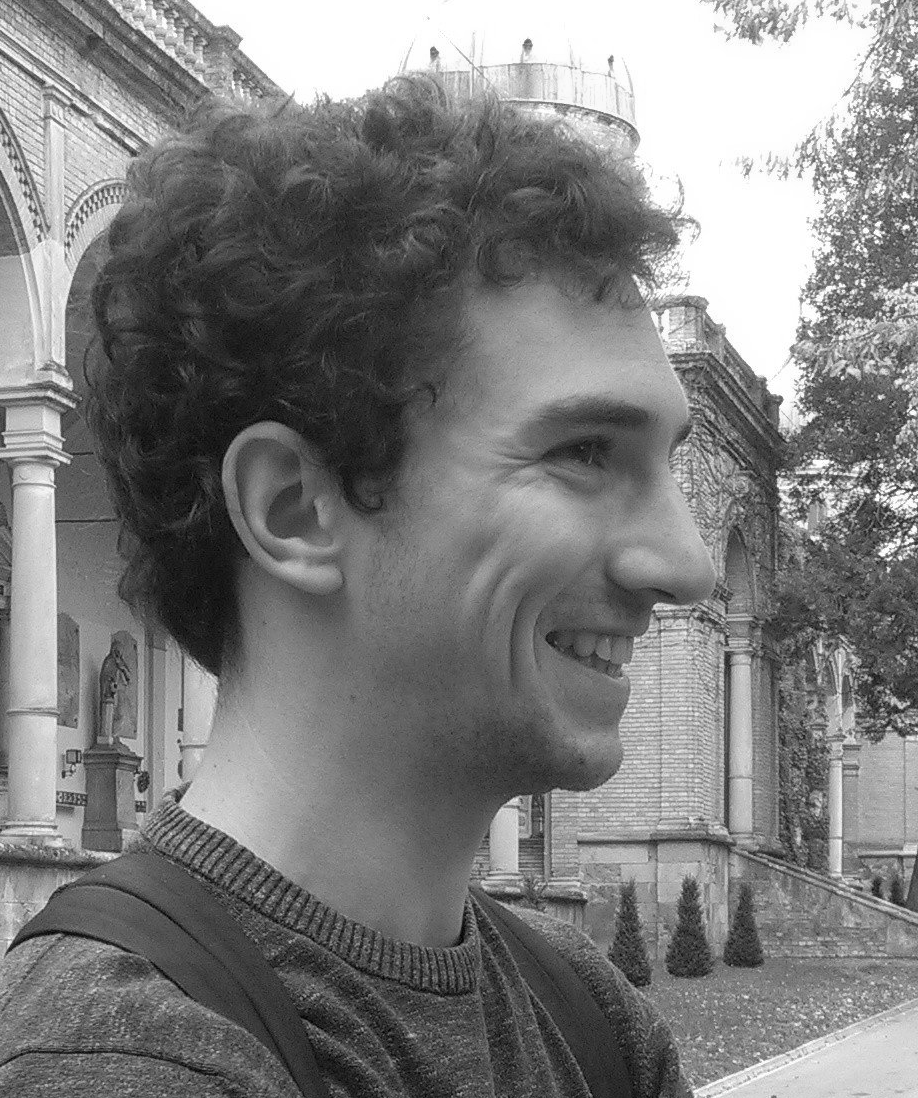}}]{Jules Vidal}
is a post-doctoral researcher, currently at Sorbonne Université, from where he received the Ph.D. degree in 2021.
He received the engineering degree
in 2018 from ENSTA Paris.
He is an active contributor to
the Topology ToolKit
(TTK), an open source library for
topological data analysis.
His notable contributions to TTK include the
efficient and progressive approximation of distances, barycenters and
clusterings of persistence diagrams.
\end{IEEEbiography}



\begin{IEEEbiography}[{\includegraphics[width=1in,height=1.25in,clip,
keepaspectratio]{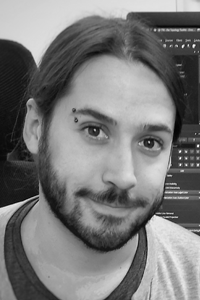}}]{Julien Tierny}
received the Ph.D. degree in Computer Science from the University
of Lille in 2008 and the Habilitation degree (HDR) from Sorbonne University in
2016. He is currently a CNRS research director, affiliated with
Sorbonne University. Prior to his CNRS tenure, he held a
Fulbright fellowship (U.S. Department of State) and was a post-doctoral
researcher at the Scientific Computing and Imaging Institute at the University
of Utah.
His research expertise lies in topological methods for data analysis
and visualization.
He is the founder and lead developer of the Topology ToolKit
(TTK), an open source library for topological data analysis.
\end{IEEEbiography}

%
%
%






\newpage
\setcounter{section}{0}
\section*{Appendix}
\includegraphics{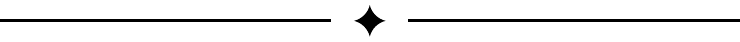}

\appendices

\usepgfplotslibrary{groupplots}
\pgfplotsset{
  every axis/.append style={
    no markers,
    grid=major,
    grid style={dashed},
    legend style={font=\tiny},
    ylabel style={font=\scriptsize},
    xlabel style={font=\scriptsize},
  },
  every axis plot/.append style={line width=1.2pt, line join=round},
  every axis legend/.append style={legend columns=1},
  group/group size=1 by 3,
  every x tick label/.append style={alias=XTick,inner xsep=0pt},
  every x tick scale label/.style={at=(XTick.base east),anchor=base west}
}

\definecolor{col1}{RGB}{53, 110, 175}
\definecolor{col2}{RGB}{204, 42, 42}
\definecolor{col3}{RGB}{255, 175, 35}
\definecolor{col4}{RGB}{79, 162, 46}
\definecolor{col5}{RGB}{97, 97, 97}
\definecolor{col6}{RGB}{103, 63, 153}
\definecolor{col7}{RGB}{0, 0, 0}
\definecolor{col8}{RGB}{123, 63, 0}

\tikzset{
  curve1/.style={col1},
  curve2/.style={col2},
  curve3/.style={col4},
  curve4/.style={col3},
  curve5/.style={col6},
  curve6/.style={cyan},
  curve7/.style={col5, densely dotted},
  curve8/.style={col7, densely dotted},
  curve9/.style={col8, densely dotted},
  curve10/.style={teal, densely dotted},
  curve11/.style={lime},
  curve12/.style={orange},
}

\section{Boundary perturbations}
\begin{figure}
  \includegraphics[width=\linewidth]{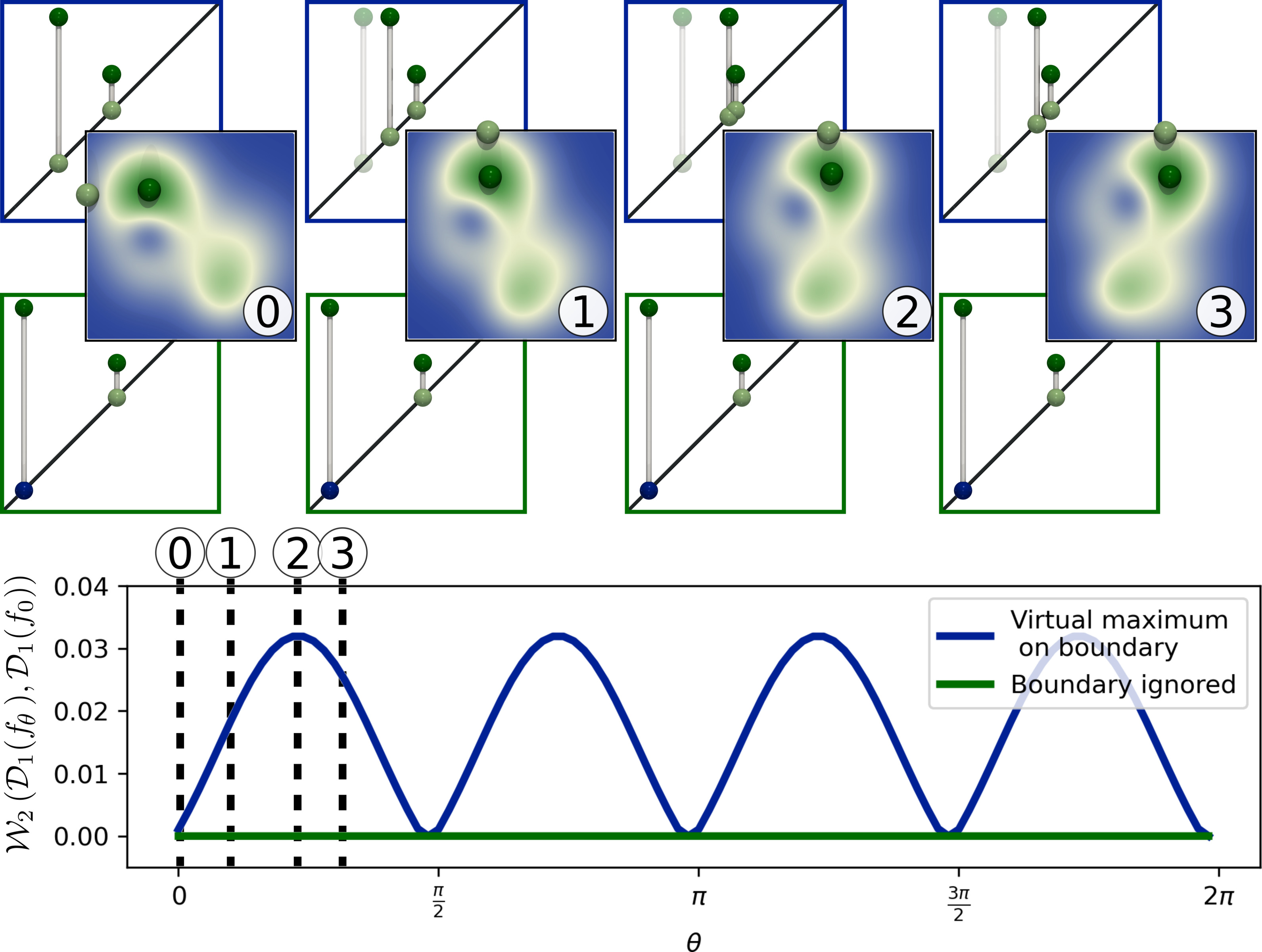}
  \caption{Instability in $\diagram_{\dimensionality-1}(f_\theta)$ induced by boundary perturbations.}
  \label{fig_boundaryHandling}
\end{figure}

As discussed in the section 5.2 of the main manuscript, the insertion of a virtual maximum on the outer boundary component of the data is optional, as it may result in an unstable assessment of the importance of the global maximum.
\autoref{fig_boundaryHandling} shows
the toy example terrain from
Fig. 2 (main manuscript)
which is rotated in the plane and cropped back to a square, for a varying angle $\theta$ (left to right), hence simulating
  a typical boundary data-cutting artifact observed in real-life data.
  In the exact diagram $\diagram_{\dimensionality-1}(f_\theta)$ (top, computed by the introduction of a virtual maximum on the outer boundary $\partial \domain$ of $\domain$,
  Sec. 5.2 of the main manuscript),
  the global maximum (dark green sphere in the data) is paired with a boundary saddle (light green sphere in the data), whose function value is dictated by the shape of the boundary (of the cut). As $\theta$ increases, the corresponding bar in $\diagram_{\dimensionality-1}(f_\theta)$ oscillates horizontally  (transparent: initial position for $\theta = 0$). Thus, the $L_2$-Wasserstein distance (blue curve, bottom,
  Sec. 2.4 of the main manuscript)
  to the original diagram $\diagram_{\dimensionality-1}(f_0)$ also oscillates with $\theta$.
  In contrast, by optionally disabling the introduction of a virtual maximum on $\partial \domain$ (center), the global maximum always gets paired (by convention,
  Sec. 6 of the main manuscript)
  to the global minimum, inducing a zero $L_2$-Wasserstein distance throughout. From our experience,
  this latter
  strategy (center) provides in practice a more stable assessment of the importance of the global maximum.

\begin{figure}
  \centering
  \resizebox{\linewidth}{!}{
  \input{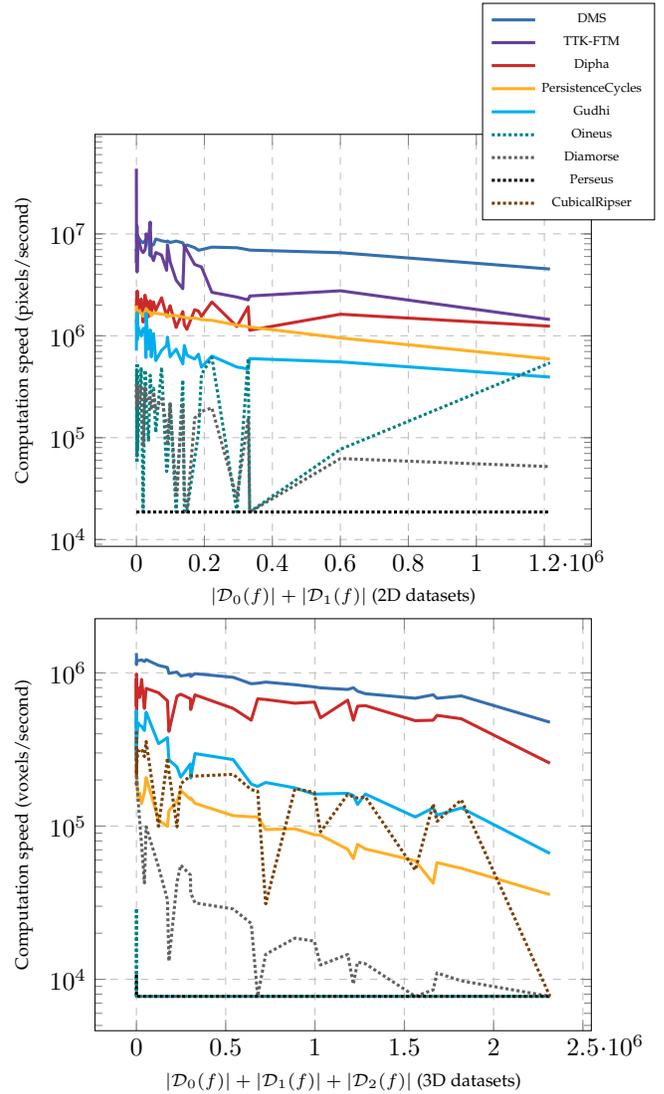}
  }
  \caption{\majorRevision{Computation speeds (top: 2D data, in pixel/s --
bottom: 3D data, in voxel/s, 8 cores) as a function of the output size.}
  }
  \label{fig_perfComparisonRegularGrid}
\end{figure}

\section{Regular grids}
As discussed in the main manuscript,
 some of the public implementations considered in our benchmark are specialized (or include specialized backends) for regular grids. However, they do not all interpret the input data in a consistent manner. For instance, some implementations (such as Gudhi) consider the input scalars to be defined on a per voxel basis, while others (such as Dipha or CubicalRipser) consider them as defined on a per vertex basis, which results in cell complexes of
 significantly
 different sizes (in particular, penalizing Gudhi).
Moreover, some implementations (such as Oineus,
\majorRevision{PairSimplices},
PersistenceCycles, TTK-FTM, DMS) implicitly triangulate the input
regular grid data \cite{freudenthal42, kuhn60}, which also changes the size of
the input complex.
First, since these internal data representations differ, the generated outputs will, consequently, not be exactly identical. Second, since these differences in internal representation result in cell complexes of
significantly
different sizes, they also induce a strong bias in runtime comparison.
For these reasons, we decided to focus our analysis on the methods which natively support simplicial complexes, for which a direct and unbiased comparison can be performed.

For completeness, we provide performance numbers for regular grids in
\autoref{fig_perfComparisonRegularGrid},
but
we would like to stress that the inconsistencies in the internal representations and in the generated outputs prevent a direct and unbiased comparison.




%

\pgfplotsset{group/group size=3 by 1,}

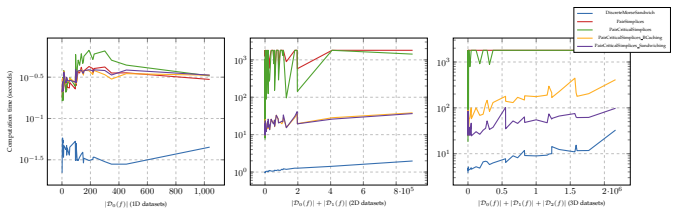
\begin{figure}
  \centering
  \resizebox{\linewidth}{!}{\begin{tikzpicture}
\begin{groupplot}[group style={group name=plots,}]

\nextgroupplot[ymode=log, ylabel=Computation time (seconds),
xlabel=\(|\diagram_0(f)|\) (1D datasets),]
\addplot[curve1] coordinates { (1, 0.022) (3, 0.058) (5, 0.058) (7, 0.054) (7, 0.051) (7, 0.054) (7, 0.04) (11, 0.034) (13, 0.047) (27, 0.042) (33, 0.047) (33, 0.036) (37, 0.044) (52, 0.036) (55, 0.045) (94, 0.053) (95, 0.032) (96, 0.031) (99, 0.038) (99, 0.05) (107, 0.031) (108, 0.032) (114, 0.031) (124, 0.034) (136, 0.037) (146, 0.039) (151, 0.029) (154, 0.033) (190, 0.031) (217, 0.032) (219, 0.034) (226, 0.034) (301, 0.03) (348, 0.028) (452, 0.028) (1037, 0.045) };
\addlegendentry{DiscreteMorseSandwich}
\addplot[curve2] coordinates { (1, 0.18) (3, 0.222) (5, 0.243) (7, 0.17) (7, 0.224) (7, 0.227) (7, 0.248) (11, 0.229) (13, 0.328) (27, 0.222) (33, 0.222) (33, 0.287) (37, 0.239) (52, 0.234) (55, 0.272) (94, 0.228) (95, 0.3) (96, 0.396) (99, 0.309) (99, 0.251) (107, 0.31) (108, 0.327) (114, 0.367) (124, 0.355) (136, 0.417) (146, 0.353) (151, 0.36) (154, 0.377) (190, 0.426) (217, 0.402) (219, 0.356) (226, 0.403) (301, 0.42) (348, 0.355) (452, 0.357) (1037, 0.297) };
\addlegendentry{PairSimplices}
\addplot[curve3] coordinates { (1, 0.162) (3, 0.218) (5, 0.282) (7, 0.165) (7,
0.213) (7, 0.221) (7, 0.316) (11, 0.242) (13, 0.38) (27, 0.208) (33, 0.219) (33,
0.305) (37, 0.256) (52, 0.287) (55, 0.312) (94, 0.245) (95, 0.333) (96, 0.594)
(99, 0.401) (99, 0.296) (107, 0.36) (108, 0.412) (114, 0.509) (124, 0.441) (136,
0.608) (146, 0.464) (151, 0.497) (154, 0.565) (190, 0.669) (217, 0.518) (219,
0.532) (226, 0.603) (301, 0.657) (348, 0.508) (452, 0.443) (1037, 0.328) };
\addlegendentry{PairCriticalSimplices}
\addplot[curve4] coordinates { (1, 0.207) (3, 0.294) (5, 0.307) (7, 0.212) (7,
0.271) (7, 0.277) (7, 0.305) (11, 0.277) (13, 0.375) (27, 0.261) (33, 0.275)
(33, 0.293) (37, 0.287) (52, 0.278) (55, 0.307) (94, 0.272) (95, 0.333) (96,
0.389) (99, 0.324) (99, 0.263) (107, 0.297) (108, 0.349) (114, 0.375) (124,
0.378) (136, 0.39) (146, 0.353) (151, 0.335) (154, 0.383) (190, 0.379) (217,
0.341) (219, 0.368) (226, 0.381) (301, 0.339) (348, 0.3) (452, 0.349) (1037,
0.334) };
\addlegendentry{PairCriticalSimplices\_BCaching}
\addplot[curve5] coordinates { (1, 0.208) (3, 0.261) (5, 0.277) (7, 0.214) (7,
0.239) (7, 0.274) (7, 0.274) (11, 0.285) (13, 0.366) (27, 0.267) (33, 0.276)
(33, 0.316) (37, 0.291) (52, 0.283) (55, 0.289) (94, 0.277) (95, 0.314) (96,
0.391) (99, 0.364) (99, 0.269) (107, 0.329) (108, 0.369) (114, 0.376) (124,
0.381) (136, 0.386) (146, 0.356) (151, 0.353) (154, 0.379) (190, 0.389) (217,
0.378) (219, 0.371) (226, 0.388) (301, 0.381) (348, 0.336) (452, 0.386) (1037,
0.335) };
\addlegendentry{PairCriticalSimplices\_Sandwiching}
\legend{}
\nextgroupplot[ymode=log, xlabel=\(|\diagram_0(f)| + |\diagram_1(f)|\) (2D
datasets),]
\addplot[curve1] coordinates { (1, 0.955) (300, 0.98) (302, 0.974) (392, 0.971)
(861, 0.965) (991, 1.026) (1083, 1.022) (1601, 1.002) (2207, 1.011) (8167,
1.052) (11836, 1.044) (15922, 1.024) (17872, 1.069) (19214, 1.047) (19236, 1.02)
(23437, 1.071) (24109, 1.061) (28144, 1.106) (37301, 1.064) (49565, 1.057)
(53773, 1.082) (57654, 1.075) (58293, 1.12) (68682, 1.095) (80046, 1.145)
(82321, 1.114) (93329, 1.109) (102782, 1.19) (108573, 1.15) (111882, 1.225)
(129954, 1.181) (181747, 1.27) (195862, 1.268) (197243, 1.268) (408577, 1.424)
(900246, 1.976) };
\addlegendentry{DiscreteMorseSandwich}
\addplot[curve2] coordinates { (1, 1800) (300, 1266.997) (302, 1526.19) (392,
1243.064) (861, 1800) (991, 1333.241) (1083, 1589.221) (1601, 1800) (2207, 1800)
(8167, 662.185) (11836, 1800) (15922, 551.377) (17872, 1800) (19214, 741.919)
(19236, 1503.05) (23437, 763.203) (24109, 591.816) (28144, 1065.914) (37301,
1800) (49565, 1800) (53773, 948.437) (57654, 1800) (58293, 375.567) (68682,
1800) (80046, 1800) (82321, 741.292) (93329, 1800) (102782, 1800) (108573, 1800)
(111882, 1800) (129954, 894.37) (181747, 1800) (195862, 1800) (197243, 579.938)
(408577, 1800) (900246, 1800) };
\addlegendentry{PairSimplices}
\addplot[curve3] coordinates { (1, 7.361) (300, 442.961) (302, 26.994) (392,
37.35) (861, 1045.83) (991, 8.306) (1083, 9.564) (1601, 1800) (2207, 1108.724)
(8167, 89.331) (11836, 1800) (15922, 310.677) (17872, 311.771) (19214, 124.59)
(19236, 1800) (23437, 351.914) (24109, 266.052) (28144, 293.425) (37301,
1753.944) (49565, 1800) (53773, 785.658) (57654, 1800) (58293, 153.227) (68682,
1800) (80046, 1800) (82321, 289.637) (93329, 1800) (102782, 1800) (108573,
1312.193) (111882, 1800) (129954, 95.975) (181747, 1800) (195862, 1800) (197243,
143.431) (408577, 1800) (900246, 1423.647) };
\addlegendentry{PairCriticalSimplices}
\addplot[curve4] coordinates { (1, 9.376) (300, 16.19) (302, 11.192) (392,
12.401) (861, 20.43) (991, 11.935) (1083, 10.09) (1601, 25.405) (2207, 15.206)
(8167, 12.21) (11836, 19.038) (15922, 22.588) (17872, 14.487) (19214, 18.202)
(19236, 18.546) (23437, 18.272) (24109, 18.52) (28144, 13.692) (37301, 25.91)
(49565, 20.467) (53773, 20.783) (57654, 24.339) (58293, 16.282) (68682, 33.205)
(80046, 26.431) (82321, 20.942) (93329, 21.573) (102782, 19.525) (108573,
22.256) (111882, 28.855) (129954, 15.555) (181747, 28.544) (195862, 36.033)
(197243, 19.713) (408577, 28.316) (900246, 38.246) };
\addlegendentry{PairCriticalSimplices\_BCaching}
\addplot[curve5] coordinates { (1, 9.311) (300, 16.144) (302, 11.152) (392,
12.31) (861, 19.773) (991, 11.917) (1083, 9.995) (1601, 23.423) (2207, 15.223)
(8167, 12.128) (11836, 19.144) (15922, 22.276) (17872, 15.739) (19214, 17.388)
(19236, 26.028) (23437, 18.061) (24109, 18.232) (28144, 13.522) (37301, 25.822)
(49565, 20.004) (53773, 19.211) (57654, 24.502) (58293, 16.821) (68682, 33.052)
(80046, 26.225) (82321, 20.568) (93329, 21.478) (102782, 22.882) (108573,
22.039) (111882, 33.725) (129954, 15.29) (181747, 30.832) (195862, 41.206)
(197243, 19.467) (408577, 26.0) (900246, 36.632) };
\addlegendentry{PairCriticalSimplices\_Sandwiching}
\legend{}
\nextgroupplot[legend to name=grouplegend_variants_para, ymode=log,
xlabel=\(|\diagram_0(f)| + |\diagram_1(f)| + |\diagram_2(f)| \) (3D datasets),]
\addplot[curve1] coordinates { (1, 3.828) (760, 4.148) (806, 4.859) (1005,
4.579) (1274, 4.344) (4067, 4.634) (7827, 5.133) (9436, 4.566) (17135, 4.458)
(28065, 4.51) (47573, 4.936) (50002, 4.584) (119112, 5.187) (173304, 4.844)
(222200, 6.645) (259088, 6.832) (280358, 6.681) (315652, 5.792) (367246, 6.291)
(550981, 7.752) (552256, 6.436) (660398, 8.993) (728357, 8.149) (827884, 11.76)
(831188, 9.122) (999052, 8.942) (1160807, 9.28) (1186371, 9.806) (1227947,
10.034) (1228299, 14.298) (1335516, 12.355) (1567356, 10.947) (1585307, 15.318)
(1598252, 11.641) (1770220, 11.725) (2163126, 32.749) };
\addlegendentry{DiscreteMorseSandwich}
\addplot[curve2] coordinates { (1, 1072.523) (760, 891.125) (806, 1800) (1005,
1800) (1274, 1255.545) (4067, 1800) (7827, 1405.001) (9436, 1800) (17135, 1800)
(28065, 1800) (47573, 1800) (50002, 1800) (119112, 1800) (173304, 1800) (222200,
1800) (259088, 1800) (280358, 1800) (315652, 1800) (367246, 1800) (550981, 1800)
(552256, 1800) (660398, 1800) (728357, 1800) (827884, 1800) (831188, 1800)
(999052, 1800) (1160807, 1800) (1186371, 1800) (1227947, 1800) (1228299, 1800)
(1335516, 1800) (1567356, 1800) (1585307, 1800) (1598252, 1800) (1770220, 1800)
(2163126, 1800) };
\addlegendentry{PairSimplices}
\addplot[curve3] coordinates { (1, 18.139) (760, 25.937) (806, 1800) (1005, 1800) (1274, 25.452) (4067, 398.316) (7827, 26.598) (9436, 1800) (17135, 1800) (28065, 790.361) (47573, 1800) (50002, 1800) (119112, 1800) (173304, 914.441) (222200, 1800) (259088, 1800) (280358, 1800) (315652, 881.886) (367246, 1800) (550981, 1800) (552256, 1800) (660398, 1800) (728357, 1800) (827884, 1800) (831188, 1800) (999052, 1800) (1160807, 1800) (1186371, 1800) (1227947, 1800) (1228299, 1800) (1335516, 1800) (1567356, 1800) (1585307, 1800) (1598252, 1800) (1770220, 1800) (2163126, 1800) };
\addlegendentry{PairCriticalSimplices}
\addplot[curve4] coordinates { (1, 24.834) (760, 25.975) (806, 89.715) (1005, 58.394) (1274, 26.688) (4067, 33.091) (7827, 31.62) (9436, 35.064) (17135, 56.396) (28065, 61.831) (47573, 100.053) (50002, 57.795) (119112, 100.01) (173304, 68.395) (222200, 72.567) (259088, 121.548) (280358, 145.938) (315652, 83.626) (367246, 130.57) (550981, 178.471) (552256, 139.038) (660398, 130.271) (728357, 169.693) (827884, 148.224) (831188, 183.115) (999052, 175.667) (1160807, 189.537) (1186371, 296.562) (1227947, 213.227) (1228299, 167.913) (1335516, 225.405) (1567356, 443.923) (1585307, 214.828) (1598252, 178.448) (1770220, 203.816) (2163126, 411.481) };
\addlegendentry{PairCriticalSimplices\_BCaching}
\addplot[curve5] coordinates { (1, 24.145) (760, 28.187) (806, 84.785) (1005, 54.762) (1274, 25.894) (4067, 31.26) (7827, 31.357) (9436, 30.487) (17135, 37.84) (28065, 25.09) (47573, 46.286) (50002, 24.473) (119112, 30.215) (173304, 26.432) (222200, 32.517) (259088, 35.853) (280358, 51.883) (315652, 33.78) (367246, 37.586) (550981, 101.894) (552256, 34.907) (660398, 51.762) (728357, 45.696) (827884, 63.441) (831188, 48.013) (999052, 54.847) (1160807, 47.337) (1186371, 73.751) (1227947, 56.798) (1228299, 58.601) (1335516, 66.21) (1567356, 75.091) (1585307, 62.834) (1598252, 61.352) (1770220, 61.803) (2163126, 97.923) };
\addlegendentry{PairCriticalSimplices\_Sandwiching}

\end{groupplot}
\node at (plots c3r1.north east)[inner sep=0pt, xshift=-2ex, yshift=2ex]
  {\pgfplotslegendfromname{grouplegend_variants_para}};
\end{tikzpicture}}
  \caption{\majorRevision{Computation times (in seconds,
log scale),
as a function of the output
size, for the distinct accelerations of our
  algorithm, in comparison to the seminal algorithm
  \majorRevision{\emph{``PairSimplices''}}
  (red, Alg. 1).
Each non-red curve corresponds to a
specific variant of our algorithm:
  \emph{``PairCriticalSimplices''} (green, Alg. 3),
  with boundary caching (yellow, Sec. 4.2),
  with \emph{``Sandwiching''} (purple, Sec. 5.1),
  in parallel (blue, $8$ cores, Sec. 7.2).
  On average, our parallel algorithm
  computes in
  $0.04$
  (1D),
  $1.10$
  (2D) and
  $\textbf{5.21}$
  (3D) seconds,
and achieves a parallel efficiency of
$30.68\%$
(1D),
$59.50\%$
(2D),
and
$\textbf{78.89\%}$
(3D) for an
overall speedup
over
\majorRevision{\emph{``PairSimplices''}}
of
$\times 8$
in 1D,
$\times 1,046$
in 2D, and
$\times \textbf{323}$
in 3D.}}
\label{appendix_timeIntra}
\end{figure}

\begin{figure}
  \centering
  \resizebox{\linewidth}{!}{\input{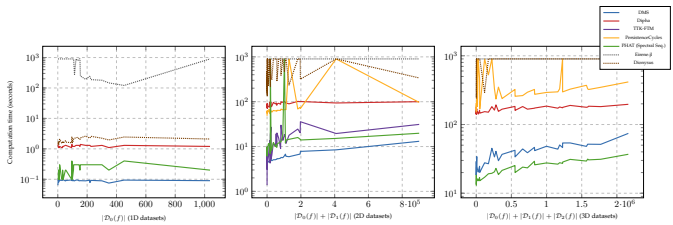}}
  \caption{\majorRevision{Benchmark of sequential computation times
(in seconds, log scale) as a function of the output size.}}
\label{appendix_timeSeq}
\end{figure}

\begin{figure}
  \centering
  \resizebox{\linewidth}{!}{\input{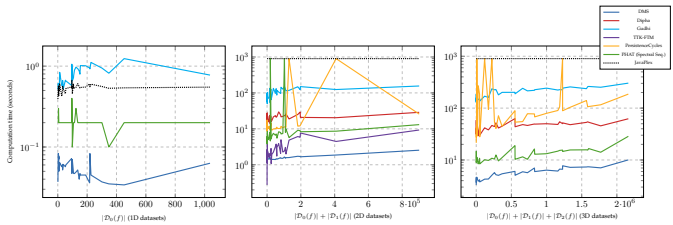}}
  \caption{\majorRevision{Benchmark of parallel computation times
(in seconds, log scale, 8 cores) as a function of the output size.}}
\label{appendix_timePara}
\end{figure}

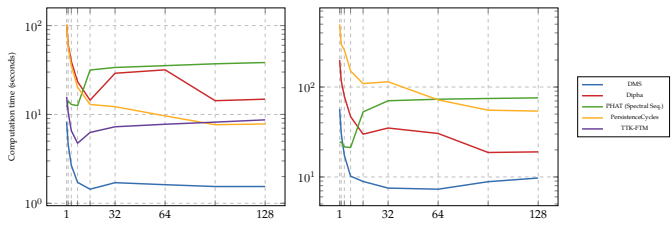
\begin{figure}
  \centering
  \resizebox{\linewidth}{!}{\begin{tikzpicture}
\begin{groupplot}[
  group style={group name=plots,},
  xlabel=\#cores]

\nextgroupplot[ymode=log, ylabel=Computation time (seconds),xtick=data,
xticklabels = {1,,,,, 32, 64,, 128},]
\addplot[curve1] coordinates { (1, 8.279972222222222) (2, 4.559333333333333) (4, 2.6808055555555548) (8, 1.720638888888889) (16, 1.4416666666666667) (32, 1.7074166666666668) (64, 1.6196388888888889) (96, 1.5448055555555558) (128, 1.5454444444444446) };
\addlegendentry{DMS}
\addplot[curve2] coordinates { (1, 102.25) (2, 61.191666666666684) (4, 39.64166666666667) (8, 23.37222222222222) (16, 14.480555555555558) (32, 29.08888888888889) (64, 31.772222222222222) (96, 14.283333333333328) (128, 14.852777777777776) };
\addlegendentry{Dipha}
\addplot[curve3] coordinates { (1, 12.108823529411763) (2, 14.141176470588233) (4, 12.952941176470588) (8, 12.552941176470588) (16, 31.629411764705885) (32, 33.78823529411765) (64, 35.42058823529412) (96, 37.15882352941177) (128, 38.40588235294117) };
\addlegendentry{PHAT (Spectral Seq.\_sequence)}
\addplot[curve4] coordinates { (1, 103.18444444444442) (2, 56.800638888888905)
(4, 35.40691666666666) (8, 20.24769444444444) (16, 12.956138888888887) (32,
12.222500000000002) (64, 9.64071875) (96, 7.666655172413793) (128,
7.815928571428571) };
\addlegendentry{PersistenceCycles}
\addplot[curve5] coordinates { (1, 15.792333333333337) (2, 10.555600000000002)
(4, 6.580611111111113) (8, 4.7461666666666655) (16, 6.273277777777779) (32,
7.26675) (64, 7.75952777777778) (96, 8.195914285714286) (128, 8.693777777777777)
};
\addlegendentry{TTK-FTM}
\legend{}
\nextgroupplot[legend to name=grouplegend_mesu, ymode=log, xtick=data,
xticklabels = {1,,,,, 32, 64,, 128},]
\addplot[curve1] coordinates { (1, 56.782388888888896) (2, 30.751) (4,
17.385194444444444) (8, 10.199583333333335) (16, 8.906138888888888) (32,
7.517333333333334) (64, 7.305944444444445) (96, 8.851861111111113) (128,
9.72927777777778) };
\addlegendentry{DMS}
\addplot[curve2] coordinates { (1, 199.39722222222218) (2, 118.30555555555556)
(4, 78.9111111111111) (8, 46.87499999999999) (16, 29.897222222222222) (32,
35.011111111111106) (64, 30.44444444444445) (96, 18.714285714285708) (128,
18.99722222222222) };
\addlegendentry{Dipha}
\addplot[curve3] coordinates { (1, 24.174999999999997) (2, 24.480555555555554)
(4, 21.60277777777778) (8, 21.25) (16, 52.897222222222226) (32,
70.35555555555554) (64, 73.19722222222222) (96, 74.58333333333333) (128,
75.79444444444445) };
\addlegendentry{\majorRevision{PHAT (Spectral Seq.)}}
\addplot[curve4] coordinates { (1, 497.6243437499999) (2, 297.28134374999996) (4, 258.69990322580645) (8, 150.54683870967742) (16, 109.20460714285714) (32, 114.43993103448277) (64, 71.95955555555554) (96, 55.28812500000001) (128, 53.990125) };
\addlegendentry{PersistenceCycles}
\addlegendimage{curve5}
\addlegendentry{TTK-FTM}

\end{groupplot}
\node at (plots c2r1.east)[inner sep=0pt, xshift=12ex]
{\pgfplotslegendfromname{grouplegend_mesu}};
\end{tikzpicture}}
  \caption{\majorRevision{Benchmark of parallel scalability (average
computation time,
  as a function of the number of used cores) in 2D (left) and 3D (right).}}
  \label{appendix_timeScalability}
\end{figure}

\section{Computation time statistics}
\majorRevision{This appendix provides further details regarding the computation
times of our approach.}

\majorRevision{First, Figures \ref{appendix_timeIntra}, \ref{appendix_timeSeq},
\ref{appendix_timePara}, \ref{appendix_timeScalability} convert the
computational speed from Figures 17, 18, 19 and 20 of the main manuscript into
computation times. This enables
the direct comparison of time performance
(instead of speed) for various implementations. For instance, in
Figures \ref{appendix_timeIntra}, \ref{appendix_timeSeq},
\ref{appendix_timePara}, each curve sample represents a single dataset from the
benchmark. Then, for a fixed X-coordinate (i.e. for a fixed dataset), one can
directly read the computation
time of each approach as the Y-coordinate of the corresponding curve. For
instance,
in \autoref{appendix_timePara}, the
rightmost point for the 3D curves (i.e. the largest output diagram)
corresponds to the
\emph{Random} stress case and reading the
corresponding Y-coordinate for each curve thus gives the computation time of the
corresponding implementation for
this
dataset.}

\begin{table}[t]
  \caption{\majorRevision{Computation statistics for our method
\emph{Discrete Morse Sandwich} on the benchmark datasets (minimum, average,
and maximum computation times),
in seconds (sequential
and parallel).}}
\label{tab_detailedStatistics}
  \rowcolors{3}{gray!20}{white}
  \resizebox{\linewidth}{!}{
\begin{tabular}[ht]{|r|l|rrr|rrr|}
\hline
& & \multicolumn{3}{c|}{Sequential} & \multicolumn{3}{c|}{Parallel (8c)} \\
\hline
 & & Min. & Avg. & Max.
 & Min. & Avg. & Max.
 \\
\hline
\hline
~ & Discrete Gradient & 0.038 & 0.045 & 0.067
& 0.010 & 0.011 & 0.013
\\
 & Sorting Edges \& Critical Simplices & 0.006 & 0.007 & 0.007
&
0.002 & 0.004
& 0.019
\\
\textbf{1D} &
$\diagram_0(f)$ \& $\diagram_2(f)$ (Sec. 5)
& 0.001 & 0.026 & 0.040
& 0.001 & 0.017 & 0.039
\\
~ &
$\diagram_1(f)$ (Sec. 4)
& 0.000 & 0.000 & 0.000
& 0.000 & 0.000 & 0.000
\\
~ & Initializations \& Allocations  & 0.014 & 0.009 & 0.001
& 0.011 &
0.010 & 0.006
\\
\hline
~ & \textbf{Total} & 0.059 & 0.087 & 0.115
& 0.024 & 0.042 & 0.077
\\
\hline
\hline
~ & Discrete Gradient & 3.041 & 3.322 & 4.521
& 0.503 & 0.565 &
0.775
\\
 & Sorting Edges \& Critical Simplices & 0.308 & 0.373 & 0.754
 & 0.034
& 0.048 &
0.125
\\
\textbf{2D} &
$\diagram_0(f)$ \& $\diagram_2(f)$ (Sec. 5)
& 0.023 & 0.743 & 4.363
& 0.021 & 0.155 & 0.753
\\
~ &
$\diagram_1(f)$ (Sec. 4)
& 0.000 & 0.000 & 0.000
& 0.000 & 0.000 & 0.000
\\
~ & Initializations \& Allocations & 0.413 & 0.428 & 0.627
& 0.363 &
0.400 & 0.322
\\
\hline
~ & \textbf{Total} & 3.785 & 4.866 & 10.265
& 0.921 & 1.168 & 1.975
\\
\hline
\hline
~
& Discrete Gradient & 12.293 & 24.674 & 32.933
& 1.761 & 2.665 &
3.223
\\
 & Sorting Edges \& Critical Simplices & 3.907 & 6.260 & 8.438
 & 0.620
& 0.814 &
1.052
\\
\textbf{3D} &
$\diagram_0(f)$ \& $\diagram_2(f)$ (Sec. 5)
& 0.019 & 1.922 & 5.194
& 0.018 & 0.346 & 1.038
\\
~ &
$\diagram_1(f)$ (Sec. 4)
& 0.000 & 3.098 & 24.331
& 0.000 & 0.432 & 2.450
\\
~ & Initializations \& Allocations & 2.180 & 1.504 & 3.371
& 0.908 &
1.303 & 2.274
\\
\hline
~ & \textbf{Total} & 18.399 & 37.458 & 74.267
& 3.307 & 5.560 & 10.037
\\
\hline
\end{tabular}
}

\end{table}

\majorRevision{
\autoref{tab_detailedStatistics} provides a breakdown of the computation times
of our approach, to further evaluate the individual performance of
each sub-step.
We provide the minimum, maximum
and average computation times of each sub-step over the benchmark datasets.
This table shows
that, for the benchmark datasets, the discrete gradient computation takes,
in 3D in sequential mode,
66\% of the computation time on average.}
\majorRevision{This can be explained by the fact that the zero persistence skip
procedure (Alg. 2, main manuscript) discards a very large number of
simplices from the computation. Then, the other steps of the approach typically
process a very small number of critical simplices (documented in the next
paragraph) and therefore tend to
only
represent a small fraction of the overall computation time.}


\majorRevision{This is
evaluated
in \autoref{stats_criticalSimplices}, which
provides the percentage of critical simplices (minimum, average and maximum
values) over the benchmark datasets.
There, the columns \emph{Min.} and
\emph{Max.} correspond to the \emph{Elevation} and \emph{Random} stress cases
respectively (i.e. the datasets with the smallest and largest number of critical
simplices). This table shows that, in 3D,
only 0.79\%  of the
input simplices are critical on average,
and therefore processed by the steps following the
discrete gradient computation.}

\begin{table}[t]
  \caption{\majorRevision{Percentage of critical simplices (minimum, average,
maximum values over the benchmark datasets) with regard to the total number of
simplices, per critical index.}}
  \label{stats_criticalSimplices}
    \rowcolors{3}{gray!20}{white}
      \resizebox{\linewidth}{!}{
  \centering
  \begin{tabular}[ht]{|ll|rrr|}
\hline
& Index & Min. & Avg. & Max. \\
\hline
\multirow{3}{*}{1D}
  & 0 & $48 \times 10^{-6}$ \%
  & 0.01 \%  & 0.05 \%  \\
\textbf{1D}  & 1 & $48 \times 10^{-6}$ \% & 0.01 \% & 0.05 \%  \\
\hline
  & \textbf{Total} & $96 \times 10^{-6}$ \% & 0.01 \%& 0.10 \%  \\
\hline
\hline
\multirow{4}{*}{2D} & 0 & $1 \times 10^{-6}$ \%& 0.04 \% & 0.50 \%  \\
\textbf{2D} & 1 &
$1 \times 10^{-6}$ \%
& 0.09 \%& 0.89 \%  \\
& 2 & 0.00 \% & 0.04 \%& 0.39 \%  \\
\hline
& \textbf{Total} & $2 \times 10^{-6}$ \% & 0.17 \%& 1.79 \%  \\
\hline
\hline
\multirow{5}{*}{3D} & 0 & $1 \times 10^{-6}$ \% & 0.11 \%& 0.45 \%  \\
& 1 & $1 \times 10^{-6}$ \% & 0.29 \%& 0.97 \%  \\
\textbf{3D} & 2 & 0.00 \%  & 0.28 \%& 0.95 \% \\
& 3 & 0.00 \%& 0.11 \%  & 0.44 \% \\
\hline
& \textbf{Total} & $2 \times 10^{-6}$ \% & 0.79 \% & 2.81 \% \\
\hline
\end{tabular}
}
\end{table}

\majorRevision{This table also shows that, in 3D, in the worst
case (\emph{Max.} column, which corresponds to the \emph{Random} dataset), the
number of minima (respectively maxima) is
at most $0.45 \%$ (respectively $0.44 \%$) of the total number of simplices.
Reported to the total number of vertices only, this means that, in practice,
at most $10 \%$ of the vertices are minima.
Similarly, at most $2 \%$ of the tetrahedra are maxima in practice.
This implies that in practice, our
fast computation of $\diagram_0(f)$ (Sec. 5.1, main manuscript) discards at
least $90 \%$
of the vertices in comparison to a standard Union-Find processing
\cite{edelsbrunner02} (and our computation of $\diagram_2(f)$ discards at
least $98 \%$ of the tetrahedra).}
%
\majorRevision{Moreover,
the same column (\emph{Max.}, 3D)
also shows that
the number of minima (respectively maxima) is
nearly half
that of $1$ (respectively $2$) saddles. This implies that our sandwiching
strategy (which efficiently computes $\diagram_0(f)$ and $\diagram_2(f)$, Sec.
5, main manuscript) discards half of the
$1$ and $2$-saddles
from the computation of the algorithm \emph{``PairCriticalSimplices''}
(Alg. 3, main manuscript).
Given the cubic worst time complexity of this step (Sec. 7.1,
main manuscript), this reduction has a significant
impact on computation times (up to a $\times (2^3)$ speedup for this sub-step).
Then, in practice,
the
algorithm \emph{``PairCriticalSimplices''} (Alg. 3, main manuscript) ends up
processing at most
$0.51 \%$
of the
simplices of the input simplicial complex.}


\majorRevision{As predicted by the time complexity analysis (Sec. 7.1,
main manuscript),
the computation of $\diagram_1(f)$ in sequential
(\autoref{tab_detailedStatistics}) takes typically more time than
the computation of  $\diagram_0(f)$ and $\diagram_2(f)$ combined
($1.6$
times on average
in sequential in 3D).
However, as shown on the right
side of \autoref{tab_detailedStatistics}, this
is mitigated in
practice by the good parallel
efficiency
of the computation of
$\diagram_1(f)$ (89.6\% on average).}

\ifCLASSOPTIONcaptionsoff
  \newpage
\fi



%
%
%



\end{document}